\documentclass[12pt]{article}
\usepackage[T1]{fontenc}
\usepackage{lmodern}
\usepackage{amsmath}
\usepackage{graphicx}
\usepackage{enumerate}
\usepackage{natbib}
\usepackage{url} 
\usepackage{xr}
\usepackage{authblk}
\usepackage{multirow}
\newcommand{\blind}{1}

\let\code=\texttt
\let\proglang=\textsf

\input{packages}
\usepackage{etoolbox}

\newcommand{\Mean}{{\mathbb{E}}}
\newcommand{\Var}{{\mbox{Var}}}
\newcommand{\Cov}{{\mbox{cov}}}

\newcommand{\prob}{{\mathbb{P}}}
\DeclareMathOperator*{\argmin}{arg\,min}

\newtheorem{thm}{Theorem}
\newtheorem{coro}{Corollary}

\begin{document}
	
	\def\spacingset#1{\renewcommand{\baselinestretch}%
		{#1}\small\normalsize} \spacingset{1}

	\if1\blind
	{
		\title{\bf Off-Policy Confidence Interval Estimation with Confounded Markov Decision Process}
		\author[1]{Chengchun Shi  \thanks{This paper is accepted at Journal of the American Statistical Association. CS' research is partly supported by the EPSRC grant EP/W014971/1. RS' research is partly supported by the NSF grants DMS-1555244 and DMS-2113637.}}
		\author[2]{Jin Zhu}
		\author[3]{Ye Shen}
		\author[4]{Shikai Luo}
		\author[5]{Hongtu Zhu}
		\author[6]{Rui Song}
		\affil[1]{London School of Economics and Political Science}
		\affil[2]{Sun Yat-sen University}
		\affil[3,6]{North Carolina State University}
		\affil[4]{Didi Chuxing}
		\affil[5]{University of North Carolina at Chapel Hill}
		\date{\empty}
		\maketitle
	} \fi
	
	\if0\blind
	{
		\bigskip
		\bigskip
		\bigskip
		\begin{center}
			{\LARGE\bf Off-Policy Confidence Interval Estimation with Confounded Markov Decision Process}
		\end{center}
		\medskip
	} \fi

	\bigskip
	
	\begin{abstract}
		This paper is concerned with constructing a confidence interval for a target policy's value offline based on a pre-collected observational data in infinite horizon settings. Most of the existing works assume no unmeasured variables exist that confound the observed actions. This assumption, however, is likely to be violated in real applications such as healthcare and technological industries. In this paper, we show that with some auxiliary variables that mediate the effect of actions on the system dynamics, the target policy's value is identifiable in a confounded Markov decision process. Based on this result, we develop an efficient off-policy value estimator that is robust to potential model misspecification and provide rigorous uncertainty quantification. Our method is justified by theoretical results, simulated and real datasets obtained from ridesharing companies. A \proglang{Python} implementation of the proposed procedure is available at \url{https://github.com/Mamba413/cope}.
	\end{abstract}

	\noindent%
	{\it Keywords:}  Reinforcement Learning; Off-Policy Evaluation; Statistical Inference; Unmeasured confounders; Infinite Horizons; Ridesourcing Platforms.
	\vfill
	
	\newpage
	\spacingset{1.5} 
	
	\setlength{\abovedisplayskip}{5.5pt}
	\setlength{\belowdisplayskip}{5.5pt}
	
	\section{Introduction}\label{secintroduction}
	We consider reinforcement learning (RL) where the goal is to learn an optimal policy that maximizes the (discounted) cumulative rewards the decision maker receives \citep{sutton2018reinforcement}. A (stationary) policy is a time-homogeneous decision rule that determines an action based on a set of observed state variables. Off-policy evaluation (OPE) aims to evaluate the impact of a given policy (called target policy) 
	using observational data generated by a potentially different policy (called behavior policy).
	OPE 
	is an important problem in settings where it is expensive or unethical to directly run an experiment that implements the target policy. This includes applications in 
	precision medicine \citep{murphy2003optimal,zhang2012robust,zhang2013robust,chakraborty2014dynamic,matsouaka2014evaluating,luedtke2016statistical,wang2018quantile,gottesman2019guidelines,wu2020resampling}, autonomous driving \citep{li2020make}, 
	robotics \citep{kober2013reinforcement}, natural language processing \citep{li2016deep}, education \citep{mandel2014offline}, among many others. 
	
	This paper is concerned with OPE under infinite horizon settings where the number of decision points is not necessarily fixed and is allowed to diverge to infinity. We remark that most works in the statistics literature focused on learning and evaluating treatment decision rules for precision medicine with only a few treatment stages \citep[see][for an overview]{tsiatis2019dynamic,kosorok2019precision}. These methods are not directly applicable to many other sequential decision making problems in reinforcement learning with infinite horizons \citep[see e.g.,][]{sutton2018reinforcement}, such as autonomous driving, robotics and mobile health (mHealth). Recently, there is a growing interest on policy learning and evaluation in mHealth applications {\citep{ertefaie2014constructing,luckett2020estimating,shi2020statistical,hu2020personalized,qi2020robust,xu2020latent,liao2020batch,liao2021off,shi2021deeply}}. In the computer science literature, existing works for OPE in infinite horizons can be roughly divided into three categories. The first type of method directly derives the value estimates by learning the system transition matrix or the Q-function under the target policy \citep[]{le2019batch,feng2020accountable,hao2021bootstrapping}. The second type of method is built upon importance sampling (IS) that re-weights the observed rewards with the density ratio of the target and behavior policies \citep{thomas2015high2,liu2018breaking,nachum2019dualdice,dai2020coindice}.
	The last type of method combines the first two for more robust and efficient value evaluation. References include \citet{jiang2016doubly,uehara2019minimax,kallus2019efficiently}. 
	In particular, \cite{kallus2019efficiently} develops a double reinforcement learning (DRL) estimator that achieves the semiparametric efficiency limits for OPE. Informally speaking,  a semiparametric efficiency bound can be viewed as the nonparametric extension of the Cramer-Rao lower bound in parametric models \cite{bickel1993efficient}. It lower bounds the asymptotic variance among all regular estimators \cite{van2000asymptotic}. 
	However, all the above cited works rely on the 
	sequential ignorability or the sequential randomization assumption \citep[see e.g.,][for a detailed definition]{robins2004optimal}. It essentially precludes the existence of unmeasured variables that confound the action-reward or action-next-state associations. However, this assumption is likely to be violated in applications such as healthcare and technological industries. We consider the following example to elaborate.

	

	Our work is motivated by the example of applying customer recommendation program in a ride-hailing platform. We consider evaluating the effects of applying certain customer recommendation program in large-scale ride-hailing platforms such as Uber, Lyft and Didi. These companies form a typical two-sided market that enables efficient interactions between passengers and drivers \citep{Rysman2009} and has substantially transformed the transportation landscape of human beings \citep{Jin2018}. 
	
	Suppose a customer launches a ride-hailing application on their smart phone. When they enter their destination, the platform will decide whether to recommend them to join a program. This corresponds to the action. Different programs will apply different coupons to the customer to discount this ride. The purpose of such recommendation is to (i) increase the chance that the customer orders this particular ride, and reduce the local drivers' vacancy periods; (ii) increase the chance that the customer uses the app more frequently in the future. We remark that (i) and (ii) correspond to the short-term and long-term benefits for the company, respectively. 
	
	We would like to evaluate the cumulative effect of a given customer recommendation program given an observational dataset collected from the ride-hailing company. In addition to a point estimate on a target policy's value, many applications would benefit from having a confidence interval (CI) that
	quantifies the uncertainty of the value estimates. For instance, it allows us to infer whether the difference between two policies' values is statistically significant. This motivates us to study the off-policy confidence interval estimation problem. 
	
	Confounding is a serious issue in data generated from these applications. This is because the behavior policy 
	involves not only an estimated automated policy to maximize the company's long term rewards but human interventions as well. 
	For example, when there is severe weather like thunderstorms or large events like sports games and concerts in a certain area,  there will be much more passengers than drivers in the local area. In that case, human interventions are needed to discourage passengers to request call orders in this
	area. However, live events and extreme weather are not recorded, 
	leading to a confounded dataset. 
	

	
	More recently, in the causal inference literature, a few methods have been proposed to deal with unmeasured confounders for treatment effects evaluation. 
	\cite{tchetgen2020introduction} proposed a proximal g-computation algorithm in single-stage and two-stage studies. \cite{shi2020multiply} proposed to learn the average treatment effect (ATE) with double-negative control adjustment. {See also \citet{kallus2021causal}.} These methods are not directly applicable to the infinite horizon setting, which is the focus of our paper. 
	In the RL literature, a few works considered reinforcement learning with  confounded datasets. 
	Among those available, 
	\cite{wang2020provably} considered learning an optimal policy in an episodic confounded MDP setting. 
	\cite{namkoong2020off} and \cite{kallus2020confounding} proposed partial identification bounds on the target policy's value under a single-decision
	confounding assumption and a memoryless unobserved confounding assumption, respectively. 
	\cite{bennett2020off} introduced an optimal balancing algorithm for OPE in a confounded MDP, without requiring the mediators to exist. {\cite{tennenholtz2020off}  adopted the POMDP model to formulate the confounded OPE problem and develop value
		estimators in tabular settings using the idea of proxy variables. More recently, there are a few works that extend their method to more general settings \citep{bennett2021proximal,nair2021spectral,shi2021minimax}.} However, none of the aforementioned methods considered constructing confidence intervals for the target policy's value in infinite horizons.

	In this paper, we model the observational data by a confounded Markov decision process \citep[CMDP,][]{zhang2016markov}. See Section \ref{sec:dgp} for a detailed description of the model. 
	To handle unmeasured confounders, we make use of some intermediate variables (mediators) that mediate the effect of actions on the system dynamics. These mediators are required to be conditionally independent of the unmeasured confounders given the actions. We remark that these auxiliary variables exist in several applications. 
		
		For instance, in the ride-hailing example, the mediator corresponds to the final discount applied to each ride. 
		It is worth mentioning that the final discount might be different from the discount included in the program, as it depends on other promotion strategies the platform applies to the ride, but is conditionally independent of other unmeasured variables that confound the action. 
		In addition, the action will affect the immediate reward and future state variables only through the mediator (see our real data section for a detailed definition of the immediate reward). Consequently, the mediator satisfies the desired condition. Predictive policing is another example. Consider the Crime Incidents dataset \citep{elzayn2019fair}. The action is whether a district is labeled as dangerous or not and the outcome is the total number of discovered crime incidents. Given the action, the police allocation (mediator) is determined by the current available policing resources and is thus conditionally independent of the confounder. In addition, in medicine, the treatment (action) and the patient's outcome might be confounded by that patient's attitude towards different treatments. For example, some patients might prefer conservative treatments, and others will strictly stick to the doctor's advice. However, given the treatment, 
		the dosage that patients receive is determined by their age, weight and clinical conditions, and is thus conditionally independent of the confounder. 
		
		To the best of our knowledge, this is the first paper that systematically studies off-policy confidence interval estimation under infinite horizon settings with unmeasured confounders. Most prior work either requires the unmeasured confounders assumption, or focuses on point estimation. More importantly, our proposal addresses an important practical question in ride-sharing companies, allowing them to evaluate different customer recommendation programs more accurately in the presence of unmeasured confounders. 
		Our proposal involves two key components. 
		We first show that in the presence of mediators, the target policy's value can be represented using the probability distribution that generates the observational data. 
		This result generalizes the front-door adjustment formula \citep[see e.g.,][]{pearl2009causality} 
		to infer the average treatment effect in single-stage decision making. 
		Based on this result, we next apply the semiparametric theory \citep[see e.g.,][]{tsiatis2007semiparametric} to derive the efficiency limits for OPE under CMDP with mediators, and outline a robust and efficient value estimate that achieves this efficiency bound and its associated CI. 

		The rest of the paper is organized as follows. Section \ref{sec:Preliminaries} lays out the basic model notation and data generating process. Section \ref{sec:causalOPE} discusses the identifiability of the policy value and construct efficient and robust interval estimation. Section \ref{SecTheory} presents the asymptotic properties of the proposed estimator with its inferential results. Section \ref{sec:simulation} presents two simulation
		studies to evaluate the performance of our proposed estimator and compare with the state-of-the-art methods by
		using synthetic data only. In Section \ref{sec:real data}, an application
		of the proposed estimator is used to analyze real data collected from a world-leading ride-hailing company. All proofs are given in the supplementary material.

		\section{Preliminaries} \label{sec:Preliminaries}
		\subsection{Data Generating Process}\label{sec:dgp}
\begin{figure}[htbp]
			\centering
			\begin{subfigure}[b]{0.45\textwidth}
				\centering
				\includegraphics[width=\textwidth]{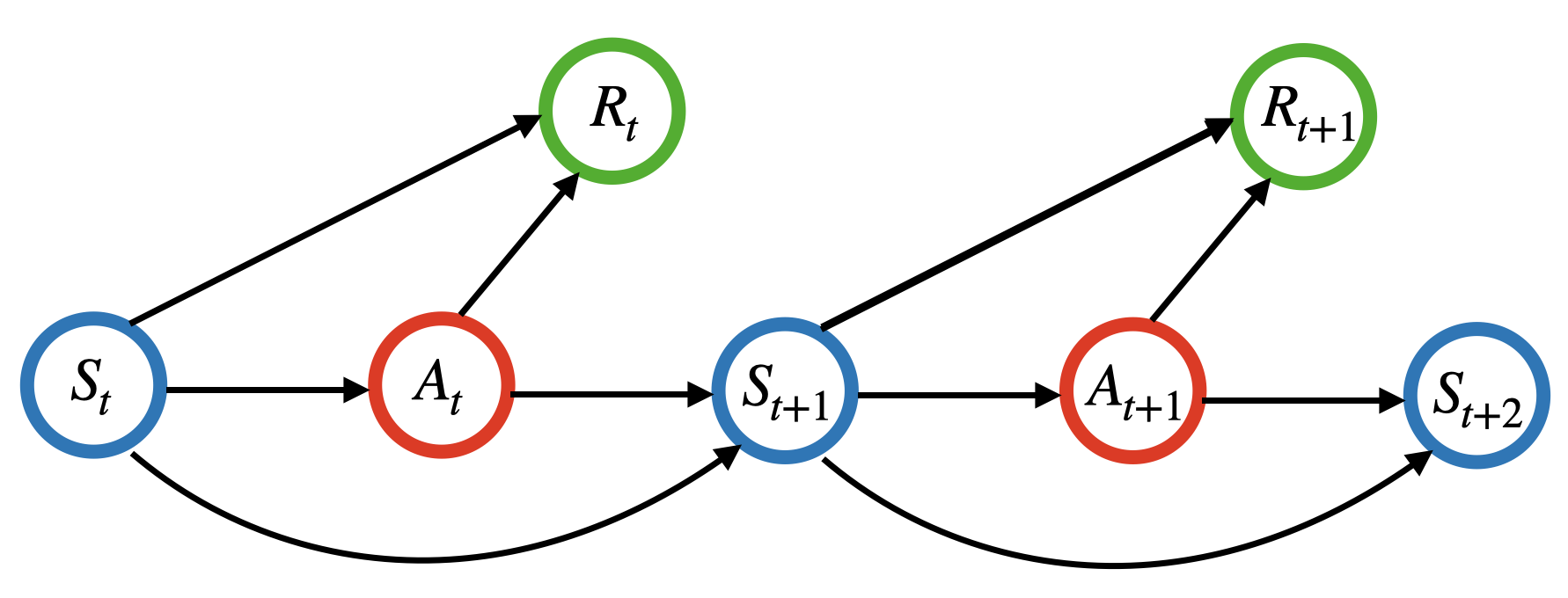}
				\caption{Standard MDP}
				\label{fig:SCMa}
			\end{subfigure}
			\hfill
			\begin{subfigure}[b]{0.45\textwidth}
				\centering
				\includegraphics[width=\textwidth]{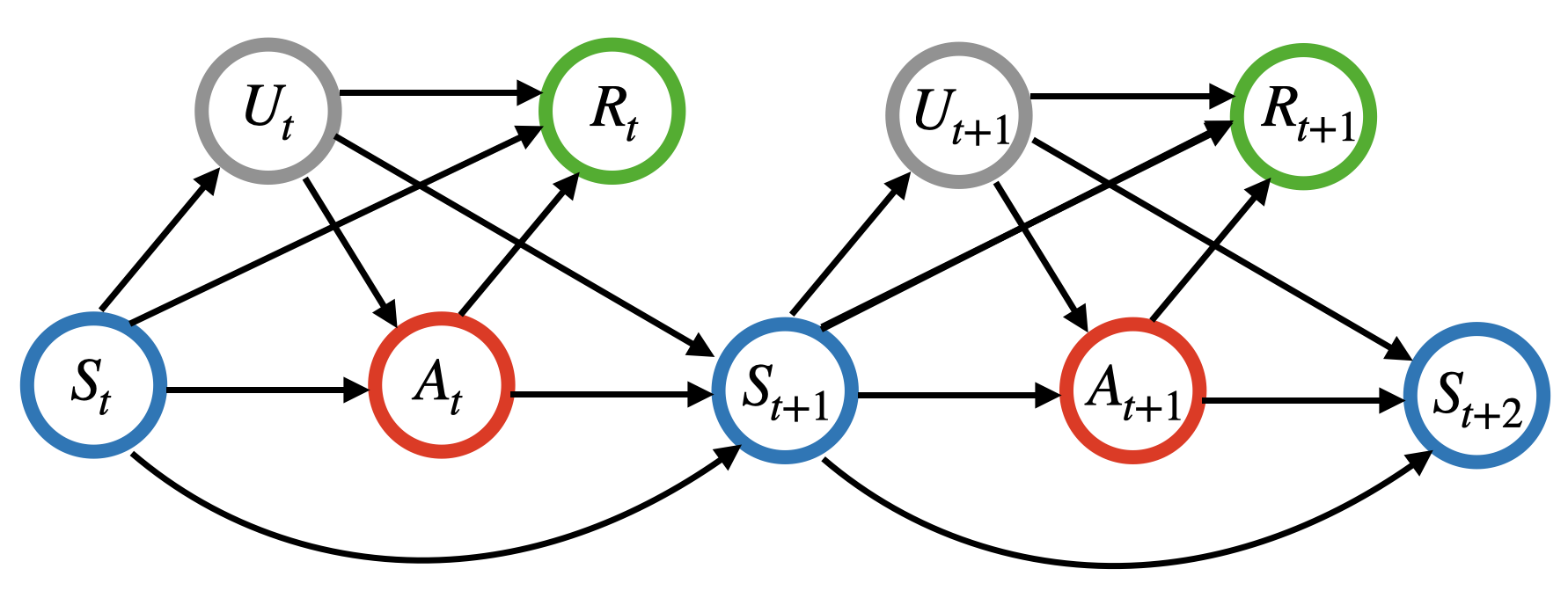}
				\caption{Confounded MDP}
				\label{fig:SCMb}
			\end{subfigure}
			\hfill
			\begin{subfigure}[b]{0.6\textwidth}
				\centering
				\includegraphics[width=\textwidth]{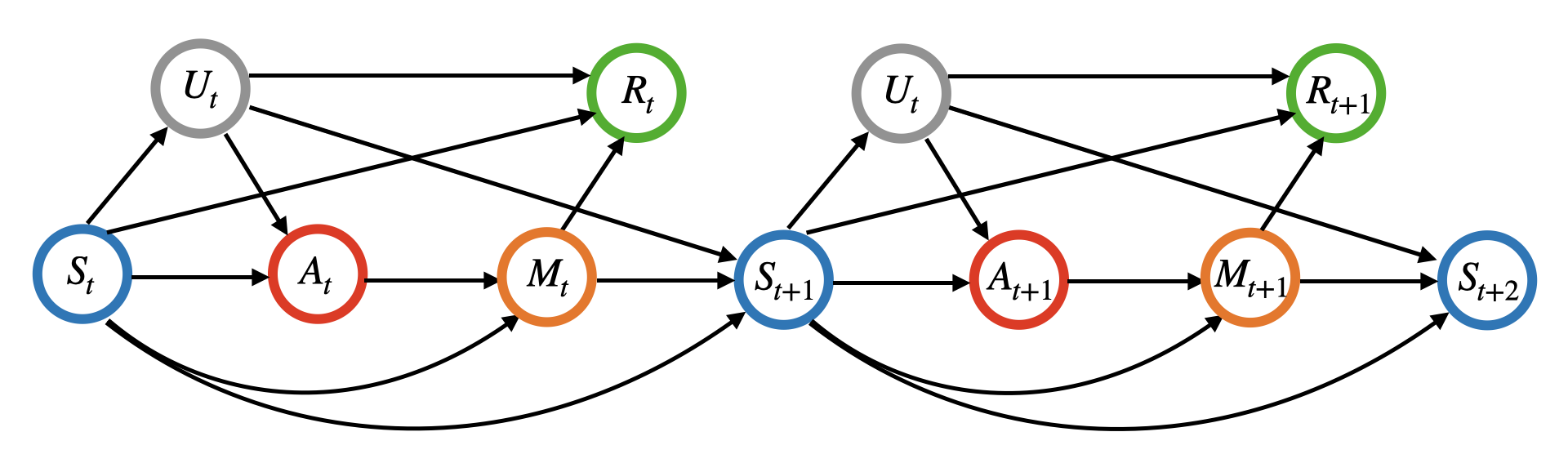}
				\caption{CMDPWM}
				\label{fig:SCMc}
			\end{subfigure}
			\caption{Causal diagrams}
			\label{fig:SCM} 
		\end{figure}

		We consider observational data generated from an confounded Markov decision process. Specifically, at a given time $t$, let $(S_t,A_t,R_t)$ denote the observed state-action-reward triplet. A standard MDP without confounding is depicted in Figure \ref{fig:SCMa}. We assume both the state and action spaces are discrete, and the immediate rewards are uniformly bounded. {\color{black}The discrete state-space assumption is imposed only to simplify the theoretical analysis. Our proposal is equally applicable to continuous state space as well.}
		Let $U_t$ denote the set of unmeasured  variables at time $t$ that confounds either the $A_t$-$R_t$ or $A_t$-$S_{t+1}$ associations, as shown in Figure \ref{fig:SCMb}. Such a data generating process excludes the existence of confounders that are influenced by past actions, leading to ``memoryless unmeasured confounding'' \citep{kallus2020confounding}. {It yields the following Markov assumption:
			
			\noindent \textbf{Assumption 1.}
			$U_t$ and other observed variables at time $t$ are conditionally independent of $\{U_j\}_{j<t}$ and past observed variables up to time $t-1$ given $S_t$. 
		}
		%

		To deal with unmeasured confounders, we assume there exist some observed immediate variables $M_t$ that mediate the effect of $A_t$ on $R_t$ and $S_{t+1}$  at time $t$, as shown in Figure \ref{fig:SCMc}. See Assumption 2 below. This assumption is similar to the front-door adjustment criterion  \citep[]{pearl2009causality} in  single-stage decision making {and is considered by \cite{wang2020provably} as well for multi-stage decision making.} 
		
		\noindent \textbf{Assumption 2.}
		(a) $M_{t}$ intercepts every directed path from $A_{t}$ to $R_t$ or to $S_{t+1}$; \\
		(b) $S_t$ blocks all backdoor paths from $A_t$ to $M_t$;\\
		(c) All back-door paths from  $M_{t}$ to $R_t$ or $S_{t+1}$ are blocked by $(S_t,A_t)$.

	For any two nodes $X$ and $Y$, a backdoor path from $X$ to $Y$ is a path that would remain if we were to remove any arrows pointing out of $X$.
	We revisit Figure \ref{fig:SCMc} to elaborate Assumption 2. Specifically, Assumption 2(a) requires the pathway that $A_t$ has a direct effect on $S_{t+1}$ absent $M_t$ to be missing. 
	Without Assumption 2(a), we can only identify the natural indirect treatment effect \citep{fulcher2020robust}  and  the policy value is not identifiable. 
	Under Assumptions 2(b) and 2(c), $U_t$ will not directly affect $M_t$. 
	Assumption 2(b) essentially requires that there are no unmeasured variables that confound the $A_t$-$M_t$ association. 
	Similarly, Assumption 2(c) requires  that there are no unmeasured variables that confound the $M_t$-$S_{t+1}$ association. 

	We next detail the data generating process. 
	At time $t$, we observe the state vector $S_t$ and the environment randomly selects some unmeasured confounder $U_t\sim p_u(\bullet|S_t)$. Then the agent takes the action $A_t\sim p_a(\bullet|S_t,U_t)$ and the mediator $M_t$ is generated using $p_m(\bullet|A_t,S_t)$ which is not confounded by $U_t$ according to Assumption 2. Finally, the agent receives a reward $R_t\sim p_r(\bullet|M_t,A_t,S_t,U_t)$ and the environment transits into the next state $S_{t+1}\sim p_s(\bullet|M_t,A_t,S_t,U_t)$. We refer to such a stochastic process as the confounded MDP with mediators, or CMDPWM for short. 
	\subsection{Problem Formulation}
	The data consist of $N$ trajectories, summarized as $\{(S_{i,t},A_{i,t},M_{i,t},R_{i,t},S_{i,t+1})\}_{1\le i\le N, 0\le t<T_i}$ where $T_i$ corresponds to the termination time of the $i$th trajectory. We assume these trajectories are i.i.d. copies of a CMDPWM model $\{(S_t,A_t,M_t,R_t,S_{t+1})\}_{t\ge 0}$. 
	
	Let $\pi$ denote a given stationary policy that maps the state space to a probability mass function on the action space $\mathcal{A}$. Following $\pi$, at each time $t$, the decision maker will set $A_t=a$ with probability $\pi(a|S_t)$ for any $a\in \mathcal{A}$. Unlike the behavior policy $p_a$, the probability mass function $\pi$ does not depend on the unmeasured confounders. For a given discounted factor $0\le \gamma<1$, we define the corresponding (state) value function as 
	\begin{eqnarray}\label{eqn:value}
		V^{\pi}(s)=\sum_{t=0}^{+\infty} \gamma^t \Mean^{\pi}(R_t|S_0=s), 
	\end{eqnarray}
	where the expectation $\mathbb{E}^{\pi}$ is defined by assuming the system follows the policy $\pi$. Based on the observed data, our objective is to learn the aggregated value $\eta^{\pi}=\Mean \{V^{\pi}(S_0)\}$ where the expectation is taken with respect to the initial state distribution, and to construct its
	associated confidence interval.
	
	{\color{black}We remark that we adopt a discounted reward formulation to investigate the policy evaluation problem.  This formulation allows us to take customers' frequency of using the app into consideration in our application (see Section \ref{sec:real data} for details). Meanwhile, our proposal can be easily extended to the average reward setting (see Appendix \ref{sec:appaveragereward}).} 
	
	\section{Off-Policy Confidence Interval Estimation}\label{sec:causalOPE}
	We first discuss the challenge of OPE in the presence of unmeasured confounders. We next show that $\eta^{\pi}$ can be represented as a function of the observed dataset. This result implies that $\eta^{\pi}$ is identifiable and forms the basis of our proposal. We then outline two potential estimators for $\eta^{\pi}$. Each estimator suffers from some limitations and requires some parts of the model to be correctly specified. This motivates our procedure that combines both estimators for more robust and efficient off-policy evaluation, based upon which a Wald-type CI is derived. 
	Finally, we detail our method. 
	
	\subsection{The Challenge with Unmeasured Confounders}
	In this section, we discuss the challenge of OPE with unmeasured confounders. {To simplify the presentation, we assume $\pi$ is a deterministic policy such that $\pi(\bullet|s)$ is a degenerate distribution for any $s$ throughout this section and Section \ref{sec:idenetapi}}. For any such policy, we use $\pi(s)$ to denote the action that the agent selects after observing the state vector $s$. 
	{To begin with, we introduce the do-operator $do$ to represent a (hard) intervention \citep[see e.g.,][]{pearl2009causality}. It amounts to lift $A_t$ from the influence of the old functional mechanism $A_t\sim p_a(\bullet|S_t,U_t)$ and place it under the influence of a new mechanism that sets the value $A_t$ while keeping all other mechanisms unperturbed. For instance, the notation $do(A_t=\pi(S_t))$ means that the action $A_t$ is set to the value $\pi(S_t)$ irrespective of the value of $U_t$. In other words, whatever relationship exists between $U_t$ and $A_t$, that relationship is no longer in effect when we perform the intervention.} Adopting the do-operator, the expectation $\Mean^{\pi}$ in \eqref{eqn:value} can be represented as
	\begin{eqnarray}\label{eqn:dooperator}
		\Mean \{R_t|do(A_j=\pi(S_j)),\forall 0\le j\le t,S_0=s\}.
	\end{eqnarray}
	In the presence of unmeasured confounders, the major challenge lies in that $\eta^{\pi}$ is defined based on the intervention distribution under the do-operator and cannot be easily approximated via the distribution of the observed data. 
	To elaborate this, we remark that the expectation in \eqref{eqn:dooperator} is generally not equal to $	\Mean \{R_t|A_j=\pi(S_j),\forall 0\le j\le t,S_0=s\}.$
	This is because the distribution under $do(A_t=\pi(S_t))$ is different from that given the observation $A_t=\pi(S_t)$. The latter corresponds to the conditional distribution generated by the causal diagram in Figure \ref{fig:SCM} given $A_t=\pi(S_t)$, whereas the former is the distribution generated by a slightly different graph, with the pathway $U_t\to A_t$ removed. 
	
\begin{figure}[t]
		\centering
		\includegraphics[width=0.6\linewidth]{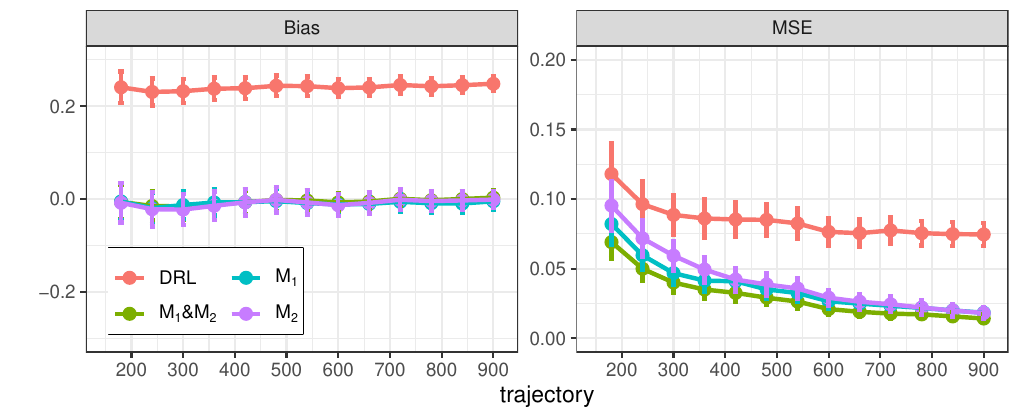}
		\caption{\small Bias and mean squared error (MSE) of DRL and the proposed estimator under different settings. $T=100$ and the results are aggregated over 200 simulations. The error bar corresponds to 95\% confidence interval for the bias and MSE, from left to right. The proposed estimator requires specification of two sets of models, $\mathcal{M}_1$ and $\mathcal{M}_2$ (see Section \ref{sec:toy} for details). The green line depicts the estimator where both sets of models are correctly specified. 
			The blue line depicts the estimator where models in $\mathcal{M}_1$ are correctly specified and $\mathcal{M}_2$ misspecified. 
			The purple line depicts the estimator where $\mathcal{M}_2$ is correctly specified and $\mathcal{M}_1$ is misspecified.
		}\label{fig:toy} 
	\end{figure}

	As an illustration, we apply DRL and the proposed method to a toy example detailed in Section \ref{sec:toy}. 
	The data are generated according to a CMDPWM model. 
	As we have commented, DRL is proposed by assuming no unmeasured confounders exist. As such, it can be seen from the left panel of Figure \ref{fig:toy} that the DRL estimator has a non-diminishing bias under this example, due to the presence of unmeasured confounders. As shown in the right panel of Figure \ref{fig:toy}, the mean squared error (MSE) of DRL does not decay to zero as the number of trajectories increases to infinity. 
	
	In the next section, we address the above mentioned challenge by making use of the auxiliary variables $M_t$ in the observed data. It can be seen from Figure \ref{fig:toy} that the proposed estimator is consistent. Both its bias and MSE decay to zero as the number of trajectories diverges to infinity. {Finally, we remark that in addition to the use of do-operator, one can adopt the potential outcome framework to formulate the policy evaluation problem \citep[see e.g.,][]{fulcher2020robust}. We omit the details to save space.}
	
	\subsection{Identification of $\eta^{\pi}$}\label{sec:idenetapi}
	In this section, we first show that $\eta^{\pi}$ is identifiable based on the observed data. The main idea is to iteratively apply the Markov property and the front-door adjustment formula to represent the intervention distribution under the do-operator via the observed data distribution.  
	
	Recall that $p_a$ is the conditional distribution of $A_t|(S_t,U_t)$. We use $p_a^*$ to denote the corresponding conditional distribution $A_t|S_t$, marginalized over $U_t$. Similarly, we use $p_{s,r}^*$ to denote the conditional distribution $(S_{t+1},R_t)|(M_t,A_t,S_t)$. We summarize the results in the following theorem. 
	
	\begin{thm}\label{thm:identifiable} Let $\tau_t$ denote the data history  $\{(s_j,a_j,m_j,r_j)\}_{0\le j\le t}$ up to time $t$ and $\nu$ denote the initial state distribution. Under Assumptions 1 and 2, $\eta^{\pi}$ is equal to
		\begin{eqnarray*}
			\begin{aligned}
				\left[ \sum_{t=0}^{+\infty}\gamma^t \sum_{\tau_t,s_{t+1}} r_t \left\{\prod_{j=0}^t p_{s,r}^*(s_{j+1},r_j|m_j,a_j,s_j)  p_m(m_j|\pi(s_j),s_j)p_a^*(a_j|s_j) \right\}\nu(s_0)\right].
			\end{aligned}
		\end{eqnarray*}
	\end{thm}
	Note that none of the distributions $p_{s,r}^*$, $p_m$ and $p_a^*$ involves the unmeasured confounders. As such, these distribution functions can be consistently estimated based on the observational data. Consequently, Theorem \ref{thm:identifiable} implies that $\eta^{\pi}$ can be rewritten using the observed data distribution. {Assumption 1 ensures the process satisfies the Markov property. Together with Assumption 2, it allows us to iteratively apply the front-door adjustment formula to replace the intervention distribution with the observed data distribution. See the proof of Theorem \ref{thm:identifiable} in Appendix \ref{sec:proofthm1} for details.} We next outline two potential estimators for $\eta^{\pi}$. 
	
	\subsection{Direct Estimator}\label{sec:de}
	{The first estimator is  Direct Estimator, where we estimate the Q-function based on the observed data and directly use it to derive the value estimator.} In our setting, we define the Q-function $Q^{\pi}(m,a,s)=
	\Mean \{R_t+\gamma V^{\pi}(S_{t+1})|M_t=m,A_t=a,S_t=s\}$.
	We make a few remarks. First, 
	our definition of the Q-function is slightly different from that in the existing RL literature, defined by $\Mean \{R_t+\gamma V^{\pi}(S_{t+1})|A_t=a,S_t=s\}$, as it involves mediators. Second, similar to Theorem \ref{thm:identifiable}, we can show $V^{\pi}$ is identifiable from the observed data. It follows that $Q^{\pi}$ is identifiable as well. 
	
	To motivate the first estimator, we notice that $\eta^{\pi}$ can be rewritten as $\Mean^{\pi} \{Q^{\pi}(M_0,A_0,S_0)\}$, or equivalently, $\sum_a\Mean [\pi(a|S_0)\Mean\{ Q^{\pi}(M_0,a,S_0)|do(A_0=a),S_0\}]$.  Applying the front-door adjustment formula, we obtain that
	\begin{eqnarray}\label{eqn:Qvalues}
		\eta^{\pi}=\sum_{m,a,a',s} p_m(m|a',s)\pi(a'|s) p_a^*(a|s) Q^{\pi}(m,a,s)\nu(s).
	\end{eqnarray}
	This motivates us to learn $p_m$, $p_a^*$, $Q^{\pi}$ and $\nu$ from the observed data and construct the value estimate by plugging-in these estimators. We refer to this estimator as the direct estimator, since the procedure shares similar spirits as the direct method in the RL literature.
	
	\subsection{Importance Sampling Estimator}\label{sec:ise}
	The second estimator is Importance Sampling (IS) Estimator. {This is motivated by the work of \cite{liu2018breaking} that develops a marginal IS estimator that breaks the curse of horizon, assuming no unmeasured confounders exist. Compared to the standard IS estimator \citep{zhang2013robust} whose variance will grow exponentially fast with respect to the number of decision points, the marginal IS estimator takes the stationary property of the state transitions into consideration and effectively breaks the curse of high variance in sequential decision making.} Specifically, let $\omega^{\pi}(\bullet)$ be the marginal density ratio, 
	\begin{eqnarray*}
		(1-\gamma) \sum_{t\ge 0}  \gamma^t \frac{p_t^{\pi}(s)}{p_{\infty}(s)},
	\end{eqnarray*}
	where $p_t^{\pi}(s)$ denotes the probability of $S_t=s$ by assuming the system follows $\pi$, and $p_{\infty}$ denotes the limiting distribution of the stochastic process $\{S_t\}_{t\ge 0}$. Similar to Theorem \ref{thm:identifiable}, we can show for any $t>1$, $p_t^{\pi}$ is identifiable. So is $\omega^{\pi}$. 
	
	A key observation is that, when the stochastic process $\{S_t\}_{t\ge 0}$ is stationary, it follows from the change of measure theorem that $\eta^{\pi}=(1-\gamma)^{-1} \sum_a \Mean \{\pi(a|S_t) R_t\omega^{\pi}(S_t)|do(A_t=a)\}$. When no unmeasured confounders exist, we have $\eta^{\pi}=(1-\gamma)^{-1} \Mean \{\pi(A_t|S_t)R_t\omega^{\pi}(S_t)/p_a^*(A_t,S_t)\}$, yielding the marginal IS estimator. 
	To replace the intervention distribution with the observed data distribution, we apply the importance sampling method again and re-weight each reward by another probability ratio
	\begin{eqnarray}\label{eqn:rho}
		\rho(M_t,A_t,S_t)=\frac{\sum_a \pi(a|S_t) p_m(M_t|a,S_t)}{p_m(M_t|A_t,S_t)}. 
	\end{eqnarray} 
	Such an importance sampling trick has been used by \citet{fulcher2020robust} to handle
	unmeasured confounders in single-stage decision making. This yields the following value
	estimate, 
	\begin{eqnarray*}
		\frac{1}{(1-\gamma)\sum_i T_i}\sum_{i,t}R_{i,t}\widehat{\omega}(S_{i,t})\frac{\sum_a \pi(a|S_t) \widehat{p}_m(M_t|a,S_t)}{\widehat{p}_m(M_t|A_t,S_t)},
	\end{eqnarray*}
	where $\widehat{\omega}$ and $\widehat{p}_m$ denote some estimators for $\omega^{\pi}$ and $p_m$. 
	
	To conclude this section, we discuss the limitations of the two estimators. First, each estimator requires some parts of the model to be correctly specified. Specifically, the direct estimator requires consistent estimates for $Q^{\pi}$, $p_m$ and $p_a^*$, and IS requires correct specification of $\omega^{\pi}$ and $p_m$. Second, generally speaking, the direct estimator suffers from a large bias due to potential model misspecification whereas the IS estimator suffers from a large variance due to inverse probability weighting. 
	To address both limitations simultaneously, we develop a robust and efficient OPE procedure by carefully combining the two estimating strategies used in Sections \ref{sec:de} and \ref{sec:ise}.
	Meanwhile, the resulting estimator requires weaker assumptions to achieve consistency. We present the main idea in the next section. 
	
	\subsection{Our Proposal}\label{sec:ourproposal}
	We begin with some notations. Let $O$ be a shorthand for a data tuple $(S,A,M,R,S')$. 
	The key to our estimator is the estimating function, 
	$\psi(O)=\psi_0+\sum_{j=1}^3 \psi_j(O)$, 
	where $\psi_0$ is the direct estimator outlined in 
	\eqref{eqn:Qvalues}, and $\psi_1(O),\psi_2(O),\psi_3(O)$ are some augmentation terms detailed below. Recall that $\psi_0$ depends on $Q^{\pi}$, $p_m$ and $p_a^*$. The purpose of adding the three augmentation terms is to offer additional protection against potential model misspecification of these nuisance functions. As such, the proposed estimator achieves the desired robustness property. See Figure \ref{fig:toy} for an illustration. {A pseudocode summarizing the proposed algorithm is given in Algorithm \ref{alg:full}.} 
	
	We next present the explicit forms of the three augmentation terms. Specifically, $\psi_1(O)$ equals 
	\begin{eqnarray*}
		\frac{1}{1-\gamma}\omega^{\pi}(S)\rho(M,A,S)\Big\{R+\gamma \sum_{m,a,a^*}Q^{\pi}(m,a,S')p_m(m,a^*,S')p_a^*(a|S')\pi(a^*|S') 
		-Q^{\pi}(M,A,S)\Big\},
	\end{eqnarray*} 
	where $\rho$ is the probability ratio defined in \eqref{eqn:rho}. 
	The last term in the curly bracket corresponds to the temporal difference error under the CMDPWM model whose conditional mean given $(M,A,S)$ equals zero. As such, $\psi_1(O)$ has zero mean when $Q^{\pi}$, $p_m$ and $p_a^*$ are correctly specified.  
	
	$\psi_2(O)$ equals 
	\begin{eqnarray*}
		(1-\gamma)^{-1}\omega^{\pi}(S)\frac{\pi(A|S)}{p_a^*(A|S)} \sum_a p_a^*(a|S) \Big\{Q^\pi(M,a,S)-\sum_m p_m(m|A,S) Q^\pi(m,a,S)\Big\}.
	\end{eqnarray*}
	When $p_m$ is correctly specified, the last term in the curly bracket can be represented as the residual $Q^{\pi}(M,a,S)-\Mean \{Q^{\pi}(M,a,S)|A,S\}$. As such, $\psi_2(O)$ has zero mean when $p_m$ is correctly specified. 
	
	$\psi_3(O)$ equals
	\begin{eqnarray*}
		(1-\gamma)^{-1}\sum_{m,a'} \omega^{\pi}(S) p_m(m|a',S)\pi(a'|S) \Big\{ Q^\pi(m,A,S)-\sum_{a} Q^\pi(m,a,S)p_a^*(a|S) \Big\}.
	\end{eqnarray*}
	Similarly, the last term in the curly bracket can be represented as the residual $Q^{\pi}(m,A,S)-\Mean \{Q^{\pi}(m,A,S)|S\}$. When $p_a^*$ is correctly specified, we have $\Mean \psi_3(O)=0$. 
	
	Based on the estimating function, the proposed estimator takes the following formula,
	\begin{eqnarray}\label{eqn:valueest}
		\widehat{\eta}=\frac{1}{\sum_i T_i}\sum_{i=1}^N\sum_{t=0}^{T_i-1}\psi(O_{i,t}),
	\end{eqnarray}
	where $O_{i,t}=(S_{i,t},A_{i,t},M_{i,t},R_{i,t},S_{i,t+1})$. {Compared to the standard DRL estimator, the proposed estimator involves additional computations due to the inclusion of the mediator distribution function in the latter two augmentation terms.} When there are no unmeasured confounders, the proposed estimator shares similar spirits with DRL.
	
	To construct such an estimator, we need to learn $Q^{\pi}$, $\omega^{\pi}$, $p_a^*$, $p_m$ and the initial state distribution $\nu$.  Note that
	estimating $p_a^*$ or $p_m$ is essentially a regression problem. These functions can be conveniently estimated via existing supervised learning algorithms. 
	We estimate $\nu$ via the empirical distribution of $\{S_{i,0}\}_{1\le i\le N}$.
	As for $Q^{\pi}$ and $\omega^{\pi}$, we discuss the corresponding estimating procedure later in Section \ref{sec:Qomega}. 
	
	Next, we discuss the relationship between the proposed estimator in \eqref{eqn:valueest} and the two estimators outlined in Sections \ref{sec:de} and \ref{sec:ise}. 
	Suppose $p_m$ is correctly specified. Let $\mathcal{M}_1$ denote the set of models $\{Q^{\pi}, p_a^*\}$, and $\mathcal{M}_2$ denote the model $\omega^{\pi}$. 
	First, when the models in $\mathcal{M}_1$ are correctly specified, the three augmentations terms have zero mean, as we have discussed earlier. By the weak law of large numbers, \eqref{eqn:valueest} is asymptotically equivalent to the direct estimator and is thus consistent. Second, when the models in $\mathcal{M}_2$ are correctly specified, we have $\Mean \psi_2(O)=0$. In addition, using similar arguments in Part 3 of the proof of Theorem \ref{thm2} in the appendix, we can show that
	\begin{eqnarray*}
		\psi_0+\Mean \psi_3(O)=\frac{1}{1-\gamma} \Mean \left[\omega^{\pi}(S)\rho(M,A,S) \{Q^{\pi}(M,A,S)-\gamma V^{\pi}(S')\}\right].
	\end{eqnarray*}
	By the definition of $\psi_1$, this in turn implies that \eqref{eqn:valueest} is unbiased to the IS estimator. It is thus consistent. 
	The above discussion informally justifies the robustness property of \eqref{eqn:valueest}. We will rigorously prove the claim  in Theorem \ref{thm2}.
	
	Finally, 
	we observe that the proposed estimator can be written as $N^{-1} \sum_{i=1}^N \eta_i$ where $\eta_i$ denotes the estimating function based on the $i$th trajectory only. Since the trajectories are independent, the proposed estimator is asymptotically normal, as shown in Theorem \ref{thm3}. A Wald-type CI
	\begin{eqnarray}\label{eqn:CI}
		[\widehat{\eta}-z_{\alpha/2} N^{-1/2}\widehat{\sigma}_{\eta}, \widehat{\eta}+ z_{\alpha/2}N^{-1/2}\widehat{\sigma}_{\eta}],
	\end{eqnarray} 
	is valid for off-policy interval estimation, where $z_{\alpha}$ denotes the upper $\alpha$th quantile of a standard normal distribution and $\widehat{\sigma}_{\eta}^2$ denotes the sampling variance estimator of $\{\eta_i\}_i$.

	\begin{algorithm}[!t]
		\caption{Proposed procedure for confounded off-policy confidence interval estimation.}
		\label{alg:full}
		\begin{algorithmic}[1]
			{\REQUIRE The data $\{(S_{i,t},A_{i,t},M_{i,t},R_{i,t})\}_{i,t}$, and the significance level $0<\alpha<1$. 
				\STATE Compute the estimators for $p_a^*$ and $p_m$ via supervised learning algorithms. Estimate $\nu$ via the empirical initial state distribution. 
				\STATE Compute the Q-function and marginal density ratio estimator according to Section \ref{sec:Qomega}. 
				\STATE Plug-in the aforementioned estimated nuisance functions into \eqref{eqn:valueest} to construct the value estimator $\widehat{\eta}$.
				\STATE Construct the Wald-type CI according to \eqref{eqn:CI}. }
		\end{algorithmic}
					
				
	\end{algorithm}
	
	\subsection{Learning $Q^{\pi}$ and $\omega^{\pi}$}\label{sec:Qomega}
	The estimating procedure for $Q^{\pi}$ is motivated by the following Bellman equation,
	\begin{eqnarray*}
		\Mean \Big\{R+\gamma \sum_{m,a,a^*}Q^{\pi}(m,a,S')p_m(m,a^*,S')p_a^*(a|S')\left.\pi(a^*|S')\right|M,A,S\Big\}=Q^{\pi}(M,A,S).
	\end{eqnarray*}
	Similar to the standard Bellman equation under settings without unmeasured confounders, it decomposes the Q-function into two parts, the immediate reward plus the discounted future state-action values. 
	
	To prove this identity, notice that similar to \eqref{eqn:Qvalues}, we can show that 
	\begin{eqnarray*}
		V^{\pi}(s)=\sum_{m,a,a^*}Q^{\pi}(m,a,s)p_m(m,a^*,s)p_a^*(a|s)\pi(a^*|s),
	\end{eqnarray*}
	based on the front-door adjustment formula. This together with our definition of the Q-function yields the above Bellman equation. 
	
	Let $\widehat{p}_m$ and $\widehat{p}_a^*$ denote consistent estimators for $p_m$ and $p_a$, based on the observed data. Given the Bellman equation, multiple methods can be applied to estimate $Q^{\pi}$. We employ the fitted Q-evaluation method \cite{le2019batch} in our setup and propose to iteratively compute $\widehat{Q}^{{\ell}+1}$ by solving
	\begin{eqnarray*}
		\begin{split}
			\argmin_{Q\in \mathcal{Q}}	\sum_{i,t}
			\left\{
			R_{i,t}- Q(M_{i,t}, A_{i, t}, S_{i, t})+\gamma 
			\widehat{V}^{\ell}(S_{i, t+1})  \right\}^2, 
		\end{split}
	\end{eqnarray*}
	where  $\widehat{V}^{\ell}(S_{i, t+1}) = \sum_{m,a,a^*} \widehat{Q}^{\ell}(m,a,S_{i,t+1})\widehat{p}_m(m,a^*,S_{i,t+1})\widehat{p}_a^*(a|S_{i,t+1})\pi(a^*|S_{i,t+1}),$
	for some function class $\mathcal{Q}$ and $\ell=0,1,\cdots$, until convergence. Similar to \cite{fan2020theoretical}, we can show that the resulting Q-estimator is consistent when $\mathcal{Q}$ is a class of universal function approximators such as neural networks.
	
	We next consider $\omega^{\pi}$. Similar to the work of \cite{liu2018breaking}, 
	we can show that when the process $\{S_t\}_{t\ge 0}$ is stationary,  $\omega^{\pi}$ satisfies the equation $L(\omega^{\pi},f)=0$ for any discriminator function $f$ in our setup, where $L(\omega^{\pi},f)$ is given by
	\begin{eqnarray}\label{eqn:learnomega}
		\begin{split}
			\Mean \omega^{\pi}(S_{i,t})\left\{f(S_{i,t})-\gamma\frac{\sum_{a\sim \pi(\bullet|S_{i,t})} p_m(M_{i,t}|a,S_{i,t})}{p_m(M_{i,t}|A_{i,t},S_{i,t})}f(S_{i,t+1}) \right\}-(1-\gamma)\sum_{s}f(s)\nu(s).
		\end{split}
	\end{eqnarray}
	As such, $\omega^{\pi}$ can be learned by solving the following mini-max problem, 
	\begin{eqnarray}\label{eqn:solveL}
		\argmin_{\omega\in \Omega} \sup_{f\in \mathcal{F}} L^2(\omega, f), 
	\end{eqnarray}
	for some function classes $\Omega$ and $\mathcal{F}$. 
	The expectation in \eqref{eqn:learnomega} can be approximated by the sample average. $p_m$ and $\nu$ in \eqref{eqn:learnomega} can be substituted with their estimators. {As pointed out by one of the referees, the minimax optimization is often not stable. To address this issue, we restrict attention to linear or kernel function classes to simply the calculation. See Appendix \ref{sec:moreexp} for details.}
	
	\section{Statistical Guarantees}\label{SecTheory}
	We prove the robustness and efficiency of our estimator as well as the validity of our CI in this section. Without loss of generality, we assume $T_i=T$ for any $i$. To derive the asymptotic theories, we require the number of trajectories $N$ to diverge to infinity. The termination time $T$ can either be bounded, or diverge with $N$. The assumption on $N$ is imposed to ensure that the initial state distribution $\nu$ can be well-approximated by the empirical distribution of $\{S_{i,0}\}_{1\le i\le N}$. We first introduce some conditions. Let $\mathcal{H}_m$ and $\mathcal{H}_a$ be the function classes used to model $p_m$ and $p_a$, respectively. 
	
	
	\noindent \textbf{Assumption 2.} The function classes $\mathcal{Q}$, $\Omega$, $\mathcal{H}_m$ and $\mathcal{H}_a$ are bounded and belong to VC type classes \citep[Definition 2.1,][]{chernozhukov2014gaussian} with VC indices upper bounded by $v=O(N^{\kappa})$ for some $0\le \kappa<1/2$. 
	
	Assumption 2 is mild as the function classes are user-specified. VC type classes contains a wide variety of functional classes, including neural networks and regression trees. The VC index controls the model complexity. It generally increases with the number of parameters in the model. We allow the VC index to diverge with the sample size to reduce the bias of the estimator due to model misspecification. 
	
	
	\begin{thm}[Robustness]\label{thm2}
		Suppose the process $\{S_t\}_{t\ge 0}$ is stationary, $p_m(M_{i,t}|A_{i,t},S_{i,t})$, $p_a(A_{i,t}|S_{i,t})$, $\widehat{p}_m(M_{i,t}|A_{i,t},S_{i,t})$, $\widehat{p}_a(A_{i,t}|S_{i,t})$ and $p_{\infty}(A_{i,t}|S_{i,t})$ are uniformly bounded away from zero, Assumptions 1, 2 hold, and $\widehat{p}_m$ is consistent. Then as $N\to\infty$, the proposed estimator is consistent when either $\widehat{Q}$, $\widehat{p}_a^*$ or $\widehat{\omega}$ converges in $L_2$-norm to their oracle values. 
	\end{thm}
	To save space, we present the detailed definition of $L_2$-norm convergence in  Appendix \ref{sec:l2conv}. 
	Theorem \ref{thm2} formally establishes the robustness property. Notice that the proposed estimator equals the direct estimator outlined in Section \ref{sec:de} when $\omega^{\pi}=0$ and equals the IS estimator in Section \ref{sec:ise} when $\widehat{Q}=0$. As a byproduct, we obtain the following corollary.
	
	\begin{coro}
		(i) Suppose the conditions in Theorem \ref{thm2} hold. Suppose $\widehat{Q}$ and $\widehat{p}_a^*$ converge in $L_2$-norm to their oracle values. Then the direct estimator is consistent as $N\to\infty$. (ii) Suppose $\widehat{\omega}$ converges in $L_2$-norm to their oracle value. Then the IS estimator is consistent as $N\to\infty$.
	\end{coro}

	To achieve efficiency, we need the following assumption: 
	
	\noindent \textbf{Assumption 3.} Suppose $\widehat{Q}$, $\widehat{p}_a^*$, $\widehat{p}_m$, $\widehat{\omega}$ converge in $L_2$-norm to their oracle values at a rate of $N^{-\kappa^*}$ for some $\kappa^*>1/4$. 
	
	Assumption 3 characterizes the theoretical requirements on the nuisance function estimators. Suppose some parametric models (e.g., linear) are imposed to learn these nuisance functions. When the models are correctly specified, then we have $\kappa^*=1/2$ \citep{uehara2021finite}. Here, we do not impose parametric assumptions and only require $\kappa^*>1/4$. For instance, when using kernels or neural networks for function approximation, the corresponding convergence rates of $\widehat{Q}$ and $\widehat{\omega}$ are provided in \cite{fan2020theoretical,liao2020batch}. $\widehat{p}_a^*$ and $\widehat{p}_m$ can be computed via standard supervised learning algorithms. Their rates of convergence are available for most often used machine learning approaches including random forests \citep{wager2018estimation} and deep learning \citep{schmidt2020nonparametric}. 
	
	\begin{thm}[Efficiency]\label{thm3}
		Suppose the conditions in Theorem \ref{thm2} hold and Assumption 3 holds. Then 
		the proposed estimator achieves the semiparametric efficiency bound.
	\end{thm}
	
	We make a few remarks. First, we show in the proof of Theorem \ref{thm3} that the proposed estimator is asymptotically normal and satisfies $\sqrt{N}(\widehat{\eta}-\eta^{\pi})\stackrel{d}{\to} N(0,\sigma_T^2)$ where the explicit form of $\sigma_T^2$ is detailed in Appendix \ref{sec:proofthm3}. The asymptotic variance estimator for $\sigma_T^2$ can be constructed via the sampling-variance formula. Consequently, a two-sided Wald-type confidence interval (CI) can be derived for $\eta^{\pi}$. 
	Second, the asymptotic variance $\sigma_T^2$ decays with $T$. Specifically, it can be decomposed into $\sigma_0^2+T^{-1} \sigma_*^2$ for some $\sigma_0,\sigma_*$. See Appendix \ref{sec:proofthm3} for the explicit forms of these quantities. 
	The first term $\sigma_0^2$ accounts for the variation of the initial state distribution in the plug-in estimator. The second term $T^{-1} \sigma_*^2$ is the variance of the augmentation terms and decays to zero as $T\to \infty$. Third, \cite{kallus2019efficiently} derives the efficiency bound for OPE in infinite horizon settings where no unmeasured confounders exist and the initial state distribution under the target policy is known. Our proof for Theorem \ref{thm3} differs from theirs in that we allow the initial state distribution to be unknown and allow unmeasured confounders to exist. {Fourth, in Assumption 3, we require all the nuisance function estimators to converge to their oracle values and thus exclude the case with model misspecification. When the model is misspecified, the semiparametric efficiency bound cannot be achieved. Finally, in our proposal, 
		we use the same dataset twice to estimate the nuisance functions and construct the final value estimator. We do not utilize cross-fitting. Because of that, we impose certain metric entropy conditions in Assumption 2 to establish the robustness and efficiency of the proposed value estimator. To remove Assumption 2, we can couple our procedure with sample-splitting and cross-fitting \citep[see e.g.,][]{chernozhukov2018double,kallus2019efficiently}. However, in our setup, we find that the proposed estimator without cross-fitting has better finite sample properties. }

\begin{thm}[Validity]\label{thm4}
	Suppose the conditions in Theorem \ref{thm3} hold. Then the coverage probability of the proposed CI approaches to the nominal level as $N$ diverges to infinity.
\end{thm}

{We remark that Theorems \ref{thm3} and \ref{thm4} are concerned with the asymptotic distribution of the value estimator under a single target policy. In Appendix \ref{sec:moretarget} of the supplementary article, we establish the joint asymptotic distribution of the proposed value estimators under multiple target policies, introduce the proposed CI for the value difference between two target policies and prove its validity.}

\section{Simulation studies}\label{sec:simulation}
In this section, we evaluate the finite sample performance of the proposed estimator using two simulation studies. The first toy example aims to illustrate the robustness properties of our estimator to unmeasured confounding and model misspecification. In the second simulation study, we demonstrate that our method is superior to 
	state-of-the-art policy evaluation methods.

	\subsection{A toy example}\label{sec:toy}
	We first describe the detailed setting for the toy example. We fix time $T=100$ and 
	the initial state is sampled from a Bernoulli  distribution with support $\{0,1\}$ and satisfies that $\prob(S_0=1)=\prob(S_0=0)=0.5$. 
	The unmeasured confounders $\{U_t\}_{t=1}^{T}$ are i.i.d. sampled from 
	a Bernoulli distribution with support $\{-1,1\}$ and satisfy that $\prob(U_t=1)=\prob(U_t=-1)=0.5$. 
	The action is discrete-valued and the behaviour policy $p_a$ satisfies that 
	$p_a(1 | S_t, U_t) =p_a(-1| S_t, U_t) = 0.5 \textup{sigmoid}(0.1  S_t + 0.9  U_t)$, and 
	$p_a(0| S_t, U_t) = 1 - \textup{sigmoid}(0.1  S_t + 0.9  U_t)$. 
	The mediator is drawn from a Bernoulli distribution with binary support. 
	We set $p_m(1|A_t,S_t)=\textup{sigmoid}(0.1 S_t -0.9  (A_t - 0.5))$ which does not depend on $U_t$. Assumption 1 is thus satisfied.
	The reward $R_t$ and the next-state $S_{t+1}$ are Bernoulli random variables with support $\{0, 10\}$ and $\{0, 1\}$, respectively, 
	and satisfy $\prob( R_t = 10|S_t,U_t,M_t)=\prob( S_{t+1} = 1|S_t,U_t,M_t)=\textup{sigmoid}(0.5  I(U_t = 1)  (S_t + M_t) - 0.1 S_t)$. 
	We are interested in evaluating a random policy that outputs 0 
	with probability $1 - \textup{sigmoid}(0.3  S_t)$, and outputs -1 or 1 
	with probability $0.5 \textup{sigmoid}(0.3  S_t)$
	after observing $S_t$. 
	Under this toy example, we are able to derive $Q^{\pi},p_{m}$, $p^*_{a}$ and $\eta^{\pi}$ theoretically, and we calculate the true value of $\omega^{\pi}$ via Monte Carlo method.
	

	Recall that $\mathcal{M}_1$ is a combination of $Q^{\pi}$ , $p^*_{a}$ and $\mathcal{M}_2=\{\omega^{\pi}\}$. We evaluated the performance of the proposed estimator under the following scenarios:\
	(i) all the models $p_{m}$, $\mathcal{M}_1$, $\mathcal{M}_2$ are correct; 
	(ii) $p_{m}$ and  $\mathcal{M}_1$ are correct, $\mathcal{M}_2$ is misspecified;
	(iii) $p_{m}$ and  $\mathcal{M}_2$ are correct, $\mathcal{M}_1$ is misspecified.
	{Specifically, to misspecify $Q^{\pi}$, we inject a Gaussian noise with unit variance to the true $Q^{\pi}$.} To misspecify $p^*_{a}$, we multiply $p_a^*$ by a variable sampled from a uniform distribution with lower boundary 0.75 and higher boundary 1.
	To misspecify $\omega^{\pi}$, we increase the value of $\omega^{\pi}(0)$ by 0.5 and reduce the value of $\omega^{\pi}(1)$ by 0.5. 
	As shown in Figure~\ref{fig:toy}, our proposed estimator is robust to unmeasured confounding and model misspecification.
	



	\subsection{Comparison with state-of-the-art methods}\label{sec:compare}
	We compare the proposed method with the state-of-the-art methods in the existing reinforcement learning literature. The simulated data are generated as follows. 
	The initial state is sampled from a standard normal distribution with dimension $d_S = 1$ or $3$. 
	The distributions of unmeasured confounders are the same as those in the toy example. 
	Let $1_{t}$ be a length $t$ vector with values 1, 
	and $C_t = 1_{d_S}^\top S_t$, the sum of the state. The action is binary-valued and is generated according to the behaviour policy 
	$p_a(1|S_t,U_t)=\textup{sigmoid}(0.1 C_t + 0.9 U_t)$. 
	The mediator is drawn from a Bernoulli distribution with binary support. 
	We set $p_m(1|A_t,S_t)=\textup{sigmoid}(0.1 C_t + 0.9 (A_t - 0.5))$ which does not depend on $U_t$. Assumption 1 is thus satisfied.
	The reward $R_t$ is sampled from a normal distribution with 
	conditional mean $0.5 I(U_t = 1) (M_t + C_t) - 0.1 C_t$ and standard deviation 0.1. 
	The future state $S_{t+1}$ is sampled from a multivariate normal distribution with mean 
	$0.5 I(U_t = 1) (M_t 1_{d_S} + S_t) - 0.1 S_t$ and covariance matrix $0.25 I_{d_S}$. 
	The target policy selects action 1 with probability $\textup{sigmoid}(0.3 C_t)$. 
	
	We compare the proposed estimator with three types of baseline methods. All these methods are developed by assuming no unmeasured confounders. The first one is the direct estimator, computed based on an estimated Q-function, $\widehat{\eta}_{\textrm{REG}} = N^{-1}\sum_{i=1}^N \widehat{Q}(S_{i, 0}, \pi(S_{i, 0}))$ (denoted by REG).  
	In our implementation, we compute $\widehat{Q}$ via the fitted Q-evaluation algorithm. 
	The second one is the marginal importance sampling (MIS) estimator \citep{liu2018breaking}. 
	To implement this method, the marginal sampling ratio is estimated by assuming no unmeasured confounders exist and is different from the proposed estimator for $\omega^{\pi}$. The third one is the DRL estimator 
	that combines the first two estimators for value evaluation. None of these methods uses the mediator. For fair comparison, we also include the mediator in the state to construct the value estimates. Denote the resulting three estimators by REG-M, MIS-M and DRL-M, respectively. We further estimate their variances based on the sampling variance formula (see Appendix \ref{sec:moreexp} for details) and construct the associated confidence interval. 
	The proposed estimator is denoted by COPE, short for confounded off-policy interval estimation. 
	{The linear basis function models $\widehat{p}_a^*$, $\widehat{p}_m$, $\widehat{Q}^\pi$, $\widehat{\omega}^\pi$ employ randomly generated Fourier features based on the \proglang{Python} \code{RBFsampler} function. We find that the performance of the value estimator is not overly sensitive to the number of basis functions (see Appendix \ref{sec:moreexp} for details).} 
	Let $\eta^\pi$ be the ground truth and $\widehat{\eta}^\pi$ be a given OPE estimator, 
	we define logBias as $ \log_{10} (|\mathbb{E}\widehat{\eta}^\pi - \eta^\pi|)$ and logMSE as $ \log_{10}\{\mathbb{E}(\widehat{\eta}^\pi - \eta^\pi)^2\}$, 
	where the expectation $\mathbb{E}(\cdot)$ is approximated by Monte Carlo simulations. 
	We report these metrics, as well as the empirical coverage probabilities of all the confidence intervals for the target policy's value in Figures~\ref{fig:savemodel} and \ref{fig:savemodel-one-dimension} (see Appendix \ref{sec:moreexp} in the supplementary article). 
	We also calculate the standard deviation of these metrics in 400 replications and report them in Appendix \ref{sec:moreexp}. 
	
	It can be seen that COPE achieves the least bias and MSE among all methods. In addition, its 
	MSE decays with $N$ and $T$ in general and  the empirical coverage rate of our CI is close to the nominal level. We also notice that the squared bias of our estimator is much smaller than its MSE.
	This demonstrates the consistency of our method and is in line with our theoretical findings. In contrast, other baseline estimators are severely biased, since they cannot handle the unmeasured confounders. 
	

	\begin{figure}[!t]
		\centering
		\includegraphics[width=0.75\linewidth]{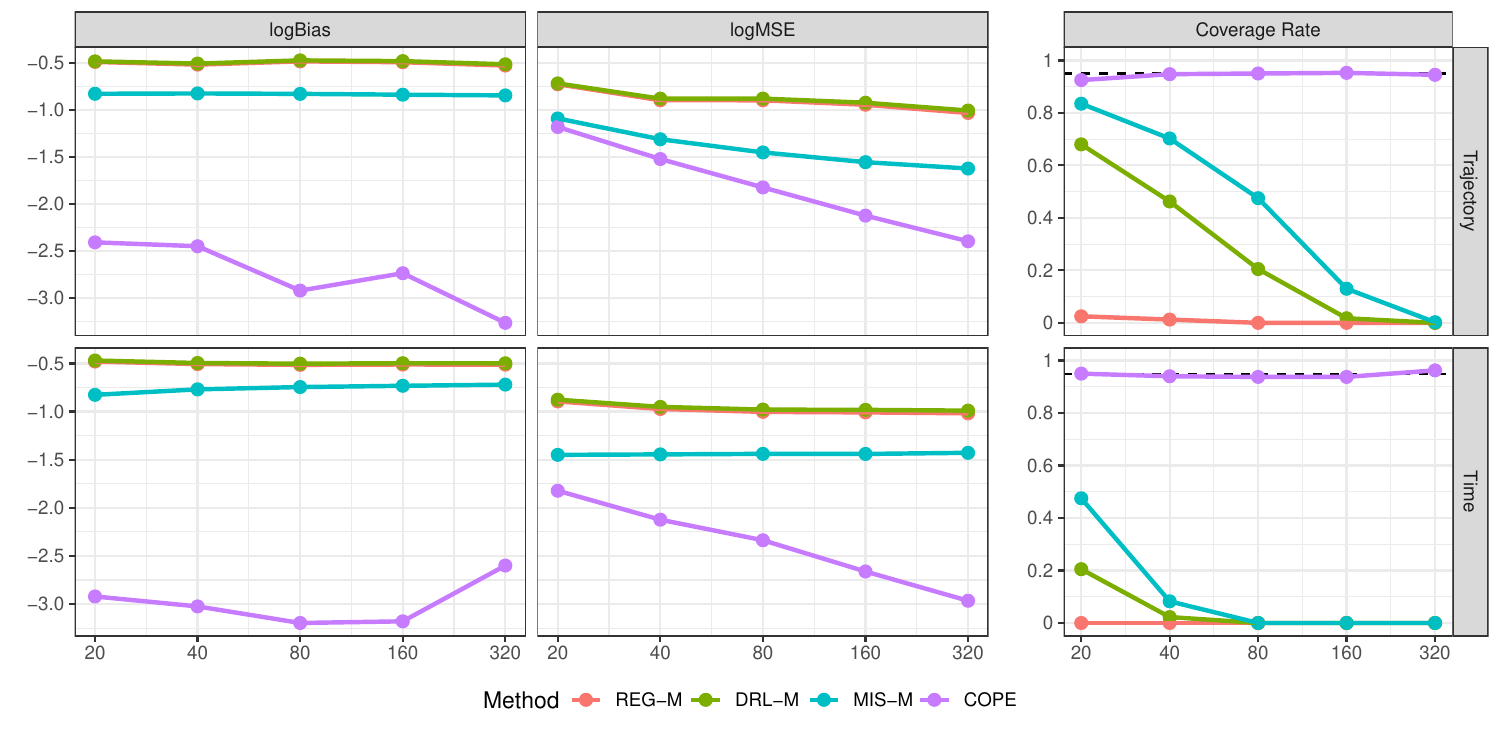}
		\caption{\small{
				Logarithms of the bias and mean square error (MSE), 
				and 95\% CI's coverage rate  of various OPE methods 
				with different combinations of $N$ and $T$ when $S_t$ is three-dimensional. 
				The black dash line corresponds to the confidence level 95\%. Top panels: $T$ is fixed to $20$ and $N\in \{20,40,80,160, 320\}$, bottom panels: $N$ is fixed to $20$ and $T\in \{20,40,80,160,320\}$. 
		}}\label{fig:savemodel}
	\end{figure}

	\section{Real Data Application} \label{sec:real data}
	In this section, we apply our method to a real dataset from a world-leading ride-hailing company. We focus on a particular recommendation program applied to customers in regions where there are more taxi drivers than the call orders. As we have commented, in the short term, this helps balance the taxi supply and passenger demand across different areas of the city. In the long term, this increases the frequency that the customer uses the app to request the trip.  

	The dataset consists of all the call orders at a given a city from September 16th to September 22th. 
	The features available to us consist of each order's time, origin, destination and 
	a supply-demand equilibrium metric that characterizes the degree that supply
	meets the demand. 
	For each of the call order, the customer might receive a coupon for 20\% off. This yields a binary action. 
	The mediator is the actual discount applied to the order. As we have commented, the mediator is calculated by the platform using
	the action and other promotion strategies and differs from the action, but is conditionally independent of those unmeasured variables that confound the action. Assumption 2(b) is thus satisfied.
	The reward is zero if the customer does not request the ride at the end, and one minus the actual discount times the price of the order otherwise. 
	We present the empirical quantiles of the reward and state in Table~\ref{tab:real-data-distribution}. 
	
	
	By definition, the reward depends on the action only through its effect on the mediator.
	In addition, the customer will observe the final discount applied to their ride on the
	application, but is not aware of which promotion strategy yields the discount. As such, it is 
	reasonable to assume that these promotion strategies will affect their behaviors through the
	final discount only. Consequently, the reward and future state are conditionally independent
	of the action and other promotion strategies given the mediator. Assumptions 2(a) and 2(c)
	thus hold.

	We first fit a MDP model to this dataset, 
	and use the estimated MDP to generate synthetic data to mimic the real dataset. 
	Specifically, the distribution of initial state is approximated by a multivariate normal distribution. 
	The state transition $S_{t+1}|A_t,S_t,M_t$ is modelled by a multivariate normal distribution $N(\mu(S_t,A_t,M_t),\sigma^2)$ where the conditional mean function $\mu$ is estimated using regularized linear basis function models. Similarly, we estimate the reward function $\Mean (R_t|S_t,A_t,M_t)$ using a regularized linear basis function model as well. 
	All the tuning parameters are selected by five-folds cross validation. 
	Based on the fitted MDP, synthetic dataset can be generated to evaluate different OPE methods. 

	
	We are interested in evaluating two recommendation policies. 
	One of them is a random policy (denote by $\pi_1$), with which each customer would have an equal chance to get a 20 percent discount with probability 0.5.  
	Another policy (denote by $\pi_2$) relies on the imbalance measure between supply and demand. 
	Specifically, for the region with extremely more vacant drivers than requests, we randomly choose $70\%$ percent of the customers for getting the discount.  
	For the rest of regions, the customers would have a $30\%$ chance for obtaining the discount. 
	We expect the second policy would yield larger values, as it has better immediate reward and encourages customers to request rides more often.

	To take customers' frequency of using the app into consideration, we use a slightly different definition of the value function as in \cite{xu2018large}, and adjust the proposed method and other baselines accordingly to reflect this change. In addition to $\{(S_{i,t},A_{i,t},M_{i,t},R_{i,t})\}_{i,t}$, the observed data consist of another sequence of variables $\{T_{i,t}\}$, corresponding to the time that the $i$th customer launches the app and enters the destination. We initialize $T_{i,0}$ to zero, for all $i$. The target policy's value is defined as $\eta^{\pi}=\sum_{t=0}^{+\infty} \Mean^{\pi}(\gamma^{T_{i,t}} R_{i,t} )$. 
To reflect this change, the proposed estimator takes the following form,
\begin{eqnarray*}
	\widehat{\eta}=\psi(O)=\psi_0+\frac{1}{NT}\sum_{j=1}^3\sum_{i,t} \psi_j'(O_{i,t}),
\end{eqnarray*}
where $\psi_0$ is the same as the direct estimator with the Q-estimator $\widehat{Q}$ replaced by $\widehat{Q}'$ detailed below, and for any $j$, $ \psi_j'(O_{i,t})$ is a version of $\psi_j(O_{i,t})$ with  $\gamma$ replaced by $\gamma^{T_{i,t+1}-T_{i,t}}$, $\widehat{Q}$ replaced by $\widehat{Q}'$ and $\widehat{\omega}$ replaced by $\widehat{\omega}'$. Specifically, $\widehat{Q}'$ is computed by solving a slightly different Bellman equation
\begin{eqnarray*}
	\Mean \Big\{R_{i,t}+\gamma^{T_{i,t+1}-T_{i,t}} \sum_{m,a',a} p_m(m|a',S_{i,t+1}) p_a^*(a|S_{i,t+1}) \pi(a'|S_{i,t+1}) Q^{\pi}(m,a,S_{i,t+1})\\ -Q^{\pi}(M_{i,t},A_{i,t},S_{i,t})\left.\right|M_{i,t},A_{i,t},S_{i,t}\Big\}=0,
\end{eqnarray*}
and $\widehat{\omega}'$ is computed by solving \eqref{eqn:solveL} with $\gamma$ replaced by $\gamma^{T_{i,t+1}-T_{i,t}}$ in \eqref{eqn:learnomega}. The DRL estimator can be similarly modified to adapt to this change. 


We apply REG-M, MIS-M, DRL-M and the proposed method COPE to evaluate the value difference $\eta^{\pi_2}-\eta^{\pi_1}$.
The ground truth OPE is approximated via Monte Carlo based on the fitted MDP model and equals 0.17. This is consistent with our expectation that the second policy yields a larger value. 
We evaluate the estimation accuracy by $\log$Bias and $\log$MSE as in the simulation section, and the coverage probability of the confidence interval. See Appendix \ref{sec:moretarget} in the supplementary article for the construction of the confidence interval of the value difference. 
The simulation results are aggregated over 500 replications. The discounted factor $\gamma$ is set to $0.99$, as we are interested in the long-term treatment effects. 

\begin{figure}[!t]
\centering
\includegraphics[width=0.8\linewidth]{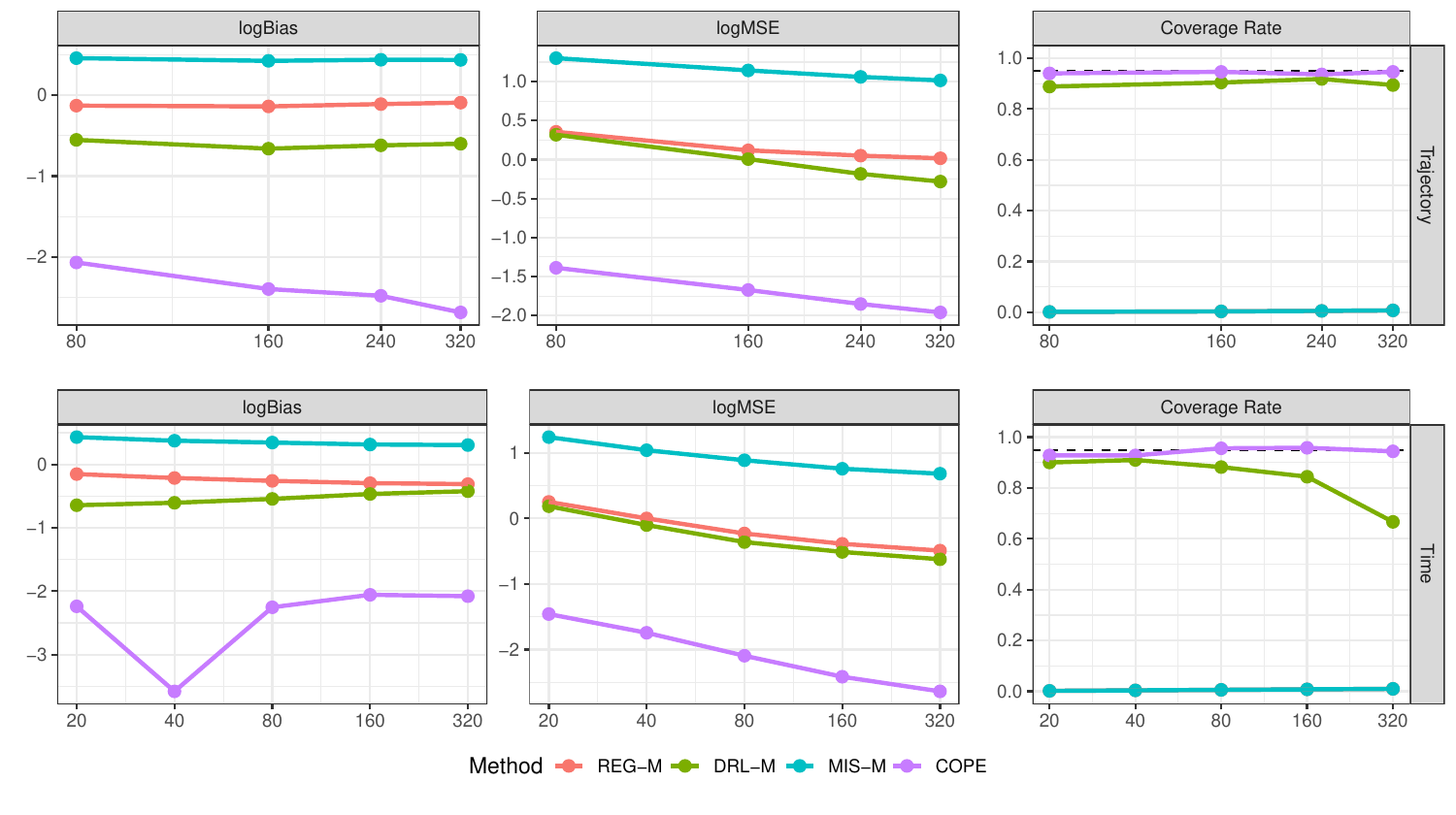}
\caption{\small Logarithms of the bias and mean square error (MSE), and 95\% CI's coverage rate  of various OPE methods with different combinations of $N$ and $T$ in simulated real data environment. 
	The black dash line in the right most panel corresponds to the confidence level 95\%. Top panels: $T$ is fixed to $20$ and $N\in \{80,160,240,320\}$, bottom panels: $N$ is fixed to $100$ and $T\in \{20,40,80,160,320\}$. }\label{fig:realmodel}
\end{figure}

Figure~\ref{fig:realmodel} depicts the performance of four methods. Results are summarized as follows. First, COPE has the best estimation accuracy among the four methods. Second, the coverage probability of the proposed CI is close to the nominal level. In contrast, the baseline methods fail to achieve the nominal coverage when $N$ or $T$ is large. These results are consistent with our simulation findings. 

We next apply our method and DRL to the real dataset to evaluate the value difference $\eta^{\pi_2}-\eta^{\pi_1}$. The proposed method yields a value difference of 0.63. The 95\% associated confidence interval is [0.03, 1.23]. As such, the second policy is significantly better than the first one. The result is consistent with our expectation. On the contrary, DRL yields a value difference
of -0.96. The associated confidence interval is [-2.07, 0.14]. According to DRL, the random policy is much better. This is due to that DRL cannot handle unmeasured confounders, leading to a biased estimator. Combining this with our theoretical and simulations results, we have more confidence about the findings of our proposed CI.

\section{Discussion}\label{Discussion}
In this section, we discuss several extensions. 
First, our current proposal relies on the ``memoryless unmeasured confounding'' assumption to simplify the derivation. 
In Appendix \ref{sec:memoryless} of the supplementary article, we discuss several possible relaxations of this assumption. 
Second, we assume the mediators are discrete to simplify the presentation. In Appendix \ref{sec:continuousM}, we extend the proposed method to settings with continuous mediators. Third, we adopt a discounted reward formulation to investigate the policy evaluation problem.  
We extend our proposal to the average reward setting in Appendix \ref{sec:appaveragereward} of the supplementary article. Fourth, we assume the mediator variable is completely observed. In the causal inference literature, \cite{chernofsky2021causal} considered settings with left censored mediators. They proposed three estimation methods, including (i) mediator model extrapolation; (ii) 
numerical integration and optimization of the observed data likelihood function; (iii)  the Monte Carlo Expectation-Maximization algorithm. In cases with partially observed mediators, we can couple their ideas with our proposal for value evaluation in confounded MDPs. We leave it for future research. 


\bibliographystyle{dcu}
\bibliography{references}

@article{uehara2021finite,
  title={Finite sample analysis of minimax offline reinforcement learning: Completeness, fast rates and first-order efficiency},
  author={Uehara, Masatoshi and Imaizumi, Masaaki and Jiang, Nan and Kallus, Nathan and Sun, Wen and Xie, Tengyang},
  journal={arXiv:2102.02981},
  year={2021}
}

@article{namkoong2020off,
	title={Off-policy policy evaluation for sequential decisions under unobserved confounding},
	author={Namkoong, Hongseok and Keramati, Ramtin and Yadlowsky, Steve and Brunskill, Emma},
	journal={Advances in Neural Information Processing Systems},
	volume={33},
	pages={18819--18831},
	year={2020}
}

@article{nair2021spectral,
  title={A spectral approach to off-policy evaluation for POMDPs},
  author={Nair, Yash and Jiang, Nan},
  journal={arXiv:2109.10502},
  year={2021}
}

@inproceedings{shi2021minimax,
	title={A minimax learning approach to off-policy evaluation in confounded Partially Observable Markov Decision Processes},
	author={Shi, Chengchun and Uehara, Masatoshi and Huang, Jiawei and Jiang, Nan},
	booktitle={International Conference on Machine Learning},
	pages={20057--20094},
	year={2022},
	organization={PMLR}
}

@inproceedings{shi2021deeply,
  title     = {Deeply-Debiased Off-Policy Interval Estimation},
  author    = {Shi, Chengchun and Wan, Runzhe and Chernozhukov, Victor and Song, Rui},
  booktitle = {Proceedings of the 38th International Conference on Machine Learning},
  pages     = {9580--9591},
  year      = {2021},
  editor    = {Meila, Marina and Zhang, Tong},
  volume    = {139},
  month     = {18--24 Jul},
  publisher = {PMLR}
}

@article{garreau2017large,
  title={Large sample analysis of the median heuristic},
  author={Garreau, Damien and Jitkrittum, Wittawat and Kanagawa, Motonobu},
  journal={arXiv:1707.07269},
  year={2017}
}

@article{shi2019sparse,
  title={A sparse random projection-based test for overall qualitative treatment effects},
  author={Shi, Chengchun and Lu, Wenbin and Song, Rui},
  journal={Journal of the American Statistical Association},
  year={2019},
  publisher={Taylor \& Francis}
}

@article{bennett2021proximal,
  title={Proximal Reinforcement Learning: Efficient Off-Policy Evaluation in Partially Observed Markov Decision Processes},
  author={Bennett, Andrew and Kallus, Nathan},
  journal={arXiv:2110.15332},
  year={2021}
}

@article{holmes2012fast,
  title={Fast nonparametric conditional density estimation},
  author={Holmes, Michael P and Gray, Alexander G and Isbell, Charles Lee},
  journal={arXiv:1206.5278},
  year={2012}
}

@article{mdn,
author = {Bishop, Christopher},
year = {1994},
month = {01},
pages = {},
title = {Mixture density networks}
}

@article{kallus2021causal,
  title={Causal Inference Under Unmeasured Confounding With Negative Controls: A Minimax Learning Approach},
  author={Kallus, Nathan and Mao, Xiaojie and Uehara, Masatoshi},
  journal={arXiv:2103.14029},
  year={2021}
}

@article{xu2020latent,
  title={Latent-state models for precision medicine},
  author={Xu, Zekun and Laber, Eric and Staicu, Ana-Maria and Severus, Emanuel},
  journal={arXiv:2005.13001},
  year={2020}
}

@inproceedings{shi2020does,
  title={Does the Markov decision process fit the data: testing for the Markov property in sequential decision making},
  author={Shi, Chengchun and Wan, Runzhe and Song, Rui and Lu, Wenbin and Leng, Ling},
  booktitle={International Conference on Machine Learning},
  pages={8807--8817},
  year={2020},
  organization={PMLR}
}

@article{qi2020robust,
  title={Robust Batch Policy Learning in Markov Decision Processes},
  author={Qi, Zhengling and Liao, Peng},
  journal={arXiv:2011.04185},
  year={2020}
}

@article{wu2020resampling,
	title={Resampling-based confidence intervals for model-free robust inference on optimal treatment regimes},
	author={Wu, Yunan and Wang, Lan},
	journal={Biometrics},
	year={2020},
	publisher={Wiley Online Library}
}

@article{zhang2013robust,
	title={Robust estimation of optimal dynamic treatment regimes for sequential treatment decisions},
	author={Zhang, Baqun and Tsiatis, Anastasios A and Laber, Eric B and Davidian, Marie},
	journal={Biometrika},
	volume={100},
	number={3},
	pages={681--694},
	year={2013},
	publisher={Oxford University Press}
}

@article{luedtke2016statistical,
	title={Statistical inference for the mean outcome under a possibly non-unique optimal treatment strategy},
	author={Luedtke, Alexander R and Van Der Laan, Mark J},
	journal={Annals of statistics},
	volume={44},
	number={2},
	pages={713},
	year={2016},
	publisher={NIH Public Access}
}

@article{liao2021off,
	title={Off-policy estimation of long-term average outcomes with applications to mobile health},
	author={Liao, Peng and Klasnja, Predrag and Murphy, Susan},
	journal={Journal of the American Statistical Association},
	volume={116},
	number={533},
	pages={382--391},
	year={2021},
	publisher={Taylor \& Francis}
}

@article{schmidt2020nonparametric,
	title={Nonparametric regression using deep neural networks with ReLU activation function},
	author={Schmidt-Hieber, Johannes and others},
	journal={Annals of Statistics},
	volume={48},
	number={4},
	pages={1875--1897},
	year={2020},
	publisher={Institute of Mathematical Statistics}
}

@article{ertefaie2014constructing,
	title={Constructing dynamic treatment regimes in infinite-horizon settings},
	author={Ertefaie, Ashkan},
	journal={arXiv:1406.0764},
	year={2014}
}

@article{hao2021bootstrapping,
	title={Bootstrapping Statistical Inference for Off-Policy Evaluation},
	author={Hao, Botao and Ji, Xiang and Duan, Yaqi and Lu, Hao and Szepesv{\'a}ri, Csaba and Wang, Mengdi},
	journal={arXiv:2102.03607},
	year={2021}
}

@article{wang2018quantile,
	title={Quantile-optimal treatment regimes},
	author={Wang, Lan and Zhou, Yu and Song, Rui and Sherwood, Ben},
	journal={Journal of the American Statistical Association},
	volume={113},
	number={523},
	pages={1243--1254},
	year={2018},
	publisher={Taylor \& Francis}
}

@article{zhang2012robust,
	title={A robust method for estimating optimal treatment regimes},
	author={Zhang, Baqun and Tsiatis, Anastasios A and Laber, Eric B and Davidian, Marie},
	journal={Biometrics},
	volume={68},
	number={4},
	pages={1010--1018},
	year={2012},
	publisher={Wiley Online Library}
}

@article{luedtke2017evaluating,
	title={Evaluating the impact of treating the optimal subgroup},
	author={Luedtke, Alexander R and van der Laan, Mark J},
	journal={Statistical methods in medical research},
	volume={26},
	number={4},
	pages={1630--1640},
	year={2017},
	publisher={SAGE Publications Sage UK: London, England}
}

@article{matsouaka2014evaluating,
	title={Evaluating marker-guided treatment selection strategies},
	author={Matsouaka, Roland A and Li, Junlong and Cai, Tianxi},
	journal={Biometrics},
	volume={70},
	number={3},
	pages={489--499},
	year={2014},
	publisher={Wiley Online Library}
}

@article{wager2018estimation,
	title={Estimation and inference of heterogeneous treatment effects using random forests},
	author={Wager, Stefan and Athey, Susan},
	journal={Journal of the American Statistical Association},
	volume={113},
	number={523},
	pages={1228--1242},
	year={2018},
	publisher={Taylor \& Francis}
}

@article{liao2020batch,
	title={Batch Policy Learning in Average Reward Markov Decision Processes},
	author={Liao, Peng and Qi, Zhengling and Murphy, Susan},
	journal={arXiv:2007.11771},
	year={2020}
}

@article{Jin2018,
	title={Ridesourcing, the sharing economy, and the future of cities},
	year={2018},
	author={Jin, S. T. and Kong, H. and Wu, R. and Sui, D. Z.},
	journal={Cities},
	volume={76},
	pages={96-104} 
}

@article{kosorok2019precision,
	title={Precision medicine},
	author={Kosorok, Michael R and Laber, Eric B},
	journal={Annual review of statistics and its application},
	volume={6},
	pages={263--286},
	year={2019},
	publisher={Annual Reviews}
}

@book{tsiatis2019dynamic,
	title={Dynamic Treatment Regimes: Statistical Methods for Precision Medicine},
	author={Tsiatis, Anastasios A and Davidian, Marie and Holloway, Shannon T and Laber, Eric B},
	year={2019},
	publisher={CRC press}
}

@inproceedings{xu2018large,
	title={Large-scale order dispatch in on-demand ride-hailing platforms: A learning and planning approach},
	author={Xu, Zhe and Li, Zhixin and Guan, Qingwen and Zhang, Dingshui and Li, Qiang and Nan, Junxiao and Liu, Chunyang and Bian, Wei and Ye, Jieping},
	booktitle={Proceedings of the 24th ACM SIGKDD International Conference on Knowledge Discovery \& Data Mining},
	pages={905--913},
	year={2018}
}

@article{hu2020personalized,
	title={Personalized Policy Learning using Longitudinal Mobile Health Data},
	author={Hu, Xinyu and Qian, Min and Cheng, Bin and Cheung, Ying Kuen},
	journal={arXiv:2001.03258},
	year={2020}
}

@inproceedings{robins2004optimal,
	title={Optimal structural nested models for optimal sequential decisions},
	author={Robins, James M},
	booktitle={Proceedings of the second seattle Symposium in Biostatistics},
	pages={189--326},
	year={2004},
	organization={Springer}
}

@article{chernozhukov2014gaussian,
	title={Gaussian approximation of suprema of empirical processes},
	author={Chernozhukov, Victor and Chetverikov, Denis and Kato, Kengo and others},
	journal={The Annals of Statistics},
	volume={42},
	number={4},
	pages={1564--1597},
	year={2014},
	publisher={Institute of Mathematical Statistics}
}

@article{dai2020coindice,
	title={CoinDICE: Off-Policy Confidence Interval Estimation},
	author={Dai, Bo and Nachum, Ofir and Chow, Yinlam and Li, Lihong and Szepesvari, Csaba and Schuurmans, Dale},
	journal={Advances in neural information processing systems},
	volume={33},
	year={2020}
}

@inproceedings{fan2020theoretical,
	title={A theoretical analysis of deep Q-learning},
	author={Fan, Jianqing and Wang, Zhaoran and Xie, Yuchen and Yang, Zhuoran},
	booktitle={Learning for Dynamics and Control},
	pages={486--489},
	year={2020},
	organization={PMLR}
}

@book{pearl2009causality,
	title={Causality},
	author={Pearl, Judea},
	year={2009},
	publisher={Cambridge university press}
}

@techreport{zhang2016markov,
	title={Markov decision processes with unobserved confounders: A causal approach},
	author={Zhang, Junzhe and Bareinboim, Elias},
	year={2016},
	institution={Technical Report R-23, Purdue AI Lab}
}

@article{kallus2019efficiently,
  title={Efficiently Breaking the Curse of Horizon in Off-Policy Evaluation with Double Reinforcement Learning},
  author={Kallus, Nathan and Uehara, Masatoshi},
  journal={arXiv},
  pages={arXiv--1909},
  year={2019}
}

@article{van1996weak,
	title={Weak Convergence and Empirical Processes},
	author={Van Der Vaart, AW and Wellner, JA},
	journal={Springer},
	volume={58},
	pages={59},
	year={1996}
}

@article{chernozhukov2018double,
  title={Double/debiased machine learning for treatment and structural parameters},
  author={Chernozhukov, Victor and Chetverikov, Denis and Demirer, Mert and Duflo, Esther and Hansen, Christian and Newey, Whitney and Robins, James},
  year={2018},
  publisher={Oxford University Press Oxford, UK}
}

@article{rahimi2008weighted,
  title={Weighted sums of random kitchen sinks: Replacing minimization with randomization in learning},
  author={Rahimi, Ali and Recht, Benjamin},
  journal={Advances in neural information processing systems},
  volume={21},
  pages={1313--1320},
  year={2008}
}

@inproceedings{jiang2016doubly,
  title={Doubly robust off-policy value evaluation for reinforcement learning},
  author={Jiang, Nan and Li, Lihong},
  booktitle={International Conference on Machine Learning},
  pages={652--661},
  year={2016},
  organization={PMLR}
}

@inproceedings{le2019batch,
	title={Batch Policy Learning under Constraints},
	author={Le, Hoang and Voloshin, Cameron and Yue, Yisong},
	booktitle={International Conference on Machine Learning},
	pages={3703--3712},
	year={2019}
}

@book{sutton2018reinforcement,
  title={Reinforcement learning: An introduction},
  author={Sutton, Richard S and Barto, Andrew G},
  year={2018},
  publisher={MIT press}
}

@inproceedings{nachum2019dualdice,
	title={Dualdice: Behavior-agnostic estimation of discounted stationary distribution corrections},
	author={Nachum, Ofir and Chow, Yinlam and Dai, Bo and Li, Lihong},
	booktitle={Advances in Neural Information Processing Systems},
	pages={2318--2328},
	year={2019}
}

@InProceedings{uehara2019minimax,
  title = 	 {Minimax Weight and Q-Function Learning for Off-Policy Evaluation},
  author =       {Uehara, Masatoshi and Huang, Jiawei and Jiang, Nan},
  booktitle = 	 {Proceedings of the 37th International Conference on Machine Learning},
  pages = 	 {9659--9668},
  year = 	 {2020},
  editor = 	 {III, Hal Daumé and Singh, Aarti},
  volume = 	 {119},
  month = 	 {13--18 Jul},
  publisher =    {PMLR}
}

@book{bickel1993efficient,
	title={Efficient and adaptive estimation for semiparametric models},
	author={Bickel, Peter J and Klaassen, Chris AJ and Bickel, Peter J and Ritov, Ya’acov and Klaassen, J and Wellner, Jon A and Ritov, YA'Acov},
	volume={4},
	year={1993},
	publisher={Johns Hopkins University Press Baltimore}
}

@book{van2000asymptotic,
	title={Asymptotic statistics},
	author={Van der Vaart, Aad W},
	volume={3},
	year={2000},
	publisher={Cambridge university press}
}

@inproceedings{thomas2015high2,
	title={High-confidence off-policy evaluation},
	author={Thomas, Philip S and Theocharous, Georgios and Ghavamzadeh, Mohammad},
	booktitle={Twenty-Ninth AAAI Conference on Artificial Intelligence},
	year={2015}
}

@article{feng2020accountable,
	title={Accountable off-policy evaluation with kernel bellman statistics},
	author={Feng, Yihao and Ren, Tongzheng and Tang, Ziyang and Liu, Qiang},
	journal={arXiv:2008.06668},
	year={2020}
}

@article{shi2020statistical,
	title={Statistical Inference of the Value Function for Reinforcement Learning in Infinite Horizon Settings},
	author={Shi, Chengchun and Zhang, Sheng and Lu, Wenbin and Song, Rui},
	journal={Journal of the Royal Statistical Society: Series B (Statistical Methodology)},
	volume={84},
	number={3},
	pages={765--793},
	year={2022},
	publisher={Wiley Online Library}
}

@book{tsiatis2007semiparametric,
  title={Semiparametric theory and missing data},
  author={Tsiatis, Anastasios},
  year={2007},
  publisher={Springer Science \& Business Media}
}

@article{li2016deep,
  title={Deep reinforcement learning for dialogue generation},
  author={Li, Jiwei and Monroe, Will and Ritter, Alan and Galley, Michel and Gao, Jianfeng and Jurafsky, Dan},
  journal={arXiv:1606.01541},
  year={2016}
}

@article{kober2013reinforcement,
  title={Reinforcement learning in robotics: A survey},
  author={Kober, Jens and Bagnell, J Andrew and Peters, Jan},
  journal={The International Journal of Robotics Research},
  volume={32},
  number={11},
  pages={1238--1274},
  year={2013},
  publisher={SAGE Publications Sage UK: London, England}
}

@article{li2020make,
  title={Who Make Drivers Stop? Towards Driver-centric Risk Assessment: Risk Object Identification via Causal Inference},
  author={Li, Chengxi and Chan, Stanley H and Chen, Yi-Ting},
  journal={arXiv:2003.02425},
  year={2020}
}

@article{murphy2003optimal,
  title={Optimal dynamic treatment regimes},
  author={Murphy, Susan A},
  journal={Journal of the Royal Statistical Society: Series B (Statistical Methodology)},
  volume={65},
  number={2},
  pages={331--355},
  year={2003},
  publisher={Wiley Online Library}
}

@article{chakraborty2014dynamic,
  title={Dynamic treatment regimes},
  author={Chakraborty, Bibhas and Murphy, Susan A},
  journal={Annual review of statistics and its application},
  volume={1},
  pages={447--464},
  year={2014},
  publisher={Annual Reviews}
}

@article{gottesman2019guidelines,
  title={Guidelines for reinforcement learning in healthcare},
  author={Gottesman, Omer and Johansson, Fredrik and Komorowski, Matthieu and Faisal, Aldo and Sontag, David and Doshi-Velez, Finale and Celi, Leo Anthony},
  journal={Nat Med},
  volume={25},
  number={1},
  pages={16--18},
  year={2019}
}

@inproceedings{mandel2014offline,
  title={Offline policy evaluation across representations with applications to educational games.},
  author={Mandel, Travis and Liu, Yun-En and Levine, Sergey and Brunskill, Emma and Popovic, Zoran},
  booktitle={AAMAS},
  pages={1077--1084},
  year={2014}
}

@article{luckett2020estimating,
  title={Estimating dynamic treatment regimes in mobile health using v-learning},
  author={Luckett, Daniel J and Laber, Eric B and Kahkoska, Anna R and Maahs, David M and Mayer-Davis, Elizabeth and Kosorok, Michael R},
  journal={Journal of the American Statistical Association},
  volume={115},
  number={530},
  pages={692--706},
  year={2020},
  publisher={Taylor \& Francis}
}

@article{wang2020provably,
  title={Provably Efficient Causal Reinforcement Learning with Confounded Observational Data},
  author={Wang, Lingxiao and Yang, Zhuoran and Wang, Zhaoran},
  journal={arXiv:2006.12311},
  year={2020}
}

@article{fulcher2020robust,
  title={Robust inference on population indirect causal effects: the generalized front door criterion},
  author={Fulcher, Isabel R and Shpitser, Ilya and Marealle, Stella and Tchetgen Tchetgen, Eric J},
  journal={Journal of the Royal Statistical Society: Series B (Statistical Methodology)},
  volume={82},
  number={1},
  pages={199--214},
  year={2020},
  publisher={Wiley Online Library}
}

@inproceedings{kallus2020confounding,
 author = {Kallus, Nathan and Zhou, Angela},
 booktitle = {Advances in Neural Information Processing Systems},
 editor = {H. Larochelle and M. Ranzato and R. Hadsell and M. F. Balcan and H. Lin},
 pages = {22293--22304},
 publisher = {Curran Associates, Inc.},
 title = {Confounding-Robust Policy Evaluation in Infinite-Horizon Reinforcement Learning},
 volume = {33},
 year = {2020}
}

@article{Rysman2009,
	title={The economics of two-sided markets},
	year={2009},
	author={Rysman, M. },
	journal={Journal of Economic Perspective},
	volume={23},
	pages={125-143} 
}

@inproceedings{tennenholtz2020off,
  title={Off-Policy Evaluation in Partially Observable Environments.},
  author={Tennenholtz, Guy and Shalit, Uri and Mannor, Shie},
  booktitle={AAAI},
  pages={10276--10283},
  year={2020}
}

@inproceedings{bennett2020off,
  title     = { Off-policy Evaluation in Infinite-Horizon Reinforcement Learning with Latent Confounders },
  author    = {Bennett, Andrew and Kallus, Nathan and Li, Lihong and Mousavi, Ali},
  booktitle = {Proceedings of The 24th International Conference on Artificial Intelligence and Statistics},
  pages     = {1999--2007},
  year      = {2021},
  editor    = {Banerjee, Arindam and Fukumizu, Kenji},
  volume    = {130},
  month     = {13--15 Apr},
  publisher = {PMLR}
}

@inproceedings{liu2018breaking,
  title={Breaking the curse of horizon: Infinite-horizon off-policy estimation},
  author={Liu, Qiang and Li, Lihong and Tang, Ziyang and Zhou, Dengyong},
  booktitle={Advances in Neural Information Processing Systems},
  pages={5356--5366},
  year={2018}
}

@inproceedings{elzayn2019fair,
  title={Fair algorithms for learning in allocation problems},
  author={Elzayn, Hadi and Jabbari, Shahin and Jung, Christopher and Kearns, Michael and Neel, Seth and Roth, Aaron and Schutzman, Zachary},
  booktitle={Proceedings of the Conference on Fairness, Accountability, and Transparency},
  pages={170--179},
  year={2019}
}

@article{shi2020multiply,
  title={Multiply robust causal inference with double-negative control adjustment for categorical unmeasured confounding},
  author={Shi, Xu and Miao, Wang and Nelson, Jennifer C and Tchetgen Tchetgen, Eric J},
  journal={Journal of the Royal Statistical Society: Series B (Statistical Methodology)},
  volume={82},
  number={2},
  pages={521--540},
  year={2020},
  publisher={Wiley Online Library}
}

@article{tchetgen2020introduction,
  title={An Introduction to Proximal Causal Learning},
  author={Tchetgen Tchetgen, Eric J and Ying, Andrew and Cui, Yifan and Shi, Xu and Miao, Wang},
  journal={arXiv:2009.10982},
  year={2020}
}

@article{chernofsky2021causal,
  title={Causal mediation analysis with mediator values below an assay limit},
  author={Chernofsky, Ariel and Bosch, Ronald J and Lok, Judith J},
  journal={arXiv:2107.14782},
  year={2021}
}
\appendix
\section{Further Discussions}\label{sec:ext}
\subsection{Memoryless Unmeasured Confounding}\label{sec:memoryless}
Our proposal relies on a ``memoryless unmeasured confounding assumption'' that requires the unmeasured confounders $U_t$ to be conditionally independent of the past observations and latent confounders given $S_t$, at each time $t$. This assumption might be violated in some applications. We discuss several possible relaxations of this assumption in this section. 

One way to relax this assumption is to allow $U_t$ to depend on past observations. 
Specifically, we impose the following ``high-order memoryless unmeasured confounding assumption'': For some integer $K\ge 1$, we assume $U_t$ is conditionally independent of $\{ (S_j,U_j,A_j,M_j,R_j): j\le t-K \}$ given $S_t \cup \{ (S_j,A_j,M_j,R_j): t-K< j<t \}$. Suppose $K$ is known to us. If we define a new state vector $S_t^*=S_t \cup \{ (S_j,A_j,M_j,R_j): t-K< j<t \}$, then $U_t$ satisfies the memoryless unmeasured confounding assumption given $S_t^*$. As such, our proposal is equally applicable to the transformed dataset. 

Another way to relax this assumption is to allow $U_t$ to depend on the past latent factors. In that case, the state vectors no longer satisfy the Markov assumption and the data follow a partially observable MDP (POMDP) model. Under partial observability, Theorem \ref{thm:identifiable} is not applicable for value identification. However, under Assumption 2, the front-door adjustment formula can still be applied to handle unmeasured confounders. It would be practically interesting to further investigate the direct, IS and DR estimators under the POMDP model, but this is beyond the scope of the current paper. We leave it for future research. 

	To summarize, we have discussed two potential relaxations of the memoryless unmeasured confounding assumption, one with a ``high-order memoryless unmeasured confounding" assumption and the other without any further assumptions. 
	These assumptions are not directly testable. However, notice that under the (high-order) memoryless unmeasured confounding assumption, the observed data process $\{(S_j,A_j,M_j,R_j):j\ge 0\}$ forms a (high-order) MDP. Without these assumptions, this process forms a POMDP. In practice, we can apply the sequential testing procedure developed by \citet{shi2020does} to the observed data for model selection. Specifically, if the MDP assumption is satisfied, then we can apply the proposed procedure in the main paper for value evaluation. If the data satisfies a $K$th-order MDP assumption, then the $K$th-order memoryless unmeasured confounding assumption is likely to hold. In that case, we can first construct the new state vector by concatenating measurements over the past $K$ time points and then apply our proposal.  Suppose we concatenate measurements over sufficiently many decision points and the test still rejects the MDP assumption. Then the ``high-order memoryless unmeasured confounding" assumption is likely to be violated and we may apply the estimators designed for POMDPs for policy evaluation.
	
		
	
	\subsection{Continuous Mediators}\label{sec:continuousM}
	As we have commented in the main paper, the proposed method is equally applicable to settings with continuous mediators. To elaborate, first, notice that the Q-function  and the marginal density ratio are well-defined regardless of whether the mediator is continuous or not. In addition, the estimating procedure discussed in Section \ref{sec:Qomega} remains consistent with continuous mediators. Second, the proposed estimator involves the conditional probability mass function of the mediator $p_m(m|a,s)=\prob(M=m|A=a,S=s)$. In settings with continuous mediators, we will replace it with the corresponding conditional probability density function of the mediator and replace the sum with the integral. This yields the following estimating function, $\psi(O)=\psi_0+\sum_{j=1}^3 \psi_j(O)$ where 
	\begin{eqnarray*}
		\psi_0=\int_m \sum_{a,a',s} p_m(m|a',s)\pi(a'|s) p_a^*(a|s) Q^{\pi}(m,a,s)\nu(s)dm,\\
		\psi_1(O)=\frac{1}{1-\gamma}\omega^{\pi}(S)\rho(M,A,S)\Big\{R+\gamma \int_m \sum_{a,a^*}Q^{\pi}(m,a,S')p_m(m,a^*,S')p_a^*(a|S')\pi(a^*|S')dm\Big. 
		\\-\Big.Q^{\pi}(M,A,S)\Big\},\\
		\psi_2(O)=\frac{1}{1-\gamma}\omega^{\pi}(S)\frac{\pi(A|S)}{p_a^*(A|S)} \sum_a p_a^*(a|S) \Big\{Q^\pi(M,a,S)-\sum_m p_m(m|A,S) Q^\pi(m,a,S)\Big\},\\
		\psi_3(O)=\frac{1}{1-\gamma}\sum_{m,a'} \omega^{\pi}(S) p_m(m|a',S)\pi(a'|S) \Big\{ Q^\pi(m,A,S)-\sum_{a} Q^\pi(m,a,S)p_a^*(a|S) \Big\}.
	\end{eqnarray*} 
	Third, in cases where the dimensions of $M$ and $S$ are small, kernel smoothers can be employed to estimate $p_m$ \citep{holmes2012fast}. In high-dimensional settings, we can apply state-of-the-art deep generative learning algorithms such as mixture density networks \citep{mdn} to learn $p_m$. 
	
	\subsection{Multiple Target Policies}\label{sec:moretarget}
	We first establish the joint asymptotic distribution of the proposed value estimators under multiple target policies $\{\pi_k\}_{k=1}^K$. We next introduce the proposed confidence interval for the value difference between two target policies and prove its validity. 
	
	Let $\widehat{\eta}_{\pi}$ denote the proposed estimator in \eqref{eqn:valueest} with the target policy given by $\pi$. Similarly, let $\widehat{Q}^{\pi}$ and $\widehat{\omega}^{\pi}$ denote the corresponding estimated Q-function and marginal density ratio, for any $\pi$. We have the following results. 
	
	\begin{thm}\label{thm6}
		Suppose the conditions in Theorem \ref{thm2} are satisfied. Suppose $\widehat{Q}^{\pi_k}$, $\widehat{\omega}^{\pi_k}$, $\widehat{p}_a^*$, $\widehat{p}_m$ converge in $L_2$-norm to their oracle values at a rate of $N^{-\kappa^*}$ for some $\kappa^*>1/4$ and any $1\le k\le K$. Then $\sqrt{N}(\widehat{\eta}^{\pi_1}-\eta^{\pi_1},\cdots,\widehat{\eta}^{\pi_K}-\eta^{\pi_K})$ are jointly asymptotically normal with covariance matrix given by $\{\sigma_{k_1,k_2}\}_{1\le k_1,k_2\le K}$ where $\sigma_{k_1,k_2}$ equals
		\begin{eqnarray*}
			\Cov\Big\{\sum_{m,a,a',S_0}Q^{\pi_{k_1}}(m,a,S_0)p_m(m,a',S_0)\pi_{k_1}(a'|S_0)p_a^*(a|S_0),Q^{\pi_{k_2}}(m,a,S_0)p_m(m,a',S_0)\pi_{k_2}(a'|S_0)p_a^*(a|S_0)\Big\}\\+\frac{1}{T}\Cov\Big\{\sum_{j=1}^3 \psi_j^{\pi_{k_1}}(O_0), \sum_{j=1}^3 \psi_j^{\pi_{k_2}}(O_0)\Big\},
		\end{eqnarray*}
		where $\psi_j^{\pi}$ is equal to $\psi_j$ defined in \eqref{sec:ourproposal} with the target policy given by $\pi$. 
	\end{thm}
	The proof of Theorem \ref{thm6} is very similar to that of Theorem \ref{thm3} and is thus omitted to save space. 
	
	We next construct a confidence interval for $\eta^{\pi_2}-\eta^{\pi_1}$. It follows from Theorem \ref{thm6} that the asymptotic variance of $\sqrt{N}(\widehat{\eta}^{\pi_2}-\widehat{\eta}^{\pi_1}-\eta^{\pi_2}+\eta^{\pi_1})$ is given by
	\begin{eqnarray}\label{eqn:diffvar}
		\begin{split}
			\Var\Big\{\sum_{m,a,a',S_0}Q^{\pi_{k_1}}(m,a,S_0)p_m(m,a',S_0)\{\pi_2(a'|S_0)-\pi_1(a'|S_0)\}p_a^*(a|S_0)\Big\}\\
			+\frac{1}{T}\Var\Big\{\sum_{j=1}^3 \{\psi_j^{\pi_2}(O_0)- \psi_j^{\pi_1}(O_0)\}\Big\}.
		\end{split}	
	\end{eqnarray}
	To estimate \eqref{eqn:diffvar}, we observe that $\widehat{\eta}^{\pi_2}-\widehat{\eta}^{\pi_1}$ can be represented as $N^{-1}\sum_{i=1}^N (\eta^{\pi_2}_i-\eta^{\pi_1}_i)$ where $\eta^{\pi}_i$ denotes the estimating function for $\eta^{\pi}$ based on the $i$th trajectory only. We thus use the sampling variance estimator of $\{\eta^{\pi_2}_i-\eta^{\pi_1}_i\}_i$, denoted by $\widehat{\sigma}^2_{\eta}$, to estimate \eqref{eqn:diffvar}. This yields the following $100(1-\alpha)\%$ confidence interval for the value difference:
	\begin{eqnarray}\label{eqn:confdiff}
		[\widehat{\eta}^{\pi_2}-\widehat{\eta}^{\pi_1}-z_{\alpha/2} N^{-1/2}\widehat{\sigma}_{\eta}, \widehat{\eta}^{\pi_2}-\widehat{\eta}^{\pi_1}+z_{\alpha/2} N^{-1/2}\widehat{\sigma}_{\eta}].
	\end{eqnarray}
	\begin{thm}\label{thm7}
		Suppose the conditions in Theorem \ref{thm6} are satisfied and \eqref{eqn:diffvar} is bounded away from zero. Then the coverage probability of the proposed CI in \eqref{eqn:confdiff} approaches to the nominal level as $N$ diverges to infinity.
	\end{thm}
	
	Finally, we remark that in the statement of Theorem \ref{thm7}, the asymptotic variance of the value difference estimator is required to be bounded away from zero. This essentially requires the two target policies $\pi_1$ and $\pi_2$ to not be too close to each other. In the extreme case where $\pi_1=\pi_2$, the asymptotic variance equals to zero and the value difference estimator has a degenerate distribution. To address this concern, we could redefine the variance estimator by setting $\widehat{\sigma}^2_{\eta}\leftarrow \max(\widehat{\sigma}^2_{\eta}, \delta^2)$ for some $\delta>0$, as in \cite{luedtke2017evaluating}. This guarantees that the variance estimator is strictly positive and the resulting CI would be valid as long as $\delta\gg T^{-1/6}$ \citep[see Theorem 3.1 of][]{shi2019sparse}. 
	
	\subsection{Average Reward}\label{sec:appaveragereward}
	We extend our proposal to the average reward setting in this section. Under this formulation, the target value can be represented as $\eta^{\pi}=\lim_{T\to \infty} T^{-1} \sum_{t=0}^{T-1}\Mean^{\pi} R_{t}$. Similar to the discounted reward setting, the proposed estimating function takes the following form $\psi(O)=\psi_0+\sum_{j=1}^3 \psi_j(O)$ where $\psi_0$ denotes the direct estimator whose form will be specified later and $\{\psi_j\}_j$ are the augmentation terms.
	
	Specifically, $\psi_1(O)$ is equal to
	\begin{eqnarray*}
		\omega^{\pi}(S)\rho(M,A,S)\Big\{R+ \sum_{m,a,a^*}Q^{\pi}(m,a,S')p_m(m,a^*,S')p_a^*(a|S')\pi(a^*|S') 
		-Q^{\pi}(M,A,S)-\psi_0\Big\},
	\end{eqnarray*}
	where $Q^{\pi}(m,a,s)$ denotes the relative value function $\sum_{t=0}^{+\infty} \Mean^{\pi} (R_t-\eta^{\pi}|M_0=m,A_0=a,S_0=s)$, $\omega^{\pi}$ denotes the density ratio of the stationary state distribution under $\pi$ and that under the behavior policy, and the definitions of $\rho$, $p_m$, $p_a^*$, $\psi_2(O)$ and $\psi_3(O)$ are consistent to those in the main paper. 
	
	We next discuss the estimating procedures for $(Q^{\pi},\psi_0)$ and $\omega^{\pi}$. Similar to Equation (3.2) in \cite{liao2020batch}, we can show that the relative value function satisfies the following Bellman equation in the presence of unmeasured confounders, 
	\begin{eqnarray*}
		\Mean \Big\{R+ \sum_{m,a,a^*}Q^{\pi}(m,a,S')p_m(m,a^*,S')p_a^*(a|S')\pi(a^*|S')|M,A,S\Big\}=Q^{\pi}(M,A,S)+\psi_0.
	\end{eqnarray*}
	Given the estimators for $p_m$, $p_a$, we can adopt the general coupled estimation framework to jointly learn $Q^{\pi}$ and $\psi_0$ from the observed data \citep[see Section 4.2 in][]{liao2020batch}. 
	
	As for $\omega^{\pi}$, we note that it satisfies the following equation,
	\begin{eqnarray*}
		\Mean \omega^{\pi}(S)\Big\{ f(S)-\rho(M|A,S) f(S')\Big\}=0,
	\end{eqnarray*}
	for any testing function $f$. The resulting estimator can thus be computed by solving a minimax problem, as detailed in Appendix \ref{sec:moreexp}. 
	
	\subsection{Joint Effect} 
	To conclude this section, we remark that in this paper, our interest lies in analyzing the effect of one individual customer recommendation program while controling for all other promotion programs. In our formulation, this particular recommendation program serves as the action whereas other promotion programs will affect the mediator. Suppose one is interested in the joint effect of all promotion programs. In that case, we may treat the mediator (e.g., the final discount) as the action for policy evaluation.

	\section{More on the Numerical Experiments}\label{sec:moreexp}
	We first discuss how we estimate $\omega^\pi$. 
	{\color{black}One way to simply the calculation is to set $\mathcal{F}$ to a unit ball of a reproducing kernel Hilbert space (RKHS) based on some symmetric positive definite kernel $\kappa$. This yields a closed form expression for the inner maximization. Specifically, based on the observed data, we aim to identify $\omega$ that minimizes the following objective function
		\begin{eqnarray*}
			-\frac{2(1-\gamma)}{\sum_i T_i}\sum_{s} \omega^{\pi}(S_{i,t})\left\{\kappa(s,S_{i,t})-\gamma \frac{\sum_{a} \pi(a|S_{i,t})p_m(M_{i,t}|a,S_{i,t}) }{p_m(M_{i,t}|A_{i,t},S_{i,t})}\kappa(s,S_{i,t+1})\right\}\nu(s)\\
			+\sum_{(i_1,t_1)\neq (i_2,t_2)}\frac{\omega^{\pi}(S_{i_1,t_1})\omega^{\pi}(S_{i_2,t_2})}{\sum_i T_i (\sum_i T_i-1)} \left\{ \gamma^2\frac{\sum_{a} \pi(a|S_{i_1,t_1})p_m(M_{i_1,t_1}|a,S_{i_1,t_1}) }{p_m(M_{i_1,t_1}|A_{i_1,t_1},S_{i_1,t_1})}\frac{\sum_{a} \pi(a|S_{i_2,t_2})p_m(M_{i_2,t_2}|a,S_{i_2,t_2}) }{p_m(M_{i_2,t_2}|A_{i_2,t_2},S_{i_2,t_2})}\right.\\
			\left.\times \kappa(S_{i_1,t_1+1},S_{i_2,t_2+1})+\kappa(S_{i_1,t_1},S_{i_2,t_2})-2\gamma\kappa(S_{i_1,t_1},S_{i_2,t_2+1})\frac{\sum_{a} \pi(a|S_{2,t})p_m(M_{i_2,t_2}|a,S_{i_2,t_2}) }{p_m(M_{i_2,t_2}|A_{i_2,t_2},S_{i_2,t_2})} \right\} \\
			+(1-\gamma)^2 \sum_{s_1,s_2} \kappa(s_1,s_2)\nu(s_1)\nu(s_2).
		\end{eqnarray*} 
		Stochastic gradient descent algorithms can be applied to update the parameters in $\omega^{\pi}$; see e.g., Algorithm 2 of \cite{shi2021deeply}. 
		
		In practice, we recommend to use the Gaussian radial basis function kernel. The bandwidth parameter in the kernel can be selected via the median heuristic \citep[see e.g.,][]{garreau2017large}. In high-dimensional settings, we can set $\Omega$ to a class of deep neural networks (DNN). The DNN structure can be adaptively determined via cross-validation. We also conduct a simulation study and find that the method is convergent in the optimization. For instance, Figure \ref{fig:ann_loss} depicts the change of loss function during the training process. It can be seen that the objective function converges as the number of training epochs reaches to 1000. In this example, we use a multilayer perceptron (MLP) neural network with a single hidden layer and five hidden nodes. The initial parameter values are randomly assigned and the learning rate is fixed to 0.01. In addition, it can be seen from Table \ref{tab:nonlinear} that MSEs of the resulting value estimators are smaller than or comparable to those presented in Section \ref{sec:compare}, based on linear function approximation.
		
		
		\begin{table}[t]
			\caption{Logarithms of MSEs of DRL and COPE. $T=20$. }\label{tab:nonlinear}
			\begin{tabular}{rrrrrr}
				\hline
				N & 20 & 40 & 80 & 160 & 320 \\ 
				\hline
				DRL & -1.26 & -1.46 & -1.49 & -1.58 & -1.71 \\ 
				COPE with linear function approximation & -1.57 & -1.85 & -2.14 & -2.48 & -2.78 \\ 
				COPE with nonlinear function approximation & -1.63 & -1.84 & -2.14 & -2.46 & -2.78 \\ 
				\hline
			\end{tabular}
		\end{table}
		
		\begin{figure}[t]
			\includegraphics[width=\linewidth]{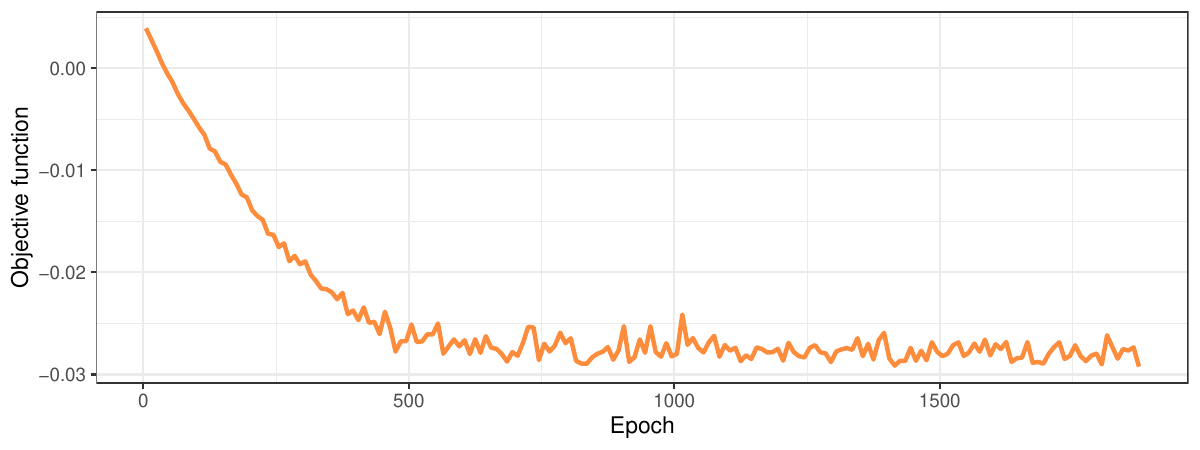}
			\caption{Line plots of objective function over training epochs on a synthetic dataset comprising 160 trajectory with $T = 20$.}\label{fig:ann_loss}
		\end{figure}
	}


	{\color{black}In our implementation, we set both $\Omega$ and $\mathcal{F}$ to linear function classes, i.e, 
		$\omega^{\pi}$ and $f$ can be characterized by a linear combination of $d_{\omega}$-dimensional  random Fourier features $\xi$ following the Random Kitchen Sinks (RKS) algorithm \citep{rahimi2008weighted}.} 
	Without loss of generality, suppose $\omega^{\pi}(s)=\xi^\top(s) \beta^*$ for some $\beta^* \in \mathbb{R}^{d_\omega}$. 
	Due to $L(\omega^\pi, f) = 0$, $\beta^*$ is the solution to
	\begin{eqnarray*}
		\Mean \beta^{*\top}  \xi(S_{i,t})\left\{\xi(S_{i,t})-\gamma \frac{ \sum_{a \sim \pi(\bullet | S_{i, t})} \widehat{p}_m(M_{i,t}|a,S_{i,t})}{\widehat{p}_m(M_{i,t}|A_{i,t},S_{i,t})}\xi(S_{i,t+1}) \right\}=(1-\gamma) \sum_s \xi(s)\nu_0(s).
	\end{eqnarray*}
	Consequently, we can estimate $\beta^*$ by
	\begin{eqnarray*}
		\widehat{\beta}&=& \left[\frac{1}{NT}\sum_{i=1}^N\sum_{t=0}^{T-1}\xi(S_{i,t})\left\{\xi^\top(S_{i,t})-\gamma \frac{ \sum_{a \sim \pi(\bullet | S_{i, t})} \widehat{p}_m(M_{i,t}|a,S_{i,t})}{\widehat{p}_m(M_{i,t}|A_{i,t},S_{i,t})}\xi^\top(S_{i,t+1}) \right\}\right]^{-1}\\
		& \times & \left\{(1-\gamma) \sum_s \xi(s)\nu_0(s)\right\}.
	\end{eqnarray*}
	{\color{black}In our simulation studies, the number of features $d_{\omega}$ is set to $6d_S$ where $d_S$ corresponds to the dimension of the state. In addition, we further conduct a sensitivity analysis to investigate the empirical performance of the proposed estimator with different number of basis functions. Figure \ref{fig:basis} reports biases, mean square errors and 95\% CIs' coverage rates. It can be seen that the performance of the proposed procedure is not overly sensitive to the number of basis functions.
		\begin{figure}[t]
			\begin{center}
				\includegraphics[width=14cm]{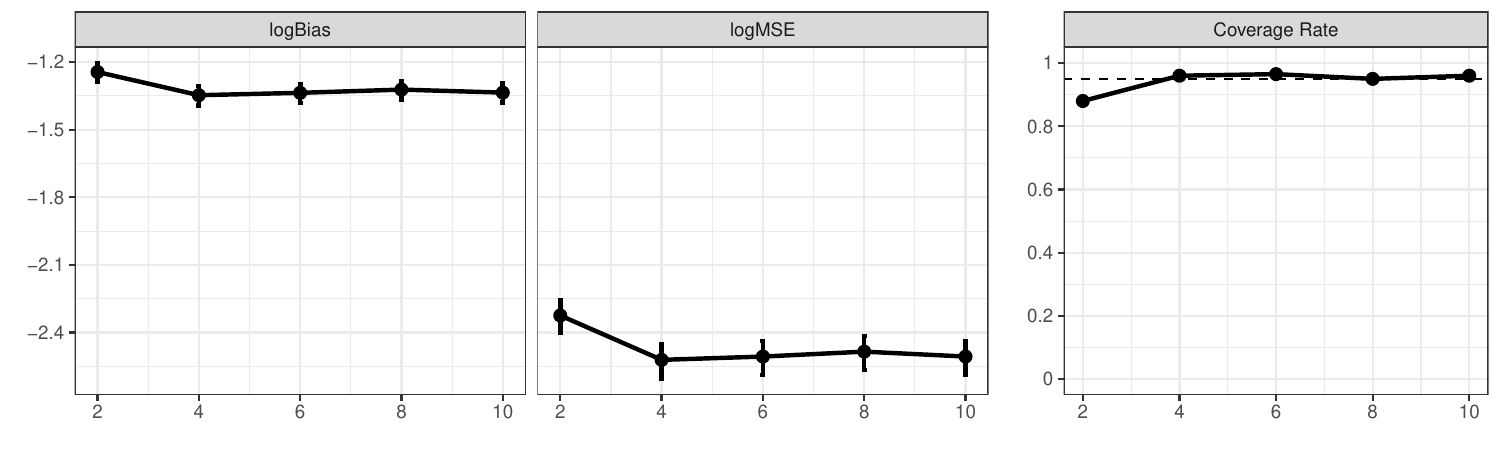}
				\includegraphics[width=14cm]{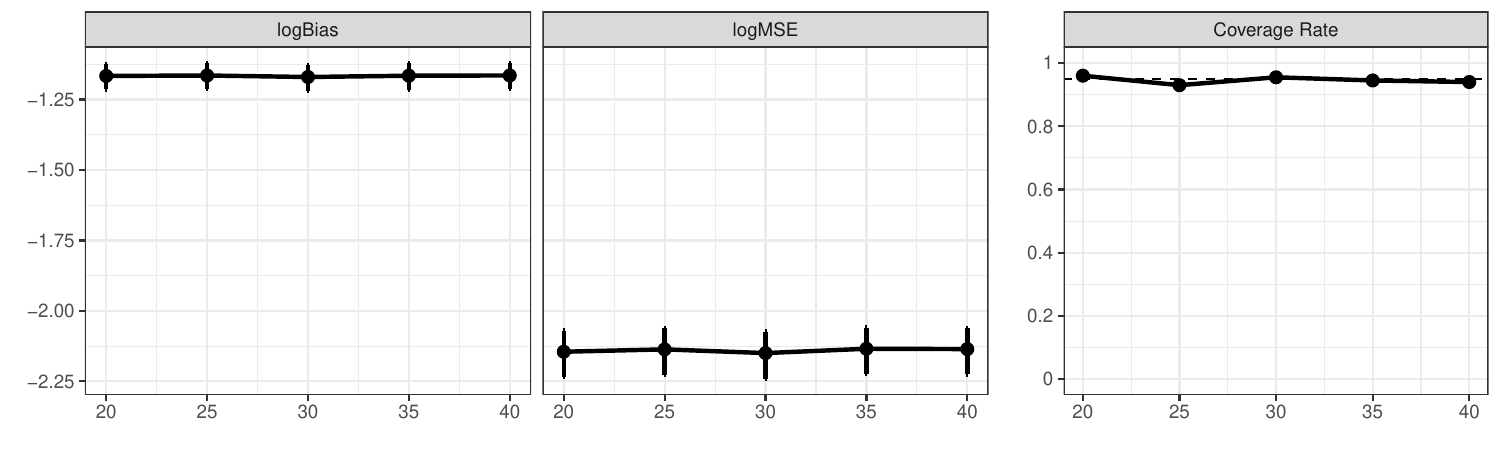}
			\end{center}
			\caption{\small Logarithms of the bias and mean square error (MSE), and 95\% CI's coverage rate  of the proposed OPE method with different number of basis functions when $S_t$ with dimension 1 (top panels) and 3 (bottom panels).}\label{fig:basis}
		\end{figure}
		%
		
		
		
		
		Next, we discuss how we estimate $Q^{\pi}$. In our implementation, we use linear basis functions with random Fourier features to model the Q-function. It suffices to solve a least-square regression during each Q-iteration. The number of basis functions is set to $5(d_S+2)$. The initial parameter is randomly generated from a multivariate normal distribution with zero mean and identity covariance matrix. To prevent overfitting, we add a ridge penalty function to shrink the least-square estimator. This guarantees that the estimated Q-function will not diverge. The regularization parameter is set to $10^{-3}$. We stop the Q-iteration when the difference between the estimated regression coefficients from one iteration to another is less than $10^{-4}$.
		
		Next, we similarly use linear basis functions with random Fourier features to parametrize $\widehat{p}_{a}^{*}$ and $\widehat{p}_{m}$. The number of basis functions are set to $d_S$ and $d_S+1$, respectively.}

	
	Next, we discuss how we estimate the variance of all baseline estimators. The variance of $\widehat{\eta}_{\textrm{REG}}$ is estimated by the sampling variance estimator of $\{\widehat{Q}(S_{i, 0}, \pi(S_{i, 0}))\}_i$. As for the MIS estimator, we first compute the value estimator based on the data for each individual trajectory as
	\begin{equation*}
		\widehat{\eta}^{\textrm{MIS}}_i = \frac{1}{T+1} \sum_{t=0}^T R_{i, t} \widehat{\omega}(S_{i, t}) \frac{\pi(A_{i, t}|S_{i, t})}{\widehat{p}_a^*(A_{i, t}|S_{i, t})}  .
	\end{equation*}
	We estimate the variance of MIS as the sampling variance estimator of $\{\widehat{\eta}^{\textrm{MIS}}_i\}_i$. The variance estimator of DRL can be similarly derived as that of the proposed estimator. Given these variance estimators, we use Wald-type confidence intervals to infer the target policy's value. {\color{black}All the experiments are conducted on a 64-bit Windows platform with Intel(R) Core(TM) i9-9940X CPU @ 3.30GHz.}

\begin{figure}[!t]
		\centering
		\includegraphics[width=0.75\linewidth]{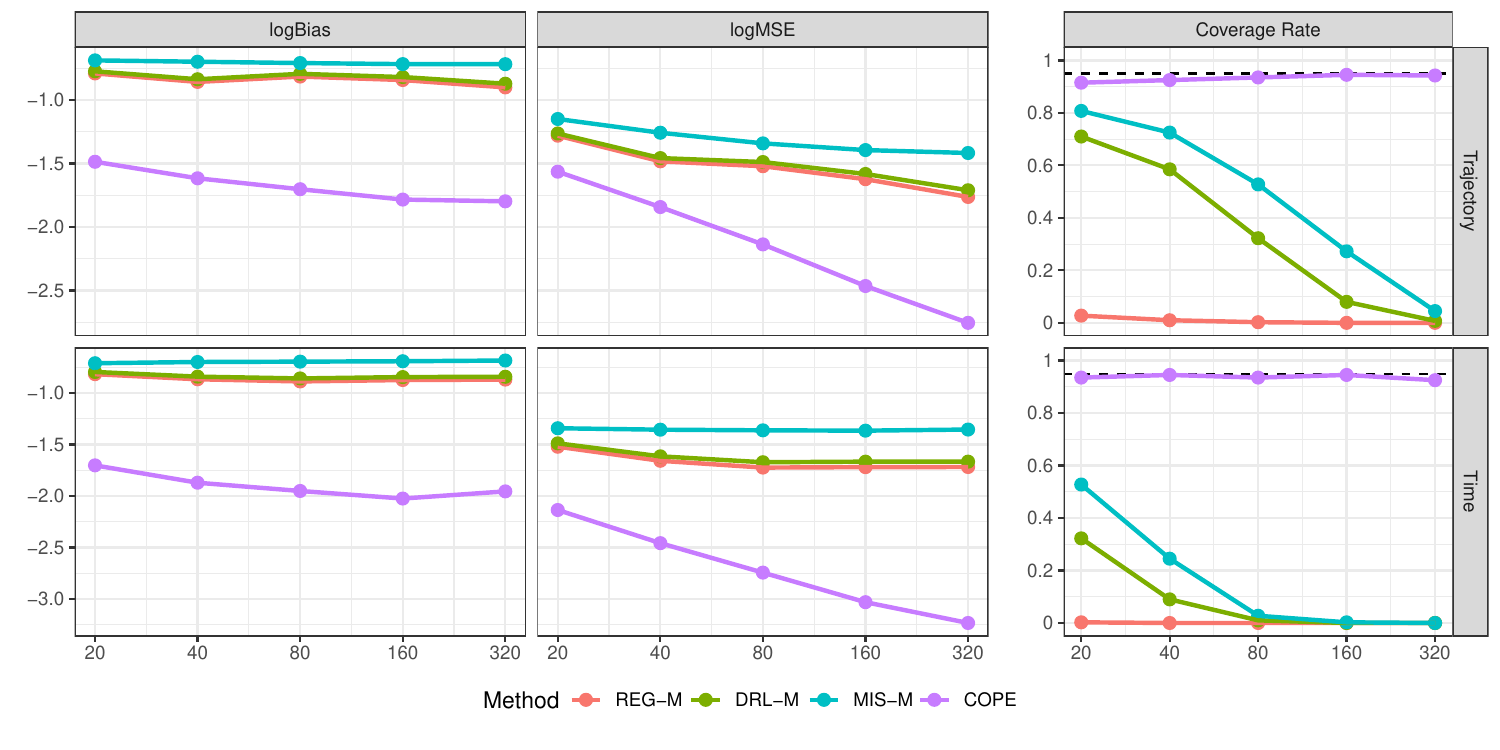}
		\vspace{-8pt}
		\caption{\small 
			Logarithms of the bias (left panels), 
			logarithms of the mean square error (middle panels), 
			and 95\% CI's coverage rate (right panels) of OPE methods 
			with different combinations of $N$ and $T$ 
			when $S_t \in \mathbb{R}$. Top panels: $T$ is fixed to $20$ and $N\in \{20,40,80,160, 320\}$, bottom panels: $N$ is fixed to $20$ and $T\in \{20,40,80,160,320\}$. 
		}\label{fig:savemodel-one-dimension}
	\end{figure}

	Finally, when state is one-dimensional, we report the bias, MSE of various OPE estimators and the coverage rate of their associated confidence intervals in Figure \ref{fig:savemodel-one-dimension}. Reported in Tables~\ref{tab:sim-N-compare-std} and~\ref{tab:sim-T-compare-std} are the standard deviations of biases and MSEs of various OPE estimators under settings described in Section~\ref{sec:compare}. Quantiles of the state and reward are reported in Table \ref{tab:real-data-distribution}.

	
	\begin{table}[ht]
		\centering
		\begin{tabular}{l|c|rrrrr}
			\toprule
			Standard deviation & Estimator & 20   & 40   & 80   & 160  & 320  \\
			\midrule
			\multirow{4}{*}{Bias}
			& REG-M   & 0.014 & 0.009 & 0.007 & 0.005 & 0.003 \\
			& DRL-M   & 0.014 & 0.009 & 0.007 & 0.005 & 0.003 \\
			& MIS-M   & 0.012 & 0.008 & 0.006 & 0.004 & 0.003 \\
			& COPE    & 0.013 & 0.009 & 0.006 & 0.004 & 0.003 \\    
			\midrule
			\multirow{4}{*}{MSE}
			& REG-M     & 0.0111 & 0.0062 & 0.0049 & 0.0032 & 0.0020 \\ 
			& DRL-M     & 0.0113 & 0.0063 & 0.0050 & 0.0033 & 0.0021 \\ 
			& MIS-M     & 0.0052 & 0.0029 & 0.0020 & 0.0013 & 0.0009 \\ 
			& COPE      & 0.0046 & 0.0020 & 0.0010 & 0.0005 & 0.0003 \\ 
			\bottomrule
		\end{tabular}
		\caption{Standard deviations of biases and MSEs of various OPE estimators under settings described in Section~\ref{sec:compare} 
			when $N$ ranges from 20 to 320.
		}
		\label{tab:sim-N-compare-std}
	\end{table}
	
	\begin{table}[ht]
		\centering
		\begin{tabular}{l|c|rrrrr}
			\toprule
			Standard deviation & Estimator & 20   & 40   & 80   & 160  & 320  \\
			\midrule
			\multirow{4}{*}{Bias}
			& REG-M     & 0.007 & 0.005 & 0.004 & 0.003 & 0.002 \\ 
			& DRL-M     & 0.007 & 0.005 & 0.004 & 0.003 & 0.002 \\ 
			& MIS-M     & 0.006 & 0.004 & 0.003 & 0.002 & 0.002 \\ 
			& COPE      & 0.006 & 0.004 & 0.003 & 0.002 & 0.002 \\ 
			\midrule
			\multirow{4}{*}{MSE}
			& REG-M     & 0.0049 & 0.0031 & 0.0025 & 0.0017 & 0.001 \\ 
			& DRL-M     & 0.0050 & 0.0032 & 0.0026 & 0.0018 & 0.001 \\ 
			& MIS-M     & 0.0020 & 0.0015 & 0.0012 & 0.0008 & 0.0006 \\ 
			& COPE      & 0.0010 & 0.0006 & 0.0004 & 0.0002 & 0.0001 \\ 
			\bottomrule
		\end{tabular}
		\caption{Standard deviations of biases and MSEs of various OPE estimators under settings described in Section~\ref{sec:compare} 
			when $T$ ranges from 20 to 320.
		}\label{tab:sim-T-compare-std}
	\end{table}
	
	\begin{table}[htbp]
		\begin{center}
			\begin{tabular}{l|ccccc}
				\toprule
				& \multicolumn{5}{c}{Quantile}                                 \\
				& 0.05 & 0.25 & 0.50 & 0.75 & 0.95                             \\
				\midrule 
				supply-demand equilibrium metric      &  1.58 &    2.60 &    3.21 &    3.82 &    4.65 \\
				road distance  &  0.57 &    1.62 &    2.30 &    2.99 &    4.00 \\
				reward         &  1.42 &    1.98 &    2.36 &    2.74 &    3.29 \\
				\bottomrule
			\end{tabular}
			\caption{The quantiles of reward and state 
				(including supply-demand equilibrium metric 
				and road distance) 
				in the real data application. 
				Out of privacy concerns, 
				the presented values are scaled.
			}
			\label{tab:real-data-distribution}
		\end{center}
	\end{table}
	
	\section{Technical Definitions and Proofs}

	We begin with some notations. In the proof of Theorems \ref{thm2} and \ref{thm3}, we write $\psi_0$, $\psi_1(O)$, $\psi_2(O)$ and $\psi_3(O)$ as $\psi_0(\widehat{Q},\widehat{p}_m,\widehat{p}_a^*)$, $\psi_1(O;\widehat{Q},\widehat{p}_m,\widehat{p}_a^*,\widehat{\omega})$, $\psi_2(O;\widehat{Q},\widehat{p}_m,\widehat{p}_a^*,\widehat{\omega})$ and $\psi_3(O;\widehat{Q},\widehat{p}_m,\widehat{p}_a^*,\widehat{\omega})$ to highlight their dependence on the nuisance function estimators $\widehat{Q}$, $\widehat{p}_m$, $\widehat{p}_a^*$ and $\widehat{\omega}$. We use $\psi_0^*$, $\psi_1^*$, $\psi_2^*$ and $\psi_3^*$ to denote the corresponding oracle versions with the estimators replaced by their oracle values. Our proofs are generally applicable to either continuous state space or discrete state space with number of elements diverging with the sample size. The action and mediator spaces are assumed to be discrete with finite number of elements. 
	
	The rest of the proof is organized as follows. We first give a detailed definition of $L_2$-norm Convergence. We next present the proofs of our major theorems. Theorem \ref{thm4} can be proven using similar arguments used in the proof of Theorems \ref{thm2} and \ref{thm3}. Hence, we omit the proof of Theorem \ref{thm4} for brevity.

	\subsection{Definition of $L_2$-norm Convergence}\label{sec:l2conv}
	
	We require all nuisance function estimators to converge in $L_2$-norm. Specifically, a sequence of variables $\{X_N\}_{N\ge 0}$ is said to converge in $L_2$-norm to $X$ if and only if $\Mean |X_N-X|^2\to 0$ as $N\to \infty$. Adopting this definition, 
	a Q-estimator $\widehat{Q}$ is said to converge in $L_2$-norm to $Q^{\pi}$ at a rate of $N^{-\alpha}$ if 
	\begin{eqnarray*}
		\sqrt{\Mean_{s\sim p_{\infty}} \Mean |\widehat{Q}(m,a,s)-Q^{\pi}(m,a,s)|^2}=O_p(N^{-\alpha}),
	\end{eqnarray*}
	for any $m$ and $a$. The Q-estimator is said to converge to its oracle value if the right-hand-side (RHS) is $o(1)$. 
	
	Similarly, a density ratio estimator $\widehat{\omega}$ is said to converge in $L_2$-norm to $\omega^{\pi}$ at a rate of $N^{-\alpha}$ if 
	\begin{eqnarray*}
		\sqrt{\Mean_{s\sim p_{\infty}} \Mean |\widehat{\omega}(s)-\omega^{\pi}(s)|^2}=O(N^{-\alpha}).
	\end{eqnarray*}
	$\widehat{\omega}$ is said to converge to its oracle value if the right-hand-side is $o(1)$. 
	
	The estimator $\widehat{p}_m$ is said to converge to $p_m$ at a rate of $N^{-\alpha}$ if  
	\begin{eqnarray*}
		\sqrt{\Mean_{s\sim p_{\infty}}   \Mean|\widehat{p}_m(m|a,s)-p_m(m|a,s)|^2}=O(N^{-\alpha}),
	\end{eqnarray*}
	for any $m$ and $a$. $\widehat{p}_m$ is said to converge to its oracle value is the RHS is $o(1)$. 
	
	Finally, the estimator $\widehat{p}_a$ is said to converge to $p_a^*$ at a rate of $N^{-\alpha}$ if  
	\begin{eqnarray*}
		\sqrt{\Mean_{s\sim p_{\infty}}   \Mean|\widehat{p}_a(a|s)-p_a^*(a|s)|^2}=O(N^{-\alpha}),
	\end{eqnarray*}
	for any $a$. $\widehat{p}_a$ is said to converge to its oracle value is the RHS is $o(1)$.

	\subsection{Proof of Theorem \ref{thm:identifiable}}\label{sec:proofthm1}
	The main idea of the proof is to iteratively apply the front-door adjustment to replace the intervention distribution under the do-calculus with the observed data distribution. 
	
	We first observe that the process $\{(S_t,A_t,M_t,R_t)\}_{t\ge 0}$ satisfies the Markov property, under Assumption 1. This is due to the fact that (i) the distribution of  $(A_t,M_t,R_t)$ is independent of the past observed data history conditional on $S_t$; (ii) the distribution of $S_{t+1}$ is independent of the observed data history up to time $t-1$ conditional on $(S_t,A_t,M_t)$. Similarly, according to the data generating mechanism, we can show that the process generated by assuming the system follows the target policy $\pi$ satisfies the Markov property as well. As such, it follows from \eqref{eqn:dooperator} that 
	\begin{eqnarray*}
		&&\Mean^{\pi}(R_t|S_0=s_0)= \Mean \{R_t|do(A_j=\pi(S_j)),\forall 0\le j\le t,S_0=s_0\}\\&=&\Mean [\Mean \{R_t|do(A_t=\pi(S_t)),S_t\}| do(A_j=\pi(S_j)),\forall 0\le j< t,S_0=s_0].
	\end{eqnarray*}
	We can apply the front-door adjustment formula to represent the inner conditional expectation $\Mean \{R_t|do(A_t=\pi(S_t)),S_t\}$ as
	\begin{eqnarray*}
		\sum_{s_{t+1},r_t,m_t,a_t} r_t p_{s,r}^*(s_{t+1},r_t|m_t,a_t,S_t)p_m(m_t|\pi(S_t),S_t)p_a^*(a_t|S_t).
	\end{eqnarray*}
	As such, 
	\begin{eqnarray*}
		\Mean^{\pi}(R_t|S_0=s_0)=\sum_{s_{t+1},r_t,m_t,a_t} \Mean \left\{r_t p_{s,r}^*(s_{t+1},r_t|m_t,a_t,S_t)p_m(m_t|\pi(S_t),S_t)p_a^*(a_t|S_t)|do(A_j=\pi(S_j)), \right. \\
		\left. \forall 0\le j< t,S_0=s_0\right\}.
	\end{eqnarray*}
	Using the Markov property again, the conditional distribution of $S_t$ given $do(A_j=\pi(S_j)),\forall 0\le j< t,S_0=s$ can be written as 
	\begin{eqnarray*}
		&&\prob(S_t=s_t|do(A_j=\pi(S_j)),\forall 0\le j< t,S_0=s_0)\\
		&=&\Mean \{\prob(S_t=s_t|do(A_{t-1}=\pi(S_{t-1})),S_{t-1})|do(A_j=\pi(S_j)),\forall 0\le j< t-1,S_0=s_0\}.
	\end{eqnarray*}
	Apply the front-door adjustment formula again, the conditional probability in the second line can be rewritten as 
	\begin{eqnarray*}
		\sum_{s_t,r_{t-1},m_{t-1},a_{t-1}} p_{s,r}^*(s_t,r_{t-1}|m_{t-1},a_{t-1},S_{t-1})p_m(m_{t-1}|\pi(S_{t-1}),S_{t-1}) p_a^*(a_{t-1}|S_{t-1}).
	\end{eqnarray*}
	This yields that
	\begin{eqnarray*}
		\Mean^{\pi}(R_t|S_0=s_0)= \sum_{s_{t-1}}\sum_{\{s_{j+1},r_j,m_j,a_j\}_{j=t-1}^t} r_t \left\{\prod_{j=t-1}^t p_{s,r}^*(s_{j+1},r_j|m_j,a_j,s_j)p_m(m_j|\pi(s_j),s_j) p_a^*(a_j|s_j)\right\}\\
		\times  \prob(S_{t-1}=s_{t-1}|do(A_j=\pi(S_j)),\forall 0\le j< t-1,S_0=s_0).
	\end{eqnarray*}
	Repeating the above procedures, we can show that
	\begin{eqnarray*}
		\Mean^{\pi}(R_t|S_0=s_0)=\sum_{\{s_{j+1},r_j,m_j,a_j\}_{j=0}^t} r_t \left\{\prod_{j=0}^t p_{s,r}^*(s_{j+1},r_j|m_j,a_j,s_j)p_m(m_j|\pi(s_j),s_j) p_a^*(a_j|s_j)\right\}.
	\end{eqnarray*}
	Sum $\Mean^{\pi}(R_t|S_0=s_0)$ from $t=0$ to $+\infty$, we obtain that
	\begin{eqnarray*}
		V^{\pi}(s_0)=\sum_{t=0}^{+\infty}\sum_{\{s_{j+1},r_j,m_j,a_j\}_{j=0}^t} \gamma^t r_t \left\{\prod_{j=0}^t p_{s,r}^*(s_{j+1},r_j|m_j,a_j,s_j)p_m(m_j|\pi(s_j),s_j) p_a^*(a_j|s_j)\right\}.
	\end{eqnarray*}
	Integrating over $s_0$ with respect to the initial state distribution $\nu$, we obtain that
	\begin{eqnarray*}
		\eta^{\pi}=\sum_{s_0} \left[\sum_{t=0}^{+\infty}\sum_{\{s_{j+1},r_j,m_j,a_j\}_{j=0}^t} \gamma^t r_t \left\{\prod_{j=0}^t p_{s,r}^*(s_{j+1},r_j|m_j,a_j,s_j)p_m(m_j|\pi(s_j),s_j) p_a^*(a_j|s_j)\right\}\right]\nu(s_0). 
	\end{eqnarray*}
	The proof is thus completed. 
	
	\subsection{Proof of Theorem \ref{thm2}}
	We break the proof into three parts. We first observe that $\widehat{p}_m$ is always consistent under the given conditions. As we have discussed in the main text, this implies that the second augmentation term $(NT)^{-1} \sum_{i,t} \psi_2(O_{i,t};\widehat{Q},\widehat{p}_m,\widehat{p}_a^*,\widehat{\omega})$ will converge to zero. We rigorously prove this claim in Part 1. 
	
	In Part 2, we focus on the case where $\widehat{Q}$, $\widehat{p}_m$ and $\widehat{p}_a^*$ are consistent and show that $(NT)^{-1} \sum_{i,t} \psi_j(O_{i,t};\widehat{Q},\widehat{p}_m,\widehat{p}_a^*,\widehat{\omega})$ converges to zero for $j=1$, $3$, and that $\psi_0(\widehat{Q},\widehat{p}_m,\widehat{p}_a^*)$ is consistent to $\eta^{\pi}$. This together with Part 1 yields the consistency of our estimator.
	
	In Part 3, we consider the case where $\widehat{\omega}$ and $\widehat{p}_m$ are consistent. We focus on showing that for any $Q\in \mathcal{Q}$, $h_a\in \mathcal{H}_a$, 
	\begin{eqnarray*}
		\psi_0(Q,p_m,h_a)+\Mean \psi_3(O;Q,p_m,h_a,\omega^{\pi})=\frac{1}{1-\gamma} \Mean [\omega^{\pi}(S)\rho(M,A,S) \{Q^{\pi}(M,A,S)\\
		-\gamma \sum_{a,a',m}Q^{\pi}(m,a,S')p_m(m,a',S')\pi(a'|S')h_a(a|S')\}],
	\end{eqnarray*}
	for a given data tuple $O=(S,A,M,R,S')$. As discussed in the main text, this further implies that the estimating function is unbiased to the IS estimator with correctly specified $\omega^{\pi}$, $p_m$ and is thus unbiased to $\eta^{\pi}$. Applying similar arguments in used in Part 1 and Part 2, we can show that the proposed estimator is consistent. The proof is thus completed.

	We next detail the proof for each part. 
	
	\textbf{Part 1.} We decompose $(NT)^{-1} \sum_{i,t} \psi_2(O_{i,t};\widehat{Q},\widehat{p}_m,\widehat{p}_a^*,\widehat{\omega})$ into
	\begin{eqnarray*}
		\underbrace{\frac{1}{NT} \sum_{i,t} \{\psi_2(O_{i,t};\widehat{Q},\widehat{p}_m,\widehat{p}_a^*,\widehat{\omega})-\psi_2(O_{i,t};\widehat{Q},p_m,\widehat{p}_a^*,\widehat{\omega})\}}_{\eta_1}+	\underbrace{\frac{1}{NT} \sum_{i,t}\psi_2(O_{i,t};\widehat{Q},p_m,\widehat{p}_a^*,\widehat{\omega})}_{\eta_2}.
	\end{eqnarray*}
	It suffices to show both $\eta_1$ and $\eta_2$ converge to zero in probability. 
	
	Consider $\eta_1$ first. Under the assumptions that $\Omega,\mathcal{Q},\mathcal{H}_m,\mathcal{H}_a$ are bounded function classes and $p_a^*(A_{i,t}|S_{i,t})$ is almost surely bounded away from zero, $|\eta_1|$ can be upper bounded by
	\begin{eqnarray}\label{eqn1}
		\frac{O(1)}{NT} \sum_{i,t} \sum_m |\widehat{p}_m(m|A_{i,t},S_{i,t})-p_m(m|A_{i,t},S_{i,t})|,
	\end{eqnarray}
	where $O(1)$ denotes some positive constant. We aim to show \eqref{eqn1} converge to zero in probability. By Markov's inequality, it suffices to show
	\begin{eqnarray}\label{eq2}
		\frac{1}{NT}\Mean\Big\{ \sum_{i,t} \sum_m |\widehat{p}_m(m|A_{i,t},S_{i,t})-p_m(m|A_{i,t},S_{i,t})|\Big\}=o(1).
	\end{eqnarray}
	Since $\widehat{p}_m$ depends on the observed data as well, the left-hand-side (LHS) is in general not equal to $(NT)^{-1}\sum_{i,t} \sum_m \Mean_{(a,s)\sim p_{\infty}}\Mean|\widehat{p}_m(m|a,s)-p_m(m|a,s)|$. To prove \eqref{eq2}, for any sufficient small constant $\varepsilon>0$, we define a set of functions $\mathcal{H}_m(\varepsilon)$ that contains all mass functions $p$ such that
	\begin{eqnarray*}
		\Mean_{(a,s)\sim p_{\infty}}  \sum_m |p(m|a,s)-p_m(m|a,s)|^2\le \varepsilon^2.
	\end{eqnarray*}
	Under the assumption that $\widehat{p}_m$ converges to $p_m$ in total variation norm, we have by Markov's inequality that $\widehat{p}_m\in \mathcal{H}_m(\varepsilon)$ with probability approaching 1 (wpa1), for sufficiently large $N$. As such, LHS of \eqref{eq2} can be upper bounded by
	\begin{eqnarray}\label{eqn3}
		\frac{1}{NT}\Mean\sup_{p\in \mathcal{H}_m(\varepsilon)} \Big\{\sum_{i,t} \sum_m |p(m|A_{i,t},S_{i,t})-p_m(m|A_{i,t},S_{i,t})|\Big\},
	\end{eqnarray}
	wpa1. We apply the empirical process theory \citep{van1996weak} to bound \eqref{eqn3}. Specifically, we further decompose \eqref{eqn3} as
	\begin{eqnarray}\label{eqn4}
		\eta_3 +\frac{1}{NT}\sup_{p\in \mathcal{H}_m(\varepsilon)} \Big\{\Mean \sum_{i,t} \sum_m |p(m|A_{i,t},S_{i,t})-p_m(m|A_{i,t},S_{i,t})|\Big\}.
	\end{eqnarray}
	where
	\begin{eqnarray*}
		\eta_3 = \frac{1}{NT}\Mean\sup_{p\in \mathcal{H}_m(\varepsilon)} \Big\{\sum_{i,t} \sum_m |p(m|A_{i,t},S_{i,t})-p_m(m|A_{i,t},S_{i,t})|\\
		-\Mean \sum_{i,t} \sum_m |p(m|A_{i,t},S_{i,t})-p_m(m|A_{i,t},S_{i,t})| \Big\}.
	\end{eqnarray*}
	By the definition of $\mathcal{H}_m(\epsilon)$ and Cauchy-Schwarz inequality, the second line is upper bounded by $\varepsilon$ and can be made arbitrarily small by setting $\varepsilon\to 0$. It suffices to show $\eta_3=o(1)$. 
	
	Applying the maximal inequality specialized to the VC-type class \citep[see e.g., Corollary 5.1 in][]{chernozhukov2014gaussian}, $\eta_3$ can be upper bounded by $CN^{-1/2}\sqrt{v}\epsilon\{\log (\epsilon)+\log N\}$ for some constant $C>0$. Under the assumption that $v=O(N^{\kappa})$ for some $\kappa<1$, $\eta_3$ converges to zero in probability. This yields that $\eta_1=o_p(1)$. 
	
	It remains to show $\eta_2=o_p(1)$. As we have discussed in the main text, for any $Q\in \mathcal{Q}$, $\omega\in \Omega$ and $h_a\in \mathcal{H}_a$, $\psi_2(O_{i,t};Q,p_m,h_a,\omega)$ has zero mean. We thus upper bound $\Mean |\eta_2|$ by
	\begin{eqnarray}\label{eqn5}
		\Mean \sup_{Q\in \mathcal{Q},h_a\in \mathcal{H}_a,\omega\in \Omega}\left|\frac{1}{NT} \sum_{i,t}\psi_2(O_{i,t};Q,p_m,h_a,\omega)\right|,
	\end{eqnarray}
	and apply the maximal inequality again to bound \eqref{eqn5}. Specifically, using similar arguments in bounding the first line of \eqref{eqn4}, we can show that \eqref{eqn5} can be upper bounded by $O(N^{-1/2} \log N)$. This implies that $\eta_2=o_p(1)$. The proof for Part 1 is thus completed. 
	
	\textbf{Part 2.} As we have discussed in the main text, when $\widehat{Q}$, $\widehat{p}_m$ and $\widehat{p}_a$ are replaced with their oracle values, the estimating functions $\psi_1(O)$ and $\psi_3(O)$ have zero means. Using similar arguments in showing $\eta_2=o_p(1)$ in Part 1, we can show that $(NT)^{-1}\sum_{i,t} \psi_1(O_{i,t};Q,p_m,p_a^*,\widehat{\omega})=o_p(1)$ and that
	\begin{equation*}
		(NT)^{-1}\sum_{i,t} \psi_3(O_{i,t};Q,p_m,p_a^*,\widehat{\omega})=o_p(1). 
	\end{equation*}
	In addition, using similar arguments in showing $\eta_1=o_p(1)$ in Part 1, we can show that 
	\begin{equation*}
		(NT)^{-1}\sum_{i,t} \{\psi_1(O_{i,t};\widehat{Q},\widehat{p}_m,\widehat{p}_a^*,\widehat{\omega})-\psi_1(O_{i,t};Q,p_m,p_a^*,\widehat{\omega})\}=o_p(1)
	\end{equation*}
	and 
	\begin{equation*}
		(NT)^{-1}\sum_{i,t} \{\psi_3(O_{i,t};\widehat{Q},\widehat{p}_m,\widehat{p}_a^*,\widehat{\omega})-\psi_3(O_{i,t};Q,p_m,p_a^*,\widehat{\omega})\}=o_p(1).
	\end{equation*}
	This yields that 
	\begin{equation*}
		(NT)^{-1}\sum_{i,t} \psi_1(O_{i,t};\widehat{Q},\widehat{p}_m,\widehat{p}_a^*,\widehat{\omega})=o_p(1)
	\end{equation*}
	and
	\begin{equation*}
		(NT)^{-1}\sum_{i,t} \psi_3(O_{i,t};\widehat{Q},\widehat{p}_m,\widehat{p}_a^*,\widehat{\omega})=o_p(1).
	\end{equation*}
	
	It remains to show $\psi_0(\widehat{Q},\widehat{p}_m,\widehat{p}_a^*)$ is consistent to $\eta^{\pi}$. Since the action and mediator spaces have finitely many elements, it suffices to show
	\begin{eqnarray}\label{eqn6}
		\Mean_{s\sim \widehat{\nu}} \widehat{Q}(m,a,s)\widehat{p}_m(m,a',s)\pi(a'|s)\widehat{p}_a^*(a|s)\stackrel{P}{\to} 	\Mean_{s\sim \nu} Q^{\pi}(m,a,s)p_m(m,a',s)\pi(a'|s)p_a^*(a|s),
	\end{eqnarray} 
	for any $a,a'$ and $m$, where $\widehat{\nu}$ denotes the initial state distribution. Under the assumption that the process $\{S_t\}_{t\ge 0}$ is stationary, we have $\nu=p_{\infty}$. Applying similar arguments in Part 1, we can show that the difference
	\begin{eqnarray*}
		\Mean_{s\sim \widehat{\nu}} \widehat{Q}(m,a,s)\widehat{p}_m(m,a',s)\pi(a'|s)\widehat{p}_a^*(a|s)-\Mean_{s\sim \widehat{\nu}} Q^{\pi}(m,a,s)p_m(m,a',s)\pi(a'|s)p_a^*(a|s)
	\end{eqnarray*}
	will converge in probability to zero. 
	
	Recall that we set $\widehat{\nu}$ to the initial state distribution. Applying the weak law of large numbers again, we have as $N\to \infty$ that 
	\begin{eqnarray*}
		\Mean_{s\sim \widehat{\nu}} Q^{\pi}(m,a,s)p_m(m,a',s)\pi(a'|s)p_a^*(a|s)\stackrel{P}{\to}\Mean_{s\sim \nu} Q^{\pi}(m,a,s)p_m(m,a',s)\pi(a'|s)p_a^*(a|s).
	\end{eqnarray*}
	This yields \eqref{eqn6}. The proof for Part 2 is hence completed. 
	
	\textbf{Part 3}. Since $p_m$ is correct, we have
	\begin{eqnarray*}
		\Mean \omega^{\pi}(S)\rho(M,A,S)Q^{\pi}(M,A,S)=\sum_{m,a',a,S}\Mean \omega^{\pi}(S) p(m|a',S)\pi(a'|S)p_a^*(a|S)Q^{\pi}(m,a,S).
	\end{eqnarray*}
	It follows from the definition of $\psi_3$ that
	\begin{eqnarray*}
		\Mean \psi_3(O;Q,p_m,h_a,\omega^{\pi})-\frac{1}{1-\gamma} \Mean \omega^{\pi}(S)\rho(M,A,S) Q(M,A,S)\\=-\frac{1}{1-\gamma}\sum_{m,a',a} \Mean \omega^{\pi}(S) p_m(m|a',S)\pi(a'|S)Q(m,a,S)h_a(a|S).
	\end{eqnarray*}
	As such, it suffices to show
	\begin{eqnarray}\label{eqn7}
		\begin{split}
			\frac{1}{1-\gamma}\sum_{m,a',a} \Mean \omega^{\pi}(S) p_m(m|a',S)\pi(a'|S)Q(m,a,S)h_a(a|S)-\psi_0(Q,p_m,h_a)\\
			=\frac{\gamma}{1-\gamma} \Mean \omega^{\pi}(S)\rho(M,A,S)  \sum_{a,a',m}Q^{\pi}(m,a,S')p_m(m,a',S')\pi(a'|S')h_a(a|S')\}].
		\end{split}	
	\end{eqnarray}
	Due the presence of the importance sampling ratio $\rho(M,A,S)$, the second line is equal to
	\begin{eqnarray*}
		\frac{\gamma}{1-\gamma} \Mean \omega^{\pi}(S) \Mean \left[\left.\sum_{a,a',m}Q^{\pi}(m,a,S')p_m(m,a',S')\pi(a'|S')h_a(a|S')\right|do(A=\pi(S)) \right].
	\end{eqnarray*}
	By the definition of $\omega^{\pi}(S)$, both lines in \eqref{eqn7} are equal to
	\begin{eqnarray*}
		\sum_{t\ge 1}\gamma^t \sum_{s,a,a',m} p_t^{\pi}(s)Q^{\pi}(m,a,s)p_m(m,a',s)\pi(a'|s)h_a(a|s).
	\end{eqnarray*}
	The proof is hence completed. 
	
	\subsection{Proof of Theorem \ref{thm3}}\label{sec:proofthm3}
	The proof is divided into two parts. In the Part 1, we show that the proposed estimator is asymptotically equivalent to the oracle estimator $\widehat{\eta}^*=\psi_0^*+(NT)^{-1} \sum_{j=1}^3 \sum_{i,t} \psi_j^*(O_{i,t})$ with the difference upper bounded by $o_p(N^{-1/2})$. In the second part, we show the oracle estimator is asymptotically normal and satisfies that $\sqrt{N}(\widehat{\eta}^*-\eta^{\pi})\stackrel{d}{\to} N(0,\sigma^2_T)$ where $\sigma^2_T$ is the semiparametric efficiency bound. By Slutsky's theorem, the proposed estimator is asymptotically normal and achieves the semiparametric efficiency bound as well. This completes the proof.
	
	Before detailing each part of the proof, we make some remarks. First, it follows from the Slutsky's theorem that the proposed estimator is asymptotically normal and satisfies $\sqrt{N}(\widehat{\eta}-\eta^{\pi})\stackrel{d}{\to}N(0,\sigma_T^2)$ as well. Second, a key observation is that, for any $j=1,2,3$, $\sum_{t} \psi_j^*(O_{i,t})$ corresponds to a sum of the martingale difference sequence with respect to the filtration, the $\sigma$-algebra generated by $\{O_{i,l}\}_{0\le l\le t-1}$. Under the stationarity assumption, we have
	\begin{eqnarray*}
		\sigma_T^2=\Var\Big\{\sum_{m,a,a',S_0}Q^{\pi}(m,a,S_0)p_m(m,a',S_0)\pi(a'|S_0)p_a^*(a|S_0)\Big\}+\frac{1}{T}\Var\Big\{\sum_{j=1}^3 \psi_j^*(O_0)\Big\},
	\end{eqnarray*}
	where the first term on the right-hand-side is due to the variation of the initial state distribution in the plug-in estimator $\psi_0^*$, the second term is due to the variation of the three augmentation terms. Consequently, $\sigma_T^2$ decreases with $T$. As $T\to \infty$, $\sigma_T^2$ converges to the first term, 
	\begin{eqnarray*}
		\Var\Big\{\sum_{m,a,a',S_0}Q^{\pi}(m,a,S_0)p_m(m,a',S_0)\pi(a'|S_0)p_a^*(a|S_0)\Big\}.
	\end{eqnarray*}
	
	\textbf{Part 1.} We provide a sketch of the proof to save brevity. We decompose the difference between the proposed value estimator and the oracle estimator as $\widehat{\eta}^{(1)}(\widehat{Q},\widehat{p}_a^*)+\widehat{\eta}^{(2)}(\widehat{Q},\widehat{p}_a^*)$ where 
	\begin{eqnarray*}
		\widehat{\eta}^{(1)}(\widehat{Q},\widehat{p}_a^*)&=&\psi_0(\widehat{Q},p_m,\widehat{p}_a^*)-\psi_0^*+\frac{1}{NT}\sum_{i,t} \sum_{j=1}^3 \{\psi_j(O_{i,t};\widehat{Q},p_m,\widehat{p}_a^*,\omega^{\pi})-\psi_j^*(O_{i,t})\},\\
		\widehat{\eta}^{(2)}(\widehat{Q},\widehat{p}_a^*)&=&\{\psi_0(\widehat{Q},\widehat{p}_m,\widehat{p}_a^*)-\psi_0(\widehat{Q},p_m,\widehat{p}_a^*)\}
		\\
		&+&\frac{1}{NT}\sum_{i,t} \sum_{j=1}^3 \{\psi_j(O_{i,t};\widehat{Q},\widehat{p}_m,\widehat{p}_a^*,\widehat{\omega})-\psi_j(O_{i,t};\widehat{Q},p_m,\widehat{p}_a^*,\omega^{\pi})\}.
	\end{eqnarray*}
	Consider $\widehat{\eta}^{(1)}(\widehat{Q},\widehat{p}_a^*)$ first. As discussed in Part 3 of the proof of Theorem \ref{thm2}, for any $Q\in \mathcal{Q}$ and $h_a\in \mathcal{H}_a$, we have $\Mean \widehat{\eta}^{(1)}(Q,h_a)=0$. Using similar arguments in the proof of Theorem \ref{thm2}, we can apply empirical process theory to show that $\widehat{\eta}^{(1)}(\widehat{Q},\widehat{p}_a^*)=o_p(N^{-1/2})$, under the conditions that $\widehat{Q}$ and $\widehat{p}_a^*$ converge in $L_2$-norm to their oracle values. 
	
	Next we consider $\widehat{\eta}^{(2)}(\widehat{Q},\widehat{p}_a^*)$. Similar to $\widehat{\eta}^{(1)}$, we can show that $\widehat{\eta}^{(2)}(Q^{\pi},p_a^*)=o_p(N^{-1/2})$. It remains to show $\widehat{\eta}^{(2)}(\widehat{Q},\widehat{p}_a^*)-\widehat{\eta}^{(2)}(Q^{\pi},p_a^*)=o_p(N^{-1/2})$. The difference $\widehat{\eta}^{(2)}(\widehat{Q},\widehat{p}_a^*)-\widehat{\eta}^{(2)}(Q^{\pi},p_a^*)$ can be decomposed into the sum
	\begin{eqnarray*}
		\{\psi_0(\widehat{Q},\widehat{p}_m,\widehat{p}_a^*)-\psi_0(\widehat{Q},p_m,\widehat{p}_a^*)\}-[\{\psi_0(Q^{\pi},\widehat{p}_m,p_a^*)-\psi_0(Q^{\pi},p_m,p_a^*)\}]\\
		+\frac{1}{NT}\sum_{i,t} \sum_{j=1}^3 [\{\psi_j(O_{i,t};\widehat{Q},\widehat{p}_m,\widehat{p}_a^*,\widehat{\omega})-\psi_j(O_{i,t};\widehat{Q},p_m,\widehat{p}_a^*,\omega^{\pi})\}\\
		-\{\psi_j(O_{i,t};Q^{\pi},\widehat{p}_m,p_a^*,\widehat{\omega})-\psi_j(O_{i,t};Q^{\pi},p_m,p_a^*,\omega^{\pi})\}].
	\end{eqnarray*}
	To save space, we focus on proving that the first line is $o_p(N^{-1/2})$. Using similar arguments, one can show that the second line is $o_p(N^{-1/2})$ as well. This completes the proof of this part. 
	
	With some calculations, we can show that the absolute value of the first line can be further bounded by $\eta_4+\eta_5$ where
	\begin{eqnarray*}
		\eta_4=\frac{1}{N}\sum_{i=1}^N \sum_{m,a',a}|\widehat{Q}(m,a',S_{i,0})| \pi(a'|S_{i,0}) |\widehat{p}_a^*(a|S_{i,0})-p_a^*(a|S_{i,0})|  |\widehat{p}_m^*(m|a,S_{i,0})-p_m(m|a,S_{i,0})|, \\
		\eta_5 = \frac{1}{N}\sum_{i=1}^N \sum_{m,a',a}|\widehat{Q}(m,a',S_{i,0})-Q^{\pi}(m,a',S_{i,0})| \pi(a'|S_{i,0}) p_a^*(a|S_{i,0})  |\widehat{p}_m^*(m|a,S_{i,0})-p_m(m|a,S_{i,0})|.
	\end{eqnarray*}
	Under the condition that $\mathcal{Q}$ is a bounded function class, $\eta_4$ can be upper bounded by
	\begin{eqnarray*}
		\frac{C}{N}\sum_{i=1}^N \sum_{m,a}  |\widehat{p}_a^*(a|S_{i,0})-p_a^*(a|S_{i,0})| |\widehat{p}_m^*(m|a,S_{i,0})-p_m(m|a,S_{i,0})|,
	\end{eqnarray*}
	for some constant $C>0$. 
	
	Applying the Cauchy-Schwarz inequality, it can be further upper bounded by
	\begin{eqnarray}\label{eqn8}
		\frac{C}{2N}\sum_{i=1}^N \sum_{m,a}  |\widehat{p}_a^*(a|S_{i,0})-p_a^*(a|S_{i,0})|^2+	\frac{C}{2N}\sum_{i=1}^N \sum_{m,a} |\widehat{p}_m^*(m|a,S_{i,0})-p_m(m|a,S_{i,0})|^2.
	\end{eqnarray}	
	Using similar arguments in bounding \eqref{eqn3}, we can show that \eqref{eqn8} is $o_p(N^{-1/2})$, under the condition that $\widehat{p}_m$ and $\widehat{p}_a^*$ converge in $L_2$-norm to their oracle values at a rate of $N^{-\kappa^*}$ for some $\kappa^*>1/4$. 
	
	\textbf{Part 2.} The asymptotic normality of $\sqrt{N}(\widehat{\eta}^*-\eta^{\pi})$ follows from the classical central limit theorem. It suffices to show  $\sigma_T^2$ achieves the minimum variance among all regular and asymptotically linear estimators. 
	
	Specifically, let $p_{s,r,\theta}$, $p_{m,\theta}$, $p_{a,\theta}$ and $\nu_{\theta}$ be the some parametric models for $p_{s,r}$, $p_m$, $p_a$ and $\nu$, respectively. Let $\mathcal{M}$ denote the set of these models. We require the existence of some oracle parameter $\theta_0$ such that $p_{s,r,\theta_0}$, $p_{m,\theta_0}$, $p_{a,\theta_0}$ and $\nu_{\theta_0}$ corresponding to the true models.  In the proof, we redefine the marginalized density ratio $\omega^{\pi}(\bullet)$ as $(1-\gamma) \sum_{t\ge 0}  \gamma^t \frac{p_t^{\pi}(s)}{p_{D}(s)}$, where $p_t^{\pi}(s)$ denotes the probability of $S_t=t$ by assuming the system follows $\pi$, and $p_{D}$ denotes the mixture distribution of observed data, i.e.
	$p_{D}(s) = \frac{1}{T}\sum_{t=0}^{T-1} p_{t}^b(s)$.
	Note that when the stochastic process is stationary, $p_{D}(s) = p_{\infty}(s)$. Similar to Theorem \ref{thm:identifiable}, we can show for any $t>1$, $p_t^{\pi}$ is identifiable. So is $\omega^{\pi}$.

	Theorem \ref{thm:identifiable} allows us to express the value as a function of $\theta$ (denote by $\eta^{\pi}_{\theta}$). 
	Define the Cramer-Rao bound as
	\begin{eqnarray*}
		\textrm{CR}(\mathcal{M})=\frac{\partial \eta_{\theta}^{\pi}}{\partial \theta_0} \left(\Mean \frac{\partial \ell(\{O_t\}_{0\le t<T};\theta)}{\partial \theta_0} \frac{\partial \ell^\top(\{O_t\}_{0\le t<T};\theta)}{\partial \theta_0}\right)^{-1} \left(\frac{\partial \eta_{\theta}^{\pi}}{\partial \theta_0}\right)^{\top},
	\end{eqnarray*}
	where $\ell(\{O_t\}_{0\le t<T};\theta)$ denote the log-likelihood function
	\begin{eqnarray*}
		\sum_{t=0}^{T-1} \log p_{s,r,\theta}(S_{t+1},R_t|M_t,A_t,S_t)+\sum_{t=0}^{T-1}\log p_{m,\theta}(M_t|A_t,S_t)+\sum_{t=0}^{T-1} \log p_{a,\theta}(A_t|S_t)+\log \nu_{\theta}(S_0).
	\end{eqnarray*}
	We aim to show
	\begin{eqnarray*}
		\sigma_T^2=\sup \textrm{CR}(\mathcal{M}),
	\end{eqnarray*}
	where supremum is taken over all regular parametric submodels $\mathcal{M}$. 
	
	We outline a sketch of the proof. In the following, we focus on proving that $\partial \eta_{\theta}^{\pi}/\partial \theta_0$ can be represented as 
	\begin{equation}\label{eqn:influence}
		\begin{aligned}
			\frac{\partial \eta_{\theta}^{\pi}}{\partial \theta_0}=\Mean \left[\left\{\sum_{m,a,a',S_0}Q^{\pi}(m,a,S_0)p_m(m,a',S_0)\pi(a'|S_0)p_a^*(a|S_0) -\eta^{\pi}  \right.\right. \\ \left.\left.
			+\frac{1}{T}\sum_{j=1}^3\sum_{t=0}^{T-1}\psi_j^*(O_t) \right\}\frac{\partial \ell(\{O_t\}_{0\le t<T};\theta)}{\partial \theta_0}\right].
		\end{aligned}
	\end{equation}
	By Cauchy-Schwarz inequality, we obtain that
	\begin{eqnarray*}
		\sup_{\mathcal{M}} \textrm{CR}(\mathcal{M})\le \Mean \left\{\sum_{m,a,a',S_0}Q^{\pi}(m,a,S_0)p_m(m,a',S_0)\pi(a'|S_0)p_a^*(a|S_0)-\eta^{\pi}+\frac{1}{T}\sum_{j=1}^3\sum_{t=0}^{T-1}\psi_j^*(O_t)\right\}^2.
	\end{eqnarray*} 
	where the RHS equals $\sigma_T^2$.

	Using similar arguments in Lemma 20 of \citet{kallus2019efficiently}, we can show that there exists some parametric model $\mathcal{M}_0$ with diverging number of parameters such that $\sigma_T^2= \textrm{CR}(\mathcal{M}_0)$. The proof is hence completed.

	We next calculate the gradient of $\eta^{\pi}_{\theta}$ with respect to $\theta$, evaluated at $\theta=\theta_0$. For simplicity, we assume $\pi$ is a deterministic policy. It follows from Theorem \ref{thm:identifiable} that 
	\begin{equation*}
		\eta^{\pi}_{\theta_0} = \left[ \sum\limits_{t=0}^{+\infty}\gamma^t \sum\limits_{\tau_t,s_{t+1}} r_t \left\{\prod_{j=0}^t p_{\theta_0}(s_{j+1},r_j,m_j,a_j|s_j)\right\}\nu_{\theta_0}(s_0)\right],
	\end{equation*}
	where $p_{\theta}$ is the product of probability mass functions $p_{s,r,\theta}^*(s,r|m,a,s)p_m(m|\pi(s),s) p_a^*(a|s)$.  
	
	Using the fact that $\nabla_\theta p_{\theta} = p_{\theta} \nabla_{\theta} \log(p_{\theta})$, we can further write $\nabla_{\theta} \eta_{\theta_0}^{\pi} $ as 
	\begin{equation} \label{EqGradient}
		\begin{aligned}
			& = \sum\limits_{t=0}^{+\infty} \gamma^{t} \sum_{\tau_t,s_{t+1}} r_{t} \prod_{j=0}^{t} p_{\theta_0}\left(s_{j+1}, r_{j}, m_{j}, a_{j} \mid s_{j}\right)\left\{\sum\limits_{j=0}^{t}\nabla_{\theta} \log p_{\theta_0}\left(s_{j+1}, r_{j}, m_{j}, a_{j} \mid s_{j}\right) \right\}
			\nu_{\theta_0}\left(s_{0}\right)  \\
			& +   \sum_{t=0}^{+\infty}\gamma^t \sum_{\tau_t,s_{t+1}} r_t \left\{\prod_{j=0}^t p_{\theta_0}(s_{j+1},r_j,m_j,a_j|s_j)\right\}\nu_{\theta_0}(s_0) \nabla_{\theta} \log\nu_{\theta_0}(s_0).
		\end{aligned}
	\end{equation}
	The first line in equation \ref{EqGradient} can be written as 
	\begin{eqnarray} \label{EqGradientP1}
		\begin{split}
			\sum_{s_0}\sum_{j=0}^{+\infty} \gamma^j \sum_{\tau_{j},s_{j+1}} \{r_j+V^{\pi}(s_{j+1})\} \left\{\prod_{k=0}^j {\color{black}p_{\theta_0}(s_{k+1},r_k,m_k,a_k|s_k)}\right\}    \\
			\times {\color{black}\nabla_{\theta}} \log p_{\theta_0}(s_{j+1},r_j,m_j,a_j|s_j)\nu_{\theta_0}(s_0).	
		\end{split}	
	\end{eqnarray}
	See section \ref{ProofOfGradient} for a detailed proof. 
	
	We can further decompose equation   \ref{EqGradientP1} into three parts and express it as $I_{1} + I_{2} +I_{3}$, where 
	\begin{equation*}
		I_j = \Mean \left[\left\{  \frac{1}{T} \sum\limits_{t=0}^{T-1}  \psi_j^*(O_t)\right\}   S(\bar{O}_{T-1})\right], j = 1, 2, 3,
	\end{equation*}
	where $\bar{O}_T$ denotes the observed data history $\{O_0,O_1,\cdots,O_{T-1}\}$ up to time $T$ and $S(\cdot)$ denotes the score function, i.e., the gradient with respect to the log-likelihood function evaluated at $\theta=\theta_0$. See section \ref{DerivationOfI1} for a detailed proof.
	
	The second line in equation \ref{EqGradient} is equal to $ 
	\Mean\left[ V^{\pi}\left(S_{0}\right) \nabla_{\theta} \log\nu_{\theta_0}(S_0)\right]$.
	Since the expectation of a score function is zero, we have $ \Mean\left[ \eta^{\pi}_{\theta_0}\nabla_{\theta} \log\nu_{\theta_0}(S_0) \right] = \eta^{\pi}_{\theta_0}\Mean\left[ \nabla_{\theta} \log\nu_{\theta_0}(S_0) \right] = 0$. It follows that 
	\begin{equation*}
		\Mean\left[ V^{\pi}\left(S_{0}\right) \nabla_{\theta} \log\nu_{\theta_0}(S_0) \right]=\Mean\left[ \left\{ V^{\pi}\left(S_{0}\right) - \eta^{\pi}_{\theta_0} \right\} \nabla_{\theta} \log\nu_{\theta_0}(S_0) \right].
	\end{equation*}
	Using the same trick again, the above quantity is equal to
	\begin{eqnarray*}
		I_4=\Mean  \left\{ V^{\pi}\left(S_{0}\right) - \eta^{\pi} \right\} S(\bar{O}_{T-1}).
	\end{eqnarray*}
	Therefore we can represent $\nabla_{\theta} \eta_{\theta_0}^{\pi}$ as 
	\begin{equation*}
		\nabla_{\theta} \eta_{\theta_0}^{\pi} =  I_1+I_2+I_3+I_4=\Mean \left[\left\{  \frac{1}{T} \sum\limits_{t=0}^{T-1}  \psi_j^*(O_t)+V^{\pi}(S_0)-\eta^{\pi}\right\}   S(\bar{O}_{T-1})\right].
	\end{equation*}
	Applying the front-door adjustment formula, $V^{\pi}(S_0)$ equals 
	\begin{equation*}
		\sum_{m,a,a'} Q^{\pi}(m,a,S_0)p_m(m|a',S_0) p_a^*(a|S_0)\pi(a'|S_0). 
	\end{equation*}

	As such, $\nabla_{\theta} \eta_{\theta_0}^{\pi}$ takes the following form
	\begin{eqnarray*}
		\Mean \left[\left\{\frac{1}{T}\sum_{t=0}^{T-1} \psi_j^*(O_j)+\sum_{m,a,a'} Q^{\pi}(m,a,S_0)p_m(m|a',S_0) p_a^*(a|S_0)\pi(a'|S_0)-\eta^{\pi}\right\}S(\bar{O}_{T-1})\right].
	\end{eqnarray*} 
	This completes the proof. 
	
	
	\subsection{Proof of equation \ref{EqGradientP1} } \label{ProofOfGradient}
	In this subsection, we will prove  that the first line in equation \ref{EqGradient} can be written as equation \ref{EqGradientP1}. Firstly, we exchange the summation of $t$ and $j$ in the first line in equation \ref{EqGradient} , which yields
	\begin{equation*}
		\sum\limits_{s_{0}}  \sum\limits_{j=0}^{+\infty} \left\{ \sum\limits_{t=j}^{+\infty}   \gamma^{t}  \sum\limits_{\tau_{t}, s_{t+1}}  r_{t}   \left [ \prod_{k=0}^{t} p_{\theta_{0}}\left(s_{k+1}, r_{k}, m_{k}, a_{k} \mid s_{k}\right)\right]\right\} \nabla_{\theta} \log p_{\theta_{0}}\left(s_{j+1}, r_{j}, m_{j}, a_{j} \mid s_{j}\right) \nu_{\theta_0}\left(s_{0}\right).
	\end{equation*}
	Note that we have changed the subscript of the product to avoid confusion. Then we split the summation $\sum_{t=j}^{+\infty}$ into the sum of $t=j$ and $\sum_{t=j+1}^{+\infty}$, and split the product $\prod_{k=0}^t$ into the product of $\prod_{k=0}^j$ and $\prod_{k=j+1}^t$, we obtain that  the first line in equation \ref{EqGradient} equals
	
	\begin{equation*}
		\begin{aligned}
			& \sum\limits_{s_{0}}  \sum\limits_{j=0}^{+\infty}\gamma^{j}     \sum\limits_{\tau_{t}, s_{t+1}}    \left [ \prod_{k=0}^{j} p_{\theta_{0}}\left(s_{k+1}, r_{k}, m_{k}, a_{k} \mid s_{k}\right)\right]  \left\{  r_{j}  +   \sum\limits_{t=j+1}^{+\infty}  \gamma^{t-j} \sum\limits_{ \tau_{t/j}, s_{t+1}} r_{t}  \right. \\
			&   \left. \times \left [  \prod_{k=j+1}^{t} p_{\theta_{0}}\left(s_{k+1}, r_{k}, m_{k}, a_{k} \mid s_{k}\right) \right]  \right\} \nabla_{\theta}  \log p_{\theta_{0}}\left(s_{j+1}, r_{j}, m_{j}, a_{j} \mid s_{j}\right) \nu_{\theta_0}\left(s_{0}\right), \\
		\end{aligned}
	\end{equation*}
	
	where $\tau_{t/j}$ is the transition from time $j$ to time $t$. 
	By the front-door adjustment formula, we have 
	\begin{equation*}
		\begin{aligned}
			\sum\limits_{m_{t},a_{t}}p_{s,r,\theta_{0}}^*(s_{t+1},r_t|m_t,a_t,s_t) p_{m,\theta_{0}}(m_t|\pi(s_t),s_t)p_{a,\theta_{0}}^*(a_t|s_t) =  p_{s,r,\theta_{0}}^*(s_{t+1}, r_t\mid do \left(\pi\left(s_{t}\right)\right),s_t).
		\end{aligned}
	\end{equation*}
		Substitute this equation, $\sum\limits_{t=j+1}^{+\infty}  \gamma^{t-j} \sum\limits_{ \tau_{t/j}, s_{t+1}} r_{t}  \left [  \prod_{k=j+1}^{t} p_{\theta_{0}}\left(s_{k+1}, r_{k}, m_{k}, a_{k} \mid s_{k}\right) \right]$ is equal to 
		\begin{eqnarray*}
			\gamma  \sum_{t=j+1}^{+\infty}\gamma^{t-(j+1)}   \sum\limits_{s_{t/j},r_{t/j}} r_t  p_{s,r,\theta_{0}}^*(s_{t+1},r_t\mid \operatorname{do} \left(\pi\left(s_{t}\right)\right),s_t)   \left[\prod_{k=j+1}^{t-1} p_{s,r,\theta_{0}}(s_{k+1},r_k|do\left(\pi\left(s_{k}\right)\right), s_{k})  \right] \\ = \gamma  \sum_{t=j+1}^{+\infty} \gamma^{t-(j+1)}  \Mean^{\pi} [R_t |S_j=s_j],
		\end{eqnarray*}
		which is $\gamma  V^{\pi}(s_{j+1})$. It follows that the first line in equation \ref{EqGradient} equals
		\begin{eqnarray*}
			\sum\limits_{s_{0}}  \sum\limits_{j=0}^{+\infty}\gamma^{j}     \sum\limits_{\tau_{j},s_{j+1}}    \left [ \prod_{k=0}^{j} p_{\theta_{0}}\left(s_{k+1}, r_{k}, m_{k}, a_{k} \mid s_{k}\right)\right]  \left\{  r_{j}  +   \gamma  V^{\pi}(s_{j+1}) \right\} \\
			\times  \nabla_{\theta}  \log p_{\theta_{0}}\left(s_{j+1}, r_{j}, m_{j}, a_{j} \mid s_{j}\right) \nu_{\theta_0}\left(s_{0}\right),
		\end{eqnarray*}
		which is exactly equation \ref{EqGradientP1}.

		\subsection{Derivations of  $I_1$, $I_2$ and $I_3$} \label{DerivationOfI1}

		Since $\nabla_{\theta} \log p_{\theta_{0}}\left(s_{j+1}, r_{j}, m_{j}, a_{j} \mid s_{j}\right)$ equals to 
		\begin{equation*}
			\nabla_{\theta} \log  p_{s, r, \theta_{0}}\left(s_{t+1}, r_{t} \mid m_{t}, a_{t}, s_{t}\right) + \nabla_{\theta}  \log p_{m, \theta_{0}}\left(m_{t} \mid \pi\left(s_{t}\right), s_{t}\right) + \nabla_{\theta}  \log  p_{a, \theta_{0}}\left(a_{t} \mid s_{t}\right),
		\end{equation*}
		we can write equation \ref{EqGradientP1} as $I_{1} + I_{2} +I_{3}$, where
		\begin{eqnarray} \label{eqn:eqnI1}
			I_1 =  \sum_{s_0}\sum_{j=0}^{+\infty} \gamma^j \sum_{\tau_{j}, s_{j+1}} \{r_j+\gamma  V^{\pi}(s_{j+1})\} \left\{\prod_{k=0}^j {\color{black}p_{\theta_0}(s_{k+1},r_k,m_k,a_k|s_k)}\right\}\\
			\nabla_{\theta} \log p_{s,r,\theta_0 }(s_{j+1},r_j|m_j,a_j,s_j)\nu_{\theta_0}(s_0),
		\end{eqnarray}
		
		\begin{eqnarray}\label{eqn:eqnI2}
			I_2 =  \sum_{s_0}\sum_{j=0}^{+\infty} \gamma^j \sum_{\tau_{j}, s_{j+1}} \{r_j+\gamma  V^{\pi}(s_{j+1})\} \left\{\prod_{k=0}^j {\color{black}p_{\theta_0}(s_{k+1},r_k,m_k,a_k|s_k)}\right\}\\	
			\nabla_{\theta} \log p_{m,\theta
				_0}(m_j|\pi(s_j),s_j)\nu_{\theta_0}(s_0),
		\end{eqnarray}
		
		\begin{eqnarray}\label{eqn:eqnI3}
			I_3  =  \sum_{s_0}\sum_{j=0}^{+\infty} \gamma^j \sum_{\tau_{j}, s_{j+1}} \{r_j+\gamma  V^{\pi}(s_{j+1})\} \left\{\prod_{k=0}^j {\color{black}p_{\theta_0}(s_{k+1},r_k,m_k,a_k|s_k)}\right\} \\ \nabla_{\theta} \log p_{a,\theta_0}(a_j|s_j)\nu_{\theta_0}(s_0).
		\end{eqnarray}

		\subsubsection{Derivations of  $I_1$} 
		
		In the following, we focus on proving equation \ref{eqn:eqnI1}.
		
		First, we note that 
		\begin{equation*}
			\sum_{s_0}\left\{\prod_{k=0}^j {\color{black}p_{\theta_0}(s_{k+1},r_k,m_k,a_k|s_k)}\right\}\nu_{\theta_0}(s_0)
		\end{equation*}
		equalsthe probability of $\{S_{j+1}=s_{j+1}$, $R_j=r_j$, $M_j=m_j$,$A_j=a_j\}$ under target policy $\pi$, i.e., 
		\begin{equation*}
			p_{\theta_0}(s_{j+1},r_j,m_j,a_j|s_j) p_{t,\theta_0}^{\pi}(s_j) = p_{s,r,\theta_0}^*(s_{j+1},r_j|m_j,a_j,s_j)p_{m,\theta_0}(m_j|\pi(s_j),s_j)p_{a,\theta_0}^*(a_j|s_j)p_{t,\theta_0}^{\pi}(s_j).
		\end{equation*}
		Using the fact that the expectation of a score function is zero, we have 
		\begin{align*}
			\sum_{s_{j+1}, r_j} p_{s,r,\theta_0}^*(s_{j+1},r_j|m_j,a_j,s_j)\nabla_{\theta} \log p_{s,r,\theta_0}^*(s_{j+1},r_j|m_j,a_j,s_j) 
		\end{align*}
		is zero for any $m_j,a_j$ and $s_j$. It follows that
		\begin{equation*}
			\begin{aligned}
				& \sum_{j=0}^{+\infty} \gamma^j \sum_{s_{j+1},
					r_j,m_j,a_j,s_j} Q^{\pi}(m_j,a_j,s_j)
				\frac{p_{m,\theta_0}(m_j|\pi(s_j),s_j)}{p_{m,\theta_0}(m_j|a_j,s_j) }  p_{s,r,\theta_
					0}(s_{j+1},r_j|m_j,a_j,s_j)\\
				& \times  p_{m,\theta_0}(m_j|a_j,s_j) p_{a,\theta_0}(a_j|s_j)p_{j,\theta_0}^{\pi}(s_j) \nabla_{\theta} \log p_{s,r,\theta_0}(s_{j+1},r_j|m_j,a_j,s_j)\\
				& =\sum_{j=0}^{+\infty} \gamma^j \sum_{
					m_j,a_j,s_j} Q^{\pi}(m_j,a_j,s_j) 
				\frac{p_{m,\theta_0}(m_j|\pi(s_j),s_j)}{p_{m,\theta_0}(m_j|a_j,s_j) }   p_{m,\theta_0}(m_j|a_j,s_j)  p_{a,\theta_0}(a_j|s_j)p_{j,\theta_0}^{\pi}(s_j)  \\
				& \times\sum_{s_{j+1}, r_j} p_{s,r,\theta_0}(s_{j+1},r_j|m_j,a_j,s_j)\nabla_{\theta} \log p_{s,r,\theta_0}(s_{j+1},r_j|m_j,a_j,s_j) =0.\\
			\end{aligned}
		\end{equation*}
		Thus, we have 
		\begin{equation*}
			\begin{aligned}
				I_1 & =\sum_{j=0}^{+\infty} \gamma^j \sum_{s_{j+1},
					r_j,m_j,a_j,s_j} \{ r_j+\gamma  V^{\pi}(s_{j+1})- Q^{\pi}(m_j,a_j,s_j) \}
				\frac{p_{m,\theta_0}(m_j|\pi(s_j),s_j)}{p_{m,\theta_0}(m_j|a_j,s_j) } \\
				& \times p_{s,r,\theta_0}^*(s_{j+1},r_j|m_j,a_j,s_j) p_{m,\theta_0}(m_j|a_j,s_j) p_{a,\theta_0}^*(a_j|s_j)p_{j,\theta_0}^{\pi}(s_j) \nabla_{\theta} \log p_{s,r,\theta_0}^*(s_{j+1},r_j|m_j,a_j,s_j).\\
			\end{aligned}
		\end{equation*}
		Second, we notice that 
		\begin{equation*}
			\begin{aligned}
				&\sum_{j=0}^{+\infty} \gamma^j \sum_{s_{j+1},
					r_j,m_j,a_j,s_j} \left[ r_j+\gamma  V^{\pi}(s_{j+1})- Q^{\pi}(m_j,a_j,s_j) \right] 
				\frac{p_{m,\theta_0}(m_j|\pi(s_j),s_j)}{p_{m,\theta_0}(m_j|a_j,s_j) } \\
				& \times p_{s,r,\theta_0}^*(s_{j+1},r_j|m_j,a_j,s_j) p_{m,\theta_0}(m_j|a_j,s_j) p_{a,\theta_0}(a_j|s_j)p_{j,\theta_0}^{\pi}(s_j)  \nabla_{\theta}  \log p_{m, \theta_0}\left(m_j \mid a_j, s_j\right) \\
			\end{aligned}
		\end{equation*}
		\begin{equation*}
			\begin{aligned}
				&=\sum_{j=0}^{+\infty} \gamma^j \sum_{m_j,a_j,s_j} 
				\frac{p_{m,\theta_0}(m_j|\pi(s_j),s_j)}{p_{m,\theta_0}(m_j|a_j,s_j) } p_{m,\theta_0}(m_j|a_j,s_j) p_{a,\theta_0} (a_j|s_j)p_{j,\theta_0}^{\pi}(s_j)   \\
				& \times  \nabla_{\theta}  \log p_{m, \theta_0}\left(m_j \mid a_j, s_j\right) \sum_{s_{j+1}, r_j}    \left[ r_j+\gamma  V^{\pi}(s_{j+1})- Q^{\pi}(m_j,a_j,s_j) \right] p_{s,r,\theta_0}(s_{j+1},r_j|m_j,a_j,s_j)  \\
			\end{aligned}
		\end{equation*}
		\begin{equation*}
			\begin{aligned}
				&=\sum_{j=0}^{+\infty} \gamma^j \sum_{\tau_{j}} 
				\frac{p_{m,\theta_0}(m_j|\pi(s_j),s_j)}{p_{m,\theta_0}(m_j|a_j,s_j) } p_{m,\theta_0}(m_j|a_j,s_j) p_{a,\theta_0} (a_j|s_j)p_{t,\theta_0}^{\pi}(s_j)  \nabla_{\theta}  \log p_{m, \theta_0}\left(m_j \mid a_j, s_j\right) \\
				& \times \left[ \Mean \{R_j+\gamma V^{\pi}(S_{j+1})|M_j=m_j,A_j=a_j,S_j=s_j\} - Q^{\pi}(m_j,a_j,s_j) \right] =0. \\
			\end{aligned}
		\end{equation*}
		The above relation also holds if we replace $\nabla_{\theta}  \log p_{m, \theta_0}\left(m_j \mid a_j, s_{j}\right)$ with $\nabla_{\theta}  \log  p_{a, \theta_0}^*\left(a_j \mid s_j\right)$. It follows that
		\begin{equation*}
			\begin{aligned}
				I_1  & = \sum_{j=0}^{+\infty} \gamma^j \sum_{s_{j+1},
					r_j,m_j,a_j,s_j} \left[ r_j+\gamma  V^{\pi}(s_{j+1})- Q^{\pi}(m_j,a_j,s_j) \right] 
				\frac{p_{m,\theta_0}(m_j|\pi(s_j),s_j)}{p_{m,\theta_0}(m_j|a_j,s_j) } \\
				& \times p_{s,r,\theta_0}^*(s_{j+1},r_j|m_j,a_j,s_j) p_{m,\theta_0}(m_j|a_j,s_j) p_{a,\theta_0}^*(a_j|s_j)
				p_{j,\theta_0}^{\pi}(s_j)
				S\left(s_{j+1}, r_{j}, m_{j}, a_{j}\mid s_j \right),\\
			\end{aligned}
		\end{equation*}	
		
		where $S(s',r,m,a|s)$ denotes the conditional score function of $(S_{t+1},R_t,M_t,A_t)$ given $S_t$. We rewrite $p_{j,\theta_0}^{\pi}(s_j)$ as the product of $p_{j,\theta_0}^{\pi}(s_j)/p_{D}(s_j)$ and $p_{D}(s_j)$. 
		Using the change of measure theorem, the RHS can be further represented as 
		\begin{eqnarray*}
			\sum_{j=0}^{+\infty}\gamma^j \Mean \{R_j+\gamma V^{\pi}(S_{j+1})-Q^{\pi}(M_j,A_j,S_j)\}\frac{p_{m}(M_j|\pi(S_j),S_j)}{p_M(M_j|A_j,S_j)}\frac{p_{j,\theta_0}^{\pi}(S_j)}{p_{D}(S_j)}S(S_{j+1},R_j,M_j,A_j|S_j)
		\end{eqnarray*}	
		Recall that $\omega^{\pi}(\bullet)=(1-\gamma) \sum_{t\ge 0}  \gamma^t \frac{p_t^{\pi}(s)}{p_{D}(s)}$ and $\rho \left(M,A,S\right) = \frac{p_{m}(M_j|\pi(S_j),S_j)}{p_M(M_j|A_j,S_j)}$, we can further write $I_1$ as 
		\begin{equation*}
			I_1  =\frac{1}{1-\gamma}\Mean \{R+\gamma V^{\pi}(S')- Q^{\pi}(M,A,S)\}\omega^{\pi}(S)\rho \left(M,A,S\right) S(S',R,M,A\mid S) 
		\end{equation*}
		where $(S,A,M,R,S')$ denotes an arbitrary data transaction from the mixture distribution. 
		Next, combining the fact that the expectation of a score function equals to zero together with the Markov property, we obtain that
		\begin{equation*}
			I_1  =\frac{1}{1-\gamma}\Mean\left\{ [R+\gamma V^{\pi}(S')- Q^{\pi} \left(M,A,S\right) ]\omega^{\pi}(S)\rho \left(M,A,S\right) S(\bar{O}_{T-1}) \right\}.
		\end{equation*}
		Notice  $(S,A,M,R,S')$ denotes an arbitrary data transaction from the mixture distribution, we have
		\begin{equation*}
			\begin{aligned}
				I_1 & =\Mean \left\{ \frac{1}{T} \sum\limits_{t=1}^{T} \frac{1}{1-\gamma}[R_t+\gamma V^{\pi}(S_{t+1})- Q^{\pi} \left(M_t,A_t,S_t\right) ]\omega^{\pi}(S_t)\rho \left(M_t,A_t,S_t\right) S(\bar{O}_{T-1}) \right\}\\
				&= \Mean \left\{\left[  \frac{1}{T} \sum\limits_{t}  \psi_1(O_{t})  \right] S(\bar{O}_{T-1})\right\}.
			\end{aligned}.
		\end{equation*}

		\subsubsection{Derivations of  $I_2$} 
		In the following, we focus on proving \ref{eqn:eqnI2}. Similarly as the derivations of  $I_1$, we note that
		\begin{eqnarray*}
			\sum_{s_0}\left\{\prod_{k=0}^j {\color{black}p_{\theta_0}(s_{k+1},r_k,m_k,a_k|s_k)}\right\}\nu_{\theta_0}(s_0) = p_{\theta_0}(s_{j+1},r_j,m_j,a_j|s_j) p_{t,\theta_0}^{\pi}(s_j) \\
			=p_{s,r,\theta_0}^*(s_{j+1},r_j|m_j,a_j,s_j)p_{m,\theta_0}(m_j|\pi(s_j),s_j)p_{a,\theta_0}^*(a_j|s_j)p_{t,\theta_0}^{\pi}(s_j)
		\end{eqnarray*}
		which is  the probability of $\{S_{j+1}=s_{j+1}, R_j=r_j, M_j=m_j,A_j=a_j\}$ under target policy $\pi$. It follows that
		\begin{equation*}
			\begin{aligned}
				& \sum_{s_{j+1},
					r_j,m_j,a_j,s_j} \{r_j+\gamma  V^{\pi}(s_{j+1})\} p_{s,r,\theta_0}(s_{j+1},r_j|m_j,a_j,s_j)p_{m,\theta_0}(m_j|\pi(s_j),s_j)p_{a,\theta_0}(a_j|s_j) \\
				& \times p_{t,\theta_0}^{\pi}(s_j)\nabla_{\theta} \log p_{m, \theta_0}(m_j|\pi(s_j),s_j)\\
				& =\sum_{ r_j,m_j,a_j,s_j} \Mean \left\{   \left[ R_j+\gamma  V^{\pi}(S_{j+1}) \right] \mid m_j,a_j,s_j  \right\}    p _{m,\theta_0}(m_j|\pi(s_j),s_j)p_{a,\theta_0}(a_j|s_j)p_{t,\theta_0}^{\pi}(s_j) \\
				& \times  \nabla_{\theta} \log p_{m, \theta_0}(m_j|\pi(s_j),s_j)\\
				& = \sum_{m_j, s_j} \left\{  \sum_{a_j}   \Mean \left\{   \left[ R_j+\gamma  V^{\pi}(S_{j+1}) \right] \mid m_j,a_j,s_j  \right\} p_{a,\theta_0}(a_j|s_j) \right\}    p _{m,\theta_0}(m_j|\pi(s_j),s_j)p_{t,\theta_0}^{\pi}(s_j) \\
				& \times
				\nabla_{\theta} \log p_{m, \theta_0}(m_j|\pi(s_j),s_j).\\
			\end{aligned}
		\end{equation*}
		Replace $p _{m,\theta_0}(m_j|\pi(s_j),s_j)$ with $I\{a_j^{*} = \pi(s_j) \}  p _{m,\theta_0}(m_j|a_j^{*},s_j)$ and rewrite $I\{a_j^{*} = \pi(s_j) \}$ as the product of $\frac{I\{a_j^{*} = \pi(s_j) \}}{p_{a,\theta_0}(a_j^{*}|s_j)}$ and $p_{a,\theta_0}(a_j^{*}|s_j)$, we have
		\begin{equation*}
			\begin{aligned}
				I_2 & = \sum_{j=0}^{+\infty} \gamma^j \sum_{m_j, s_j}  \sum_{a_j^{*}} \left\{ \sum_{a_j} \Mean \left\{   \left[ R_j+\gamma  V^{\pi}(R_{j+1}) \right] \mid m_j,a_j,s_j  \right\} p_{a,\theta_0}(a_j|s_j) \right\}\frac{I\{a_j^{*} = \pi(s_j) \}}{p_{a,\theta_0}(a_j^{*}|s_j)}    \\  
				&  \times p _{m,\theta_0}(m_j|a_j^{*},s_j) p_{a,\theta_0}(a_j^{*}|s_j) p_{t,\theta_0}^{\pi}(s_j)   \nabla_{\theta} \log p_{m, \theta_0}(m_j|a_j^{*},s_j).\\
			\end{aligned}
		\end{equation*}
		Using the fact that the expectation of a score function is zero, we have 
		\begin{equation*}
			\begin{aligned}
				&  \sum_{j=0}^{+\infty} \gamma^j \sum_{m_j, s_j}  \sum_{a_j^{*}} \left\{ \sum_{a_{t},m_t^{*}}  \Mean \{R_t+\gamma V^{\pi}(S_{t+1})|M_t=m_t^{*},A_t=a_t,s_t\} p_{m,\theta_0}(m_t^{*}|a_j^{*},s_t) p_{a,\theta_0}(a_j|s_j)  \right\} \\  
				&  \times \frac{I\{a_j^{*} = \pi(s_j) \}}{p_{a,\theta_0}(a_j^{*}|s_j)}  p _{m,\theta_0}(m_j|a_j^{*},s_j) p_{a,\theta_0}(a_j^{*}|s_j) p_{t,\theta_0}^{\pi}(s_j)   \nabla_{\theta} \log p_{m, \theta_0}(m_j|a_j^{*},s_j)\\
				&  = \sum_{j=0}^{+\infty} \gamma^j \sum_{s_j}  \sum_{a_j^{*}} \left\{ \sum_{a_{t},m_t^{*}}  \Mean \{R_t+\gamma V^{\pi}(S_{t+1})|M_t=m_t^{*},A_t=a_t,s_t\} p_{m,\theta_0}(m_t^{*}|a_j^{*},s_t) p_{a,\theta_0}(a_j|s_j)  \right\} \\  
				&  \times \frac{I\{a_j^{*} = \pi(s_j) \}}{p_{a,\theta_0}(a_j^{*}|s_j)}  p_{a,\theta_0}(a_j^{*}|s_j) p_{t,\theta_0}^{\pi}(s_j)  \sum_{m_j}p _{m,\theta_0}(m_j|a_j^{*},s_j)  \nabla_{\theta} \log p_{m, \theta_0}(m_j|a_j^{*},s_j) = 0.\\
			\end{aligned}
		\end{equation*}
		Thus,
		\begin{equation*}
			\begin{aligned}
				I_2 & =\sum_{j=0}^{+\infty} \gamma^j \sum_{m_j, s_j}  \sum_{a_j^{*}} \left\{ \sum_{a_j} \Mean \left\{   \left[ R_j+\gamma  V^{\pi}(S_{j+1}) \right] \mid m_j,a_j,s_j  \right\} p_{a,\theta_0}(a_j|s_j)  \right.\\
				& - \left.\sum_{a_{t},m_t^{*}}  \Mean \{R_t+\gamma V^{\pi}(S_{t+1})|M_t=m_t^{*},A_t=a_t,s_t\} p_{m,\theta_0}(m_t^{*}|a_j^{*},s_t) p_{a,\theta_0}(a_j|s_j)  \right\} \\  
				&  \times \frac{I\{a_j^{*} = \pi(s_j) \}}{p_{a,\theta_0}(a_j^{*}|s_j)}  p _{m,\theta_0}(m_j|a_j^{*},s_j) p_{a,\theta_0}(a_j^{*}|s_j) p_{t,\theta_0}^{\pi}(s_j)   \nabla_{\theta} \log p_{m, \theta_0}(m_j|a_j^{*},s_j),\\
			\end{aligned}
		\end{equation*}
		Replace $\Mean \left\{   \left[ R_j+\gamma  V^{\pi}(S_{j+1}) \right] \mid m_j,a_j,s_j  \right\}$ with $Q^{\pi}(m_j,a_j,s_j)$, and replace $\Mean \{R_t+\gamma V^{\pi}(S_{t+1})|M_t=m_t^{*},A_t=a_t,s_t\}$ with $Q^{\pi}(m_t^{*},a_t,s_t)$. Then 
		use the fact that the expectation of a score function is zero again, we have
		\begin{equation*}
			\begin{aligned}
				I_2  & = \sum_{j=0}^{+\infty} \gamma^j \sum_{ s_{j+1},r_{j}, m_j, s_j}  \sum_{a_j^{*}} \frac{I\{a_j^{*} = \pi(s_j) \}}{p_{a,\theta_0}(a_j^{*}|s_j)} \left\{ \sum_{a_j}Q^{\pi}(m_j,a_j,s_j)  p_{a,\theta_0}(a_j|s_j) \right. \\
				& \left. -\sum_{a_j,m_t^{*}} Q^{\pi}(m_t^{*},a_t,s_t) p_{m,\theta_0}(m_t^{*}|a_j^{*},s_t)  p_{a,\theta_0}(a_j|s_j)   \right\}\\  
				&  \times  p_{\theta_0}(s_{j+1},r_j,m_j,a_j\mid s_j) \frac{p_{t,\theta_0}^{\pi}(s_j)  }{p_{D}(s_j)}p_{D}(s_j)\nabla_{\theta} \log  p_{\theta_0}\left(s_{j+1}, r_{j},  m_{j}, a_{j}^{*} \mid s_j\right). \\
			\end{aligned}
		\end{equation*}
		Similarly as the derivations of  $I_1$, we rewrite $p_{j,\theta_0}^{\pi}(s_j)$ as the product of $p_{j,\theta_0}^{\pi}(s_j)/p_{D}(s_j)$ and $p_{D}(s_j)$.  Using the change of measure theorem, $I_2$ can be further represented as 
		\begin{equation*}
			\begin{aligned}
				& \sum_{j=0}^{+\infty} \gamma^j  \Mean \frac{\mathbb{I}(A=\pi(S))}{p_{a,\theta_0}(A|S)} \sum_{a} p_{a,\theta_0}(a|S) \left[Q^{\pi}(M,a,S) -\sum_{m} p_m(m|A,S)Q^{\pi}(m,a,S)\right] \\ & \times \frac{p_{j,\theta_0}^{\pi}(S)}{p_{D}(S)} S(S',R,M,A\mid S) \\
				=& \frac{1}{1-\gamma} \Mean  \omega^{\pi}(S) \frac{\mathbb{I}(A=\pi(S))}{p_{a,\theta_0}(A|S)} \sum_{a} p_{a,\theta_0}(a|S) \left[Q^{\pi}(M,a,S) -\sum_{m} p_m(m|A,S)Q^{\pi}(m,a,S)\right]\\
				& \times S(S',R,M,A\mid S),
			\end{aligned}
		\end{equation*}
		where $(S,A,M,R,S')$ denotes an arbitrary data transaction from the mixture distribution. Thus, 
		
		Next, combining the fact that the expectation of a score function equals to zero together with the Markov property, we obtain that $I_2$ equals to 
		\begin{eqnarray*}
			\frac{1}{1-\gamma} \Mean  \omega^{\pi}(S) \frac{\mathbb{I}(A=\pi(S))}{p_{a,\theta_0}(A|S)} \sum_{a} p_{a,\theta_0}(a|S)  \left[Q^{\pi}(M,a,S) -\sum_{m} p_{m, \theta_0}(m|A,S)Q^{\pi}(m,a,S)\right]S(\bar{O}_{T-1}).
		\end{eqnarray*}
		Notice  $(S,A,M,R,S')$ denotes an arbitrary data transaction from the mixture distribution, so $I_2 $ can be further expressed as 
		\begin{equation*}
			\begin{aligned}
				& \Mean \left\{ \frac{1}{T} \sum\limits_{t=1}^{T} \frac{1}{1-\gamma}\omega^{\pi}(S) \frac{\mathbb{I}(A=\pi(S))}{p_{a,\theta_0}(A|S)} \sum_{a} p_{a,\theta_0}(a|S) \left[Q^{\pi}(M,a,S) \right. \right. \\
				& \left. \left. -\sum_{m} p_{m, \theta_0}(m|A,S)Q^{\pi}(m,a,S)\right]  S(\bar{O}_{T-1})
				\right\}= \Mean \left\{\left[  \frac{1}{T} \sum\limits_{t}  \psi_2(O_{t})  \right] S(\bar{O}_{T-1})\right\}.
			\end{aligned}
		\end{equation*} 
		Note that when the target policy is a random policy, $\psi_2(O)$ has the form as 
		\begin{equation*}
			\frac{1}{1-\gamma} \omega^{\pi}(S) \frac{\pi(A \mid S)}{p_{a,\theta_0}(A|S)} \sum_{a} p_{a,\theta_0}(A|S) \times\left\{Q^{\pi}(M, a, S)-\sum_{m} p_{m,\theta_0}(m \mid A, S) Q^{\pi}(m, a, S)\right\},
		\end{equation*}
		and the proof is similar.
		
		\subsubsection{Derivations of  $I_3$} 
		In the following, we focus on proving \ref{eqn:eqnI3}. Similarly as before, we note that 
		\begin{eqnarray*}
			\sum_{s_0}\left\{\prod_{k=0}^j {\color{black}p_{\theta_0}(s_{k+1},r_k,m_k,a_k|s_k)}\right\}\nu_{\theta_0}(s_0)  = p_{\theta_0}(s_{j+1},r_j,m_j,a_j|s_j) p_{t,\theta_0}^{\pi}(s_j) \\
			= p_{s,r,\theta_0}^*(s_{j+1},r_j|m_j,a_j,s_j)p_{m,\theta_0}(m_j|\pi(s_j),s_j)p_{a,\theta_0}^*(a_j|s_j)p_{t,\theta_0}^{\pi}(s_j)
		\end{eqnarray*}
		which is  the probability of $\{S_{j+1}=s_{j+1}, R_j=r_j, M_j=m_j,A_j=a_j\}$ under target policy $\pi$. It follows that
		\begin{equation*}
			\begin{aligned}
				I_3 &=\sum_{j=0}^{+\infty} \gamma^j \sum_{s_{j+1},
					m_j,a_j,s_j} \{r_j+\gamma  V^{\pi}(s_{j+1})\} p_{s,r,\theta}(s_{j+1},r_j|m_j,a_j,s_j)p_{m,\theta}(m_j|\pi(s_j),s_j) \\
				&\times p_{a,\theta_0}(a_j|s_j)p_{t,\theta_0}^{\pi}(s_j) \nabla_{\theta} \log p_{a, \theta_0}(a_j|s_j),
			\end{aligned}
		\end{equation*}
		which can be further expressed as 
		\begin{equation*}
			\begin{aligned}
				I_3 & =\sum_{j=0}^{+\infty} \gamma^j \sum_{m_j,a_j,s_j} \Mean \left\{   \left[ R_j+\gamma  V^{\pi}(S_{j+1}) \right] \mid m_j,a_j,s_j  \right\}    p _{m,\theta_0}(m_j|\pi(s_j),s_j)p_{a,\theta_0}(a_j|s_j)   \\
				& \times  p_{t,\theta_0}^{\pi}(s_j) \nabla_{\theta} \log p_{a, \theta_0}(a_j|s_j)\\
				& =\sum_{j=0}^{+\infty} \gamma^j \sum_{a_j,s_j} \left\{ \sum_{m_j} \Mean \left\{   \left[ R_j+\gamma  V^{\pi}(S_{j+1}) \right] \mid m_j,a_j,s_j  \right\}    p _{m,\theta_0}(m_j|\pi(s_j),s_j) \right\}   p_{a,\theta_0}(a_j|s_j)\\
				& \times  p_{t,\theta_0}^{\pi}(s_j)  \nabla_{\theta} \log p_{a, \theta_0}(a_j|s_j).\\
			\end{aligned}
		\end{equation*}
		Since the expectation of a score function is zero, we have
		\begin{equation*}
			\begin{aligned}
				& \sum_{j=0}^{+\infty} \gamma^j \sum_{a_j,s_j} \left\{ \sum_{m_j,a_j} \Mean \left\{   \left[ R_j+\gamma  V^{\pi}(S_{j+1}) \right] \mid m_j,a_j,s_j  \right\}   p _{m,\theta_0}(m_j|\pi(s_j),s_j)  \right\} p_{a,\theta_0}(a_j|s_j)\\
				& \times p_{t,\theta_0}^{\pi}(s_j)   \nabla_{\theta} \log p_{a, \theta_0}(a_j|s_j)\\
				= &\sum_{j=0}^{+\infty} \gamma^j \sum_{s_j} \left\{ \sum_{m_j,a_j} \Mean \left\{   \left[ R_j+\gamma  V^{\pi}(S_{j+1}) \right] \mid m_j,a_j,s_j  \right\}   p _{m,\theta_0}(m_j|\pi(s_j),s_j)  \right\} p_{t,\theta_0}^{\pi}(s_j)   \\
				& \times \sum_{a_j}p_{a,\theta_0}(a_j|s_j)\nabla_{\theta} \log p_{a, \theta_0}(a_j|s_j)= 0.
			\end{aligned}
		\end{equation*}
		Thus we have
		\begin{equation*}
			\begin{aligned}
				I_3& =\sum_{j=0}^{+\infty} \gamma^j \sum_{a_j,s_j} \left\{ \sum_{m_j} \Mean \left\{   \left[ R_j+\gamma  V^{\pi}(S_{j+1}) \right] \mid m_j,a_j,s_j  \right\}   p _{m,\theta_0}(m_j|\pi(s_j),s_j) \right. \\
				& \left. - \sum_{m_j,a_j} \Mean \left\{   \left[ R_j+\gamma  V^{\pi}(S_{j+1}) \right] \mid m_j,a_j,s_j  \right\}   p _{m,\theta_0}(m_j|\pi(s_j),s_j)  \right\} p_{a,\theta_0}(a_j|s_j)p_{t,\theta_0}^{\pi}(s_j)  \\
				& \times \nabla_{\theta} \log p_{a, \theta_0}(a_j|s_j)\\
				&  = \sum_{j=0}^{+\infty} \gamma^j \sum_{m_j^{*},s_{j+1},r_j} \sum_{a_j,s_j} \left\{ \sum_{m_j} \Mean \left\{   \left[ R_j+\gamma  V^{\pi}(S_{j+1}) \right] \mid m_j,a_j,s_j  \right\}   p _{m,\theta_0}(m_j|\pi(s_j),s_j) \right. \\
				& \left. - \sum_{m_j,a_j} \Mean \left\{   \left[ R_j+\gamma  V^{\pi}(S_{j+1}) \right] \mid m_j,a_j,s_j  \right\}   p _{m,\theta_0}(m_j|\pi(s_j),s_j)  \right\} \\
				& p_{s,r,\theta}(s_{j+1},r_j|m_j^{*},a_j,s_j) p _{m,\theta_0}(m_j^{*}|a_j,s_j)  p_{a,\theta_0}(a_j|s_j)p_{t,\theta_0}^{\pi}(s_j)  \nabla_{\theta} \log p_{a, \theta_0}(a_j|s_j). \\
			\end{aligned}
		\end{equation*}
		Using similar trick about score functions again and we rewrite $p_{t,\theta_0}^{\pi}(s_j)$ as the product of $ \frac{p_{t,\theta_0}^{\pi}(s_j) }{p_{D}(s_j)}$ and $ p_{D}(s_j)$, we have
		\begin{equation*}
			\begin{aligned}
				I_3 & = \sum_{j=0}^{+\infty} \gamma^j \sum_{m_j^{*},s_{j+1},r_j,a_j,s_j} \left\{ \sum_{m_j} \Mean \left\{   \left[ R_j+\gamma  V^{\pi}(S_{j+1}) \right] \mid m_j,a_j,s_j  \right\}   p _{m,\theta_0}(m_j|\pi(s_j),s_j) \right. \\
				& \left. - \sum_{m_j,a_j} \Mean \left\{   \left[ R_j+\gamma  V^{\pi}(S_{j+1}) \right] \mid m_j,a_j,s_j  \right\}   p _{m,\theta_0}(m_j|\pi(s_j),s_j)  \right\} p_{s,r,\theta}(s_{j+1},r_j,m_j^{*},a_j|s_j)  \\
				&  \frac{p_{t,\theta_0}^{\pi}(s_j) }{p_{D}(s_j)} p_{D}(s_j)\nabla_{\theta} \log  p_{s, r, \theta}\left(s_{j+1}, r_{j}, m_{j}^{*}, a_{j}\mid  s_{j}\right). \\
			\end{aligned}
		\end{equation*}
		Replace $\Mean \left\{   \left[ R_j+\gamma  V^{\pi}(S_{j+1}) \right] \mid m_j,a_j,s_j  \right\} $ with $Q^{\pi}(m_j,a_j,s_j)$, then
		\begin{equation*}
			\begin{aligned}
				I_3 & = \sum_{j=0}^{+\infty} \gamma^j \sum_{m_j^{*},s_{j+1},r_j,a_j,s_j} \left\{ \sum_{m_j} Q^{\pi}(m_j,a_j,s_j)   p _{m,\theta_0}(m_j|\pi(s_j),s_j) \right.\\
				& \left. - \sum_{m_j,a_j} Q^{\pi}(m_j,a_j,s_j)  p _{m,\theta_0}(m_j|\pi(s_j),s_j)  \right\} p_{s,r,\theta}(s_{j+1},r_j,m_j^{*},a_j|s_j) \\
				&  \times   \frac{p_{t,\theta_0}^{\pi}(s_j) }{p_{D}(s_j)} p_{D}(s_j)\nabla_{\theta} \log  p_{s, r, \theta}\left(s_{j+1}, r_{j}, m_{j}^{*}, a_{j}\mid  s_{j}\right). \\
			\end{aligned}
		\end{equation*}
		
		Using the change of measure theorem, $I_3$ can be further represented as 
		\begin{equation*}
			\begin{aligned}
				I_3 & = \sum_{j=0}^{+\infty} \gamma^j \Mean \left\{ \sum_{m_j} Q^{\pi}(m_j,A,S)   p _{m,\theta_0}(m_j|\pi(S),S) - \sum_{m_j,a_j} Q^{\pi}(m_j,a_j,S)  \right\} \\
				&  \times  p _{m,\theta_0}(m_j|\pi(S),S)  \frac{p_{t,\theta_0}^{\pi}(S) }{p_{D}(S)} S\left(S', R,M, A\mid  S\right) = \frac{1}{1-\gamma} \Mean  \omega^{\pi}(S)  \sum_{m_j}p _{m,\theta_0}(m_j|\pi(S),S)    \\
				&  \times  \left\{  Q^{\pi}(m_j,A,S)    - \sum_{a_j} Q^{\pi}(m_j,a_j,S)  p _{m,\theta_0}(m_j|\pi(S),S)  \right\}  S\left(S', R,M, A\mid  S\right), 
			\end{aligned}
		\end{equation*}
		where $(S,A,M,R,S')$ denotes an arbitrary data transaction from the mixture distribution. 
		
		Next, combining the fact that the expectation of a score function equals to zero together with the Markov property, we obtain that $I_3$ equals to
		\begin{eqnarray*}
			\frac{1}{1-\gamma} \Mean  \omega^{\pi}(S)  \sum_{m_j}p _{m,\theta_0}(m_j|\pi(S),S)  \left\{  Q^{\pi}(m_j,A,S)   - \sum_{a_j} Q^{\pi}(m_j,a_j,S)  p _{m,\theta_0}(m_j|\pi(S),S)  \right\}   S(\bar{O}_{T-1}) .
		\end{eqnarray*}
		Notice  $(S,A,M,R,S')$ denotes an arbitrary data transaction from the mixture distribution, so $I_3$ can be further expressed as 
		\begin{eqnarray*}
			\Mean \left\{ \frac{1}{T} \sum\limits_{t=1}^{T} \frac{1}{1-\gamma}\omega^{\pi}(S) \sum_{m_j}p _{m,\theta_0}(m_j|\pi(S),S)  \left\{  Q^{\pi}(m_j,A,S)   \right.  \right.\\\left. - \sum_{a_j} Q^{\pi}(m_j,a_j,S)  p _{m,\theta_0}(m_j|\pi(S),S)  \}   S(\bar{O}_{T-1}) \right\},
		\end{eqnarray*}
		which equals to 
		\begin{equation*}
			\Mean \left\{\left[  \frac{1}{T} \sum\limits_{t}  \psi_3(O_{t})  \right] S(\bar{O}_{T-1})\right\}.
		\end{equation*}
		Then completes the proof.

\end{document}